\documentclass[twoside,english,3p]{elsarticle}
\biboptions{numbers,sort&compress}

\usepackage{subfigure}
\usepackage{amssymb}
\usepackage{amsthm}
\usepackage{amsmath}
\usepackage{booktabs} 
\usepackage{caption}
\usepackage{algorithm}
\usepackage{algpseudocode}
\usepackage{makecell}
\usepackage{hyperref}
\usepackage[hyphenbreaks]{breakurl}
\newtheorem{remark}{Remark}
\journal{Computer Methods in Applied Mechanics and Engineering}

\begin{document}
\subfigcapmargin=-1cm
\raggedbottom 
\begin{frontmatter}



\title{BINN: A deep learning approach for computational mechanics problems based on boundary integral equations}


\author[inst1]{Jia Sun}
\author[inst1]{Yinghua Liu}
\author[inst1,inst2]{Yizheng Wang}
\author[inst1]{Zhenhan Yao}
\author[inst1]{Xiaoping Zheng\corref{cor1}}
\ead{zhengxp@mail.tsinghua.edu.cn}
\cortext[cor1]{Corresponding author}
\affiliation[inst1]{organization={Department of Engineering Mechanics, Tsinghua University },
            city={Beijing},
            postcode={100084}, 
            country={China}}
\affiliation[inst2]{organization={Microsoft Research AI4Science},
            city={Beijing},
            postcode={100080}, 
            country={China}}

\begin{abstract}
We proposed the boundary-integral type neural networks (BINN) for the boundary value problems in computational mechanics. The boundary integral equations are employed to transfer all the unknowns to the boundary, then the unknowns are approximated using neural networks and solved through a training process. The loss function is chosen as the residuals of the boundary integral equations. Regularization techniques are adopted to efficiently evaluate the weakly singular and Cauchy principle integrals in boundary integral equations. Potential problems and elastostatic problems are mainly concerned in this article as a demonstration. The proposed method has several outstanding advantages: First, the dimensions of the original problem are reduced by one, thus the freedoms are greatly reduced. Second, the proposed method does not require any extra treatment to introduce the boundary conditions, since they are naturally considered through the boundary integral equations. Therefore, the method is suitable for complex geometries. Third, BINN is suitable for problems on the infinite or semi-infinite domains. Moreover, BINN can easily handle heterogeneous problems with a single neural network without domain decomposition.

\end{abstract}

\begin{highlights}
\item A BIE-based deep learning framework is proposed to solve boundary value problems.
\item The advantage of BINN is that it can reduce the problem dimension by 1.
\item Boundary conditions are automatically considered in the proposed method. 
\item BINN is convenient to solve problems with infinite and semi-infinite regions.
\item BINN can conveniently solve heterogeneous problems with only a single network.
\end{highlights}

\begin{keyword}
Physics-informed neural networks\sep Deep learning\sep Boundary integral equations\sep Mesh-free method \sep Inverse statement \sep Reduced-order modeling
\end{keyword}

\end{frontmatter}


\section{Introduction}
\label{sec1}
In the past decades, machine learning algorithms have been widely employed in various tasks such as computer vision \cite{RN64}, natural language processing \cite{RN114}, and image synthesis \cite{RN113}. 
The universal approximation theorem \cite{1990Universal,RN86} indicates the powerful capacity of feed-forward neural networks for approximating any continuous functions with arbitrary accuracy. In recent years, a novel machine learning framework that introduces the laws of physics described with partial differential equations (PDEs) as constraints into neural networks, namely the physics-informed neural networks (PINNs) \cite{RN29,Karniadakis_1}, has gained much attention. Research that uses neural networks in solving PDEs can be traced back to the last century \cite{RN98,RN100,RN101}. However, these earliest works lacked attention at the time due to the limitation of the hardware and software. In the last decades, with the progress of artificial intelligence, several mature machine learning frameworks like Pytorch \cite{Paszke_1} and TensorFlow \cite{Abadi_1} have been developed and make it convenient to build and train a neural network. Raissi et.al. \cite{RN29} formally put forward the theory of PINNs. They systematically stated the key idea of PINNs and showed its powerful potential on several classical PDEs, including forward and inverse problems.  Over the past few years, many related works were published. E and Yu \cite{RN97} proposed the Deep Ritz method that solves variational problems with deep learning algorithms. Samaniego et.al. \cite{Samaniego_1} proposed the deep energy method (DEM) to solve variational problems and employed it on various physical problems.  Lu et.al. \cite{Lu1} published DeepXDE, a python library for PINNs. Lu et.al. \cite{Lu_2021} proposed DeepONet, a framework that could directly learn the nonlinear operator instead of a specific function, to name a few. For problems with discontinuity, Jagtap et.al. \cite{RN53} suggested the idea of the subdomain, where each subdomain was allocated with an individual network that coupled with others on the interface. Wang et.al. \cite{RN115} extended the idea of subdomain into variational problems.

There are two modern implementations among PINN-based methods, namely the Deep Collocation method (DCM) \cite{RN29} and the Deep Ritz method (DRM) \cite{RN97,Samaniego_1}. 
The DCM can be derived from the original statement (also known as the strong form) of the weighted residual method (WRM), which is employed in the original literature of PINNs \cite{RN29}. Some other PINN-based methods are also based on the original statement but with different test functions, such as the deep Galerkin method \cite{Justin_1}, VPINN \cite{Ehsan_1} and hp-VPINN \cite{Ehsan_2}. In the DCM, the neural networks are constrained to meet the governing equations and boundary conditions (BCs) at a set of collocation points. In boundary value problems, the loss function is mainly comprised of three parts \cite{Ehsan_3}: the residuals of the government equations at the interior collocation points; the residual error of the essential boundary conditions at the boundary collocation points; the residual error of the natural boundary conditions at the boundary collocation points. Thus the loss function can be taken as the weighted summation of the mean squared error among the three sets of collocation points. The Deep Collocation method based on the original statement is a powerful algorithm that can be applied to any PDEs \cite{RN29,Sina_1,Mao_1}. In practice, a large set of PDEs have an equivalent variational form, thus the Deep Ritz method based on the weak statement (also known as the weak form) of the weighted residual method can be performed \cite{RN97,RN105}, where neural networks are still employed as the trail function, and an energy functional will be computed by integrating among a set of quadrature points on the interior domain and the natural boundary. The solution of the PDEs will be exactly the minimum of the functional if the trail function is admissible, which requires the approximate function to meet the essential boundary conditions. The essential boundary conditions can be imposed through a penalty term \cite{RN97} or Nitsche's method \cite{RN110}.
Thus the loss function in DRM usually contains the energy functional and a residual term to apply the essential boundary conditions. 
The Deep Ritz method has the advantage of less requirement for the continuity to the approximation function \cite{Li_1}.  Furthermore, the natural boundary conditions are naturally considered in the energy functional. But not all the PDEs have a variational form. 

In both Deep Collocation and Deep Ritz methods, a single loss function will be trained to minimize more than one objective functions, including the interior term involving the governing equations and the boundary term involving the boundary conditions. Therefore, the weights of the terms are important to balance the gradient of each objective function in the gradient descent algorithm. There are many studies that focus on improving the evolution of the gradient during the training process of PINN-based methods \cite{RN102,RN103,Ameya_1,RN136,Mao_1}, such as using a adaptive strategy to decide the weights of the terms in the loss function \cite{RN102}, using adaptive activation functions \cite{Ameya_1} and so on. 
Alternatively, the exact imposition of the boundary conditions in the approximate function is an effective idea to reduce the cost on selecting the artificial weights, which eliminates all the boundary terms in the loss function. Such strategies have already been proposed in some early works  \cite{RN101} of PINNs. The key idea is to divide the approximate function into two parts \cite{RN108}: The networks are embedded in the first part that is specially constructed to satisfy the homogeneous boundary conditions at corresponding boundaries, while the given boundary conditions are satisfied in the second part. Such construction is easy for simple geometries such as rectangular region, and there are plenty of works aimed at extending the strategy to arbitrary-complex domains. For the construction of the first part, Lagaris et.al \cite{RN106} used a radial basis function network that vanished on the boundary. The methods of using an approximate distance function to the boundary also gain much attention. The distance function can be evaluated using thin plates splines mapping function \cite{RN99}, radial basis function \cite{RN105}, another neural network  \cite{RN109}, R-function or the theory of mean potential value fields \cite{RN108}. The second part can be constructed in several ways, such as analytical construction \cite{RN99}, an extra neural network \cite{RN105},  transformation using the property of R-functions \cite{RN108}, and so on. 

Another problem in PINN-based methods is the treatment of discontinuity. In the present work, we focus on the weak discontinuity in heterogeneous problems, where the derivatives of the field will be discontinuous across the interface, although the field itself is continuous. In both DCM and DRM, the unknown field is approximated using a single network, and the continuity of the trail function is important to accurately evaluate the differential operators in PDEs. However, for the networks with $\tanh()$ or other high-order continuous functions as activation functions, the derivatives are usually unique everywhere. Hence DCM and DRM will be inaccurate on the interfaces in heterogeneous problems due to the inherent nature of the network. An effective idea is domain decomposition, where each domain will be assigned with a network that couples with adjacent ones on the interface. Such strategies are implemented to DCM by Jagtap et.al. \cite{RN53} and extended to DRM by Wang et.al \cite{RN115}.


Beyond the original statement and the weak statement of WRM, the inverse statement and the derived boundary integral equations (BIEs) are also important in the history of computational mechanics \cite{RN55}. BIE-based methods such as the boundary element method (BEM) have many superiorities such as dimension reduction, easy treatment for infinite/semi-infinite regions, and automatic implementation of the boundary conditions. The BIE-based methods have been widely employed in potential theory, elastostatics, acoustic wave scattering, electromagnetism, and so on \cite{RN55,RN90}. Some recent works are combining BEM with deep neural networks to solve inverse problems based on data-driven techniques, where BEM is implemented only to generate the data set, such as the research by Han et.al  \cite{Han_1} and the research by the author's group \cite{Sun_1}. However, to the best of our knowledge, there lack work that implements PINN based on BIEs. 

In this article, we proposed the boundary-integral type neural networks (BINN) as an alternative scheme to the Deep Collocation and Deep Ritz method, which solves the PDEs with boundary integral equations based on the concept of PINNs. The boundary integral equations can be derived from the inverse statement of WRM. A well-known problem in BIE-based methods is the evaluation of the singular integrals induced from the singularity of the kernel function. We demonstrated that the common treatments for the singular integrals in traditional BEM are not suitable for BINN, while regularization techniques are suggested to compute the singular integrals. As a demonstration, we implemented BINN to potential problems and elastostatic problems in this article, including the problem with the complex-shaped region, the infinite/semi-infinite region, and heterogeneous materials.
The proposed scheme has the following advantages:
\begin{enumerate}
    \item In BINN, all the unknowns are transferred to the boundary, hence the dimension of the problem is reduced by 1. Only the unknowns on the boundary are approximated by neural networks (or the derivatives of the networks), which leads to fewer integral points and less computational cost.
    
    \item In BINN, the loss function only contains the residual of the boundary integral equations, in which all the boundary conditions have been naturally considered, and the network itself does not require to satisfy the given boundary conditions. Therefore, the proposed scheme is suitable for the complex-shaped region, without the request for any special constructions to the approximation function or penalty terms.
    
    \item BINN is based on the inverse statement of WRM, where the continuity requirements of the approximate function are less than methods based on the original statement such as DCM.
    
    \item BINN is a mesh-free method. The unknowns are approximated with a neural network with high order continuous and strong capacity, hence mesh generation is not required and the proposed method may be suitable for problems with large deformation.
    
    \item As a BIE-based method, BINN can be easily implemented to problems with the infinite or semi-infinite region. Moreover, BINN can easily handle heterogeneous problems with a single network.
\end{enumerate}


The remainder of this article is organized as follows: In section \ref{sec2}, we will briefly recall the original and weak statements of the weighted residual method, and outline the derived two models: the Deep Collocation method and the Deep Ritz method. Then we will propose our basic idea of BINN, which can be derived from the inverse statement of the weighted residual method. In section \ref{sec_3}, we will detail the numerical implementation of BINN. Then we will show some numerical examples for BINN in section \ref{sec_results}, including potential problems and elastostatic problems. Conclusions and discussions are given in section \ref{sec_conclusion}.


\section{Methodology}
\label{sec2}
In this section, we will introduce the main idea of BINN. We will first outline the basics of deep neural networks, then we will give a brief introduction to the two most popular deep neural network-based models, namely the Deep Collocation method (DCM) and Deep Ritz method (DRM), which can be derived from the original statement and the weak statement of the weighted residual method (WRM), respectively. Then we will propose the basic idea of BINN, which can be derived from the inverse statement of WRM. 

\subsection{Deep neural networks}
In the past few years, deep learning has reached great success in the area of computer vision and natural language processing tasks. The universal approximation capabilities of neural networks have also gained much attention to be employed as function approximation machine in solving PDEs. 
\begin{figure}
\centering  
\includegraphics[height=8cm]{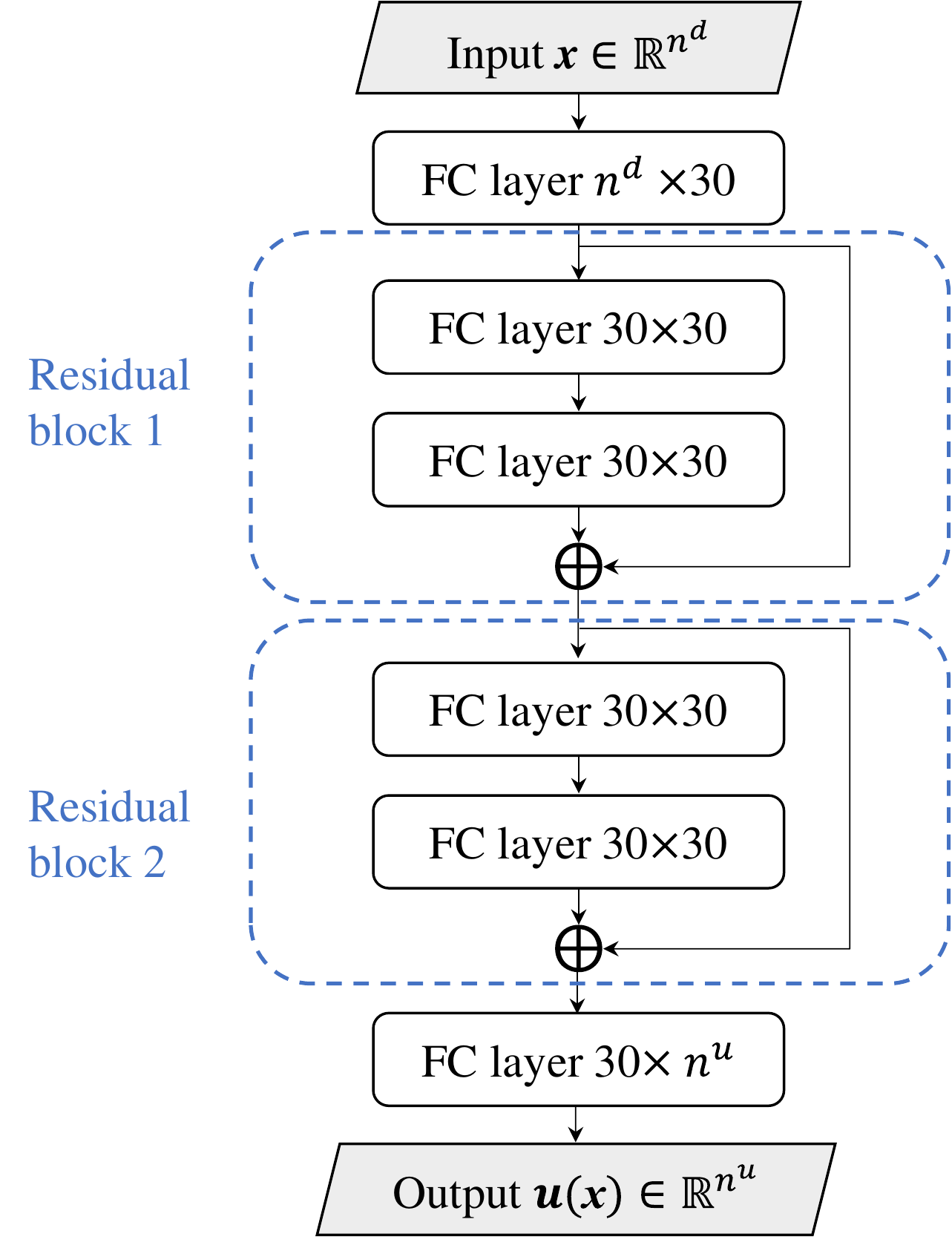}
\caption{The network structure in the present work. Two residual blocks and two extra fully connected layers are employed. }  
\label{NN}  
\end{figure}
In this article, we employed the architecture of full-connected networks with shortcut connections as the approximation function in BINN. The operation of a fully connected layer can be written as:
\begin{equation}
\begin{aligned}
    \boldsymbol{a}^{out} = \boldsymbol{\sigma}\left(\boldsymbol{W}\cdot\boldsymbol{a}^{in}+\boldsymbol{b}\right),
\end{aligned}
\label{FC}
\end{equation}
where $\boldsymbol{a}^{in}$ and $\boldsymbol{a}^{out}$ are the input and output of the layer, respectively. $\boldsymbol{W}$ and $\boldsymbol{b}$ are the weight matrix and bias vector of the layer, respectively.
$\sigma(\cdot)$ is a non-linear function called the activation function. In the algorithms of PINNs, DEM, and BINN, the smoothness of the activation function usually plays a key role to ensure accuracy. In the present work, we adopt the $\tanh()$ function as the activation function:
\begin{equation}
\sigma(x) = \frac{e^{x}-e^{-x}}{e^{x}+e^{-x}}.
\label{actvation}
\end{equation}

A residual block is comprised of several fully connected layers and a shortcut connection. In this article, each block contains two fully connected layers, hence the operation of the residual block can be written as:
\begin{equation}
\boldsymbol{a}^{out} = \boldsymbol{a}^{in}+ \boldsymbol{\sigma}\left(\boldsymbol{W}^2\cdot(\boldsymbol{\sigma}\left(\boldsymbol{W}^1\cdot\boldsymbol{a}^{in}+\boldsymbol{b}^1\right))+\boldsymbol{b}^2\right).
\label{resblock}
\end{equation}

The residual block is also a mature architecture in deep learning to enhance the performance of the model \cite{RN111,RN112,RN97}. As shown in fig.\ref{NN}, the network we used in this article contains two residual blocks and two extra fully connected layers at the beginning and the end of the network, respectively. Like in PINNs and DEM, the networks in BINN serve as function approximation machines. The input of the networks is the coordinate $\boldsymbol{x}\in \mathbb{R}^{n_d}$, where $n_d$ is the dimension of the problem. The output is the field variable $\boldsymbol{u(x)}\in \mathbb{R}^{n_u}$ which could be either scalar or vector, where $n_u$ is the dimension of the output. The network is built with the framework Pytorch \cite{Paszke_1} in this article.

\subsection{Introduction to Deep Collocation method and Deep Ritz method}
\subsubsection{The original and weak statement of the weighted residual method}
In this section, we will briefly recall the original statement and the weak statement of the weighted residual method (WRM) on boundary value problems (BVPs). They are the fundamentals of the DCM and DRM, respectively. Consider a BVP of the general form:
\begin{equation}
 \left\{\begin{aligned}
 \boldsymbol{\mathcal{A}}\left(\boldsymbol{u}(\boldsymbol{x})\right)=\boldsymbol{f}(\boldsymbol{x}),\quad&\boldsymbol{x}\in\Omega,
 \\
 \boldsymbol{\mathcal{B}}\left(\boldsymbol{u}(\boldsymbol{x})\right)=\boldsymbol{g}(\boldsymbol{x}),\quad&\boldsymbol{x}\in\Gamma,
\end{aligned}
\right.    
\label{PDE}
\end{equation}
where $\boldsymbol{u}(\boldsymbol{x})$ denotes the unknown field which may be either scalar or vectorial, $\boldsymbol{\mathcal{A}}(\cdot)$ denotes the differential operator, $\boldsymbol{f}(\boldsymbol{x})$ is the non-homogeneous term. $\boldsymbol{\mathcal{B}}(\cdot)$ denotes the boundary operator, and $\boldsymbol{g}(\boldsymbol{x})$ is the given boundary condition. $\Omega$ and $\Gamma$ denote the interior region and the boundary, respectively. In the present work, we mainly focus on potential problems and elastostatic problems. 
Following the original statement of the weighted residual method, eq(\ref{PDE}) can be approximated as the following form:
\begin{equation}
\left\{
\begin{aligned}
    &\int_{\Omega}\left(\boldsymbol{\mathcal{A}}\left(\boldsymbol{u}(\boldsymbol{x})\right)-\boldsymbol{f}(\boldsymbol{x})\right)\cdot\boldsymbol{w}(\boldsymbol{x})d\Omega=0,\\
    &\int_{\Gamma}\left(\boldsymbol{\mathcal{B}}\left(\boldsymbol{u}(\boldsymbol{x})\right)-\boldsymbol{g}(\boldsymbol{x})\right)\cdot\boldsymbol{\bar{w}}(\boldsymbol{x})d\Gamma=0,\\
\end{aligned}
\right.
\label{WRM}
\end{equation}
where $\boldsymbol{w}(\boldsymbol{x})$ and $\bar{\boldsymbol{w}}(\boldsymbol{x})$ denote the weighting function (also called the test function) on $\Omega$ and $\Gamma$, respectively. Eq(\ref{WRM}) will be equivalent to eq(\ref{PDE}) if it holds for arbitrary $\boldsymbol{w}(\boldsymbol{x})$ and $\bar{\boldsymbol{w}}(\boldsymbol{x})$. In practice, the test function will be chosen from a given function basis. One of the common choices is the Dirac delta function $\Delta(\boldsymbol{x-x}^i)$, where $\{\boldsymbol{x}^i\}$ is a set of collocation points. The derived collocation method can be written as
\begin{equation}
\left\{
\begin{aligned}
    \boldsymbol{\mathcal{A}}\left(\boldsymbol{u} (\boldsymbol{x}^i)\right)-\boldsymbol{f}(\boldsymbol{x}^i)=0, \boldsymbol{x}^i\in\Omega, i=1,2,...,N_{in}, \\
    \boldsymbol{\mathcal{B}}\left(\boldsymbol{u}(\bar{\boldsymbol{x}}^j)\right)-\boldsymbol{g}(\bar{\boldsymbol{x}}^j)=0, \bar{\boldsymbol{x}}^j\in\Gamma, j=1,2,...,N_{bc},
\end{aligned}
\right.
\label{colloc}
\end{equation}
where $N_{in}$ and $N_{bc}$ denote the number of collocation points on $\Omega$ and $\Gamma$, respectively.
 Eq(\ref{colloc}) or eq(\ref{WRM}) requires the evaluation of $\boldsymbol{\mathcal{A}}(\boldsymbol{u} (\boldsymbol{x}))$, which may contain high order derivatives of $\boldsymbol{u}(\boldsymbol{x})$. Therefore, the trail function $\boldsymbol{u}(\boldsymbol{x})$ is expected to have higher order continuity than the test function $\boldsymbol{w}(\boldsymbol{x})$. To reduce the requirement of the continuity, a common treatment is to integrate by parts $\boldsymbol{\mathcal{A}}(\boldsymbol{u}(\boldsymbol{x}))$ and employ the Gauss theorem to get the weak statement:
\begin{equation}
\begin{aligned}
    \int_{\Omega}\left[\boldsymbol{\mathcal{C}}\left(\boldsymbol{u} (\boldsymbol{x})\right)\cdot
    \boldsymbol{\mathcal{D}}\left(\boldsymbol{w} ( \boldsymbol{x})\right)-\boldsymbol{f}( \boldsymbol{x})\cdot\boldsymbol{w}( \boldsymbol{x})\right]d\Omega+\text{b.t.}(\boldsymbol{u,w})=0,
\end{aligned}
\label{weakform}
\end{equation}
where $\boldsymbol{\mathcal{C}}(\cdot),\boldsymbol{\mathcal{D}}(\cdot)$ are differential operators whose order is lower than $\boldsymbol{\mathcal{A}}(\cdot)$, b.t.$(\boldsymbol{u,w})$ denotes the terms of the boundary integrals derived from the Gauss theorem. Moreover, in many physical problems, with some constraints to the space of the trial function, eq(\ref{weakform}) will further lead to an energy form, and can be transferred into a minimization problem of the energy functional $\pi(\boldsymbol{u})$:
\begin{equation}
\boldsymbol{u}^{*} = \mathop{\arg\min}_{\boldsymbol{u}\in \mathcal{U}} \pi(\boldsymbol{u}),
\label{energy_f}
\end{equation}
where $\boldsymbol{u}^{*}$ denotes the approximate solution, $\mathcal{U}$ denotes the space of the trail function. The weak statement reduces the continuity request of the trail function. Moreover, the natural BC can be automatically considered in the energy functional. 

As a demonstration, consider the Poisson equation:
\begin{equation}
\left\{
\begin{aligned}
   -\nabla^{2}u(\boldsymbol{x})&=f(\boldsymbol{x}), &\boldsymbol{x}\in\Omega, \\
    u(\boldsymbol{x})&=\bar{u} (\boldsymbol{x}), &\boldsymbol{x}\in\Gamma_1,\\
    \frac{\partial u(\boldsymbol{x})}{\partial \boldsymbol{n}}&= \bar{q}(\boldsymbol{x}), &\boldsymbol{x}\in\Gamma_2,
\end{aligned}
\right.
\label{poisson}
\end{equation}
where $\Gamma_1$ and $\Gamma_2$ denote the essential and the natural boundary, respectively. The original statement of the WRM can be written as:
\begin{equation}
    \int_{\Omega}\left[ -\nabla^{2}u(\boldsymbol{x})-f(\boldsymbol{x})\right]w(\boldsymbol{x})=0.\\
\label{poi_strong}
\end{equation}
Taking the test function as the Galerkin form $w(\boldsymbol{x})=\delta u(\boldsymbol{x})$, where $\delta$ denotes the variational operator, i.e, the space of the test function is the same as the trail function, integrating by parts the Laplacian and employing the Gauss theorem we will get the weak statement:
\begin{equation}
    \int_{\Omega} \left[ \nabla u(\boldsymbol{x})\cdot\nabla \delta u(\boldsymbol{x})-f(\boldsymbol{x})\delta u(\boldsymbol{x})\right]d\Omega+b.t.(u,\delta u)=0,\\
\label{poi_weak}
\end{equation}
where
\begin{equation}
    b.t.(u,\delta u)= -\int_{\Gamma}\frac{\partial u(\boldsymbol{x})}{\partial \boldsymbol{n}} \delta u d\Gamma.\\
\label{poi_weak2}
\end{equation}

Suppose the space of the trail function $u(\boldsymbol{x})$ is constrained to identically satisfy the essential BC, then we have $\delta u=0$ on $\Gamma_{1}$. Substituting the Neumann boundary conditions in eq(\ref{poisson}) to eq(\ref{poi_weak}) we get the stationary problem of the functional:
\begin{equation}
    \delta \pi(u)=0,
\label{poi_functional}
\end{equation}
where
\begin{equation}
    \pi(u)=\int_{\Omega} \left[\frac{1}{2}\nabla u(\boldsymbol{x})\cdot\nabla u(\boldsymbol{x})-f(\boldsymbol{x})u(\boldsymbol{x})\right]d\Omega-\int_{\Gamma_2} \bar{q}(\boldsymbol{x}) u d\Gamma.\\
\label{poi_functional2}
\end{equation}

Note that the Neumann BC has been naturally considered in the energy functional. Then the stationary problem eq(\ref{poi_functional}) is exactly a minimization problem, which can be easily proved by verifying the sign of the second variation $\delta^2\pi(u)$.

\subsubsection{The Deep Collocation method and Deep Ritz method}
 The DCM can be derived from the original statement, which is employed in the original literature of PINNs \cite{RN29}.
The DCM can be directly formulated from eq(\ref{colloc}) by replacing $\boldsymbol{u}(\boldsymbol{x})$ with a deep neural network $\boldsymbol{u}(\boldsymbol{x})\approx \boldsymbol{\phi}(\boldsymbol{x};\boldsymbol{\theta})$. And the loss function can be directly taken as the sum of the residuals on the collocation points \cite{RN29}:
\begin{equation}
\begin{aligned}
    &\textbf{MSE}= \textbf{MSE}_{in}+ \textbf{MSE}_{bc},\\
    &\textbf{MSE}_{in}=\frac{1}{N_{in}}\sum^{N_{in}}_{i=1}\left\|\boldsymbol{\mathcal{A}}\left(\boldsymbol{\phi}(\boldsymbol{x}^i;\boldsymbol{\theta})\right)-\boldsymbol{f}(\boldsymbol{x}^i)\right\|^2,\\
    &\textbf{MSE}_{bc}=\frac{1}{N_{bc}}\sum^{N_{bc}}_{i=1}\left\|\boldsymbol{\mathcal{B}}\left(\boldsymbol{\phi}( \boldsymbol{x}^i;\boldsymbol{\theta})\right)-\boldsymbol{g}(\boldsymbol{x}^i)\right\|^2,\\
\end{aligned}
\label{mse}
\end{equation}
then the approximate solution will be obtained by minimizing the loss function:
\begin{equation}
    \boldsymbol{\theta}=\mathop{\arg\min} _{\boldsymbol{\theta} \in \boldsymbol{\Theta}}\textbf{MSE},\\
\label{opt_colloc}
\end{equation} 
where $\boldsymbol{\Theta}$ denotes the parameter space of $\boldsymbol{\theta}$. The DCM is a powerful method that can be implemented in any BVPs. If the BVP has an energy form, the DRM can be applied similarly by substituting the approximation $\boldsymbol{u} (\boldsymbol{x})\approx \boldsymbol{\phi} ( \boldsymbol{x};\boldsymbol{\theta} )$ into eq(\ref{weakform}), and the unknowns will be solved through a training process by minimizing the energy functional eq(\ref{energy_f}). For example, in the Poisson equations, DRM can be formulated by replacing the trail function with neural networks $\boldsymbol{\phi}(\boldsymbol{x};\boldsymbol{\theta})$ in eq(\ref{poi_weak}). Then the networks are trained to minimize the functional to produce the approximate solution:
\begin{equation}
    \boldsymbol{\theta}=\mathop{\arg\min} _{\boldsymbol{\theta} \in \boldsymbol{\Theta}}\pi(\boldsymbol{\phi}(\boldsymbol{x};\boldsymbol{\theta})).\\
\end{equation}

An advantage of DRM is that the natural boundary conditions are automatically considered in the functional eq(\ref{poi_functional2}), hence fewer artificial parameters are required. 
Remember that in eq(\ref{poi_functional2}) we assumed that the trail function $\boldsymbol{u}$ has been constrained to satisfy the essential BC. One way is to employ the penalty method and consider the modified functional \cite{RN97}
\begin{equation}
    \pi^{\ast}(\boldsymbol{\phi}(\boldsymbol{x};\boldsymbol{\theta}))=\pi(\boldsymbol{\phi}(\boldsymbol{x};\boldsymbol{\theta}))+\beta \int_{\Gamma_2}\|\boldsymbol{\phi}(\boldsymbol{x};\boldsymbol{\theta})-\bar{\boldsymbol{u}}(\boldsymbol{x})\|^2d\Gamma,\\
\label{Ritz_loss}
\end{equation}
where $\beta$ is the penalty factor, $\Gamma_2$ denotes the essential boundary and $\bar{u}(\boldsymbol{x})$ is the given essential boundary condition. Another strategy is directly imposing the boundary conditions into the trail function with some special construction, which can be implemented in both DCM and DRM. For example, the essential boundary $\boldsymbol{u}\vert _{x=x_0}=0$ can be imposed with the trail function of the form $\boldsymbol{u}=(x-x_0)\boldsymbol{\phi}(\boldsymbol{x};\boldsymbol{\theta})$. For more complex geometries, several researchers have discussed how to construct the form of trail function \cite{RN99,RN101,RN106,RN105,RN108,RN109}. If the boundary conditions have been imposed in advance, the corresponding boundary terms will not appear in the loss function. 

\subsection{The boundary-integral type neural networks (BINN)}
In both DCM and DRM, the neural networks are requested to approximate the interested field on the whole domain $\Omega$, and satisfy all the given boundary conditions whether through a penalty term in the loss function or exact imposition in the trail function. In this section, we will introduce the boundary-integral type neural networks (BINN) as a more efficient strategy, where the networks (or the derivatives of the networks) are only required to approximate the boundary values instead of the whole field. Moreover, the given boundary conditions can be naturally considered and there is no request for the network itself to fit the given boundary conditions in BINN. Therefore, only the unknowns on the boundary should be approximated.
\subsubsection{The inverse statement of weighted residual method}
In the previous section, we outlined the weak statement eq(\ref{weakform}), which can be derived from the original statement by integrating by parts the differential operator and employing the Gauss theorem. In a wide range of PDEs, such as the Poisson equations, Navier equations, Helmholtz equations and so on, if we keep integrating by parts the operator $\boldsymbol{\mathcal{C}}(\cdot)$ in eq(\ref{weakform}) until the derivatives of $\boldsymbol{u}$ vanishes, we will get the inverse statement \cite{RN55}:
\begin{equation}
    \int_{\Omega}\left[\boldsymbol{u}(\boldsymbol{x})\cdot\boldsymbol{\mathcal{E}}\left(\boldsymbol{w}(\boldsymbol{x})\right)-\boldsymbol{f}(\boldsymbol{x})\cdot\boldsymbol{w}(\boldsymbol{x})\right]d\Omega+\text{b.t.}(\boldsymbol{u,w})=0,
\label{inverse}
\end{equation}
where $\boldsymbol{\mathcal{E}}(\cdot)$ is the differential operator with the same order of $\boldsymbol{\mathcal{A}}(\cdot)$. $\text{b.t.}(\boldsymbol{u,w})$ contains all the boundary terms derived from the Gauss theorem, including the value and derivatives of $\boldsymbol{u}$ and $\boldsymbol{w}$ on the boundary. It can be seen that all the derivatives have been transformed to the test function $\boldsymbol{w}$. Eq(\ref{inverse}) can be transferred into a much more elegant form where the unknowns are only on the boundary and the boundary conditions are naturally considered, by taking the test function as the fundamental solution, i.e., the solution of:
\begin{equation}
   \boldsymbol{\mathcal{E}}(\boldsymbol{w}(\boldsymbol{x}))=\Delta (\boldsymbol{x-y})\boldsymbol{e}_i,\quad i=1,2,\dots,n^d,
\label{fund}
\end{equation}
where $\Delta(\cdot)$ denotes the Dirac delta function. $\boldsymbol{e}_i$ is the unit vector along the $i-$th direction. $n^d$ is the dimension of the problem. $\boldsymbol{y}$ is a chosen point called the source point. As a distinction, we will use $\boldsymbol{u}^{s}(\boldsymbol{x;y})$ to specify all the fundamental solutions in eq(\ref{fund}), which is a $n^d\times n^d$ tensor function where $u_{ij}^{s}(\boldsymbol{x;y})$ denotes the $j-$th component of the solution for $\boldsymbol{e}_i$. Let $\boldsymbol{y}\in \Omega$, substituting eq(\ref{fund}) into eq(\ref{inverse}) and noting the property of the Dirac delta function will produce:
\begin{equation}
    \boldsymbol{u}(\boldsymbol{y})=\int_{\Omega}\left[\boldsymbol{u}^{s}(\boldsymbol{x;y})\cdot\boldsymbol{f(x)}\right]d\Omega-\text{b.t.}(\boldsymbol{u,u}^s),\quad \boldsymbol{y}\in \Omega.
\label{eq_INNER}
\end{equation}

In eq(\ref{eq_INNER}), the first term in the right-hand side is a domain integral, where the integrand is composed of the non-homogeneous term $\boldsymbol{f}(\boldsymbol{x})$ and the fundamental solution $\boldsymbol{u}^{s}(\boldsymbol{x;y})$, both of which are given functions. The unknowns are involved in the second term that only contains the boundary integrals. Eq(\ref{eq_INNER}) indicates that once we obtain all the boundary results in $\text{b.t.}(\boldsymbol{u,u}^s)$, the value of any interior point $\boldsymbol{y}\in \Omega$ can be directly calculated. Then our main goal is transferred to solve all the boundary unknowns, including the value and derivatives of $\boldsymbol{u}$. Thus the dimension of the problem is reduced by 1, which is one of the major advantages of the inverse statement. 

In eq(\ref{eq_INNER}), let $\boldsymbol{y}\rightarrow \Gamma$ and we will get the well-known boundary integral equations (BIEs):
\begin{equation}
    \boldsymbol{C}(\boldsymbol{y})\cdot\boldsymbol{u}(\boldsymbol{y})+\int_{\Omega}\left[-\boldsymbol{u}^{s}(\boldsymbol{x;y})\cdot\boldsymbol{f}(\boldsymbol{x})\right]d\Omega+\text{b.t.}(\boldsymbol{u,u}^s)=0,\quad \boldsymbol{y}\in \Gamma,
\label{eq_BIE}
\end{equation}
where $\boldsymbol{C}(\boldsymbol{y})$ is a diagonal matrix that depends on the smoothness of the boundary at $\boldsymbol{y}$. Again we emphasize that the integrand of the domain integral in eq(\ref{eq_BIE}) given function, and the unknowns are only involved in $\boldsymbol{C}(\boldsymbol{y})\cdot\boldsymbol{u}(\boldsymbol{y})$ and the boundary term $\text{b.t.}(\boldsymbol{u,u}^s)$. Similar to the weak statement, where the natural BC can be naturally considered, in BINN, all the boundary conditions have been naturally considered in $\text{b.t.}(\boldsymbol{u,u}^s)$. 

Again we take the Poisson equation as a demonstration. We will start from the weighted residual form eq(\ref{poi_strong}). 
If we integrate by part the Laplacian twice, we will get the inverse statement:
\begin{equation}
    \int_{\Omega} \left[-u(\boldsymbol{x})\nabla ^{2}  w( \boldsymbol{x})-f( \boldsymbol{x})w( \boldsymbol{x})\right]d\Omega+b.t.(u,w)=0,
\label{poi_inv}
\end{equation}
where
\begin{equation}
    b.t.(u,w)=\int_{\Gamma}\frac{\partial w(\boldsymbol{x})}{\partial \boldsymbol{n}}u(\boldsymbol{x})d\Gamma-\int_{\Gamma}\frac{\partial u(\boldsymbol{x})}{\partial \boldsymbol{n}}w(\boldsymbol{x})d\Gamma.
\label{poi_inv2}
\end{equation}

Substituting eq(\ref{poi_inv2}) and the boundary conditions in eq(\ref{poisson}) into eq(\ref{poi_inv}) will produce:
\begin{equation}
\begin{aligned}
    &\int_{\Omega} \left[-u(\boldsymbol{x})\nabla ^{2} w(\boldsymbol{x})-f(\boldsymbol{x})w(\boldsymbol{x})\right]d\Omega
    +\int_{\Gamma_1}\frac{\partial w(\boldsymbol{x})}{\partial n}\bar{u}(\boldsymbol{x})d\Gamma
    +\int_{\Gamma_2}\frac{\partial w(\boldsymbol{x})}{\partial  \boldsymbol{n}}u(\boldsymbol{x})d\Gamma\\
    &-\int_{\Gamma_1}\frac{\partial u(\boldsymbol{x})}{\partial \boldsymbol{n}}w(\boldsymbol{x})d\Gamma-\int_{\Gamma_2}\bar{q}w(\boldsymbol{x})d\Gamma=0.
\end{aligned}
\label{poi_inv3}
\end{equation}

Note that all the boundary conditions are considered in eq(\ref{poi_inv3}). The fundamental solution can be obtained by solving:
\begin{equation}
    \nabla^{2}w(\boldsymbol{x})=\Delta(\boldsymbol{x-y}).
\label{poi_weight}
\end{equation}

For 2D problem, the solution of eq(\ref{poi_weight}) is:
\begin{equation}
    u^{s}(\boldsymbol{x;y})=-\frac{1}{2\pi}\ln \left|\boldsymbol{r }\right|,
\label{poi_fund1}
\end{equation}
where $\boldsymbol{r}=\boldsymbol{x-y}$, and we have:
\begin{equation}
    \frac{\partial u^{s}(\boldsymbol{x;y})}{\partial \boldsymbol{n}}=-\frac{1}{2\pi}\frac{\boldsymbol{r\cdot n}}{|\boldsymbol{r}|^2}.
\label{poi_fund2}
\end{equation}
Substitute eq(\ref{poi_weight}) into eq(\ref{poi_inv3}) and let $\boldsymbol{y}\rightarrow \Gamma$ then we will get the boundary integral equations:
\begin{equation}
\begin{aligned}
    &c(\boldsymbol{y})u(\boldsymbol{y})
    +\int_{\Gamma_1}\frac{\partial u(\boldsymbol{x})}{\partial \boldsymbol{n}}u^{s}(\boldsymbol{x;y})d\Gamma
    -\int_{\Gamma_2}\frac{\partial u^{s}(\boldsymbol{x;y})}{\partial \boldsymbol{n}}u(\boldsymbol{x})d\Gamma=\\
    &\int_{\Gamma_1}\frac{\partial u^{s}(\boldsymbol{x;y})}{\partial \boldsymbol{n}}\bar{u}(\boldsymbol{x})d\Gamma
    -\int_{\Gamma_2}\bar{q}(\boldsymbol{x})u^{s}(\boldsymbol{x;y})d\Gamma
    -\int_{\Omega} f(\boldsymbol{x})u^{s}(\boldsymbol{x;y})d\Omega,\quad \boldsymbol{y}\in \Gamma,
\end{aligned}
\label{poi_bie}
\end{equation}
where $c(\boldsymbol{y})$ is a parameter that depends on the continuity of the boundary on $\boldsymbol{y}$. For smooth boundary, we have $c(\boldsymbol{y})=0.5$. In eq(\ref{poi_bie}) we have re-arranged the terms such that all the unknowns are on the left-hand side. It can be seen that only boundary terms are included on the left-hand side, while the right-hand side contains all the given boundary conditions $\bar{u},\bar{q}$ and the non-homogeneous term $f$.

Note that the BIE formula eq(\ref{eq_BIE}) or eq(\ref{poi_bie}) is still valid for problems on exterior region, i.e., the problems on infinite or semi-infinite region, under some constraints to the field properties at infinity. The detailed derivation can be found in many monographs of BEM \cite{RN55,RN90}. In these problems, the scale can be greatly reduced with the BIE formula, since only boundary values are concerned and we do not have to model the infinite region. The convenient treatment for problems on infinite/semi-infinite region is also an advantage of BIE-based methods.

\subsubsection{Basic idea of BINN}
\begin{figure}
\centering  
\subfigure[]{   
\centering    
\includegraphics[width=12cm]{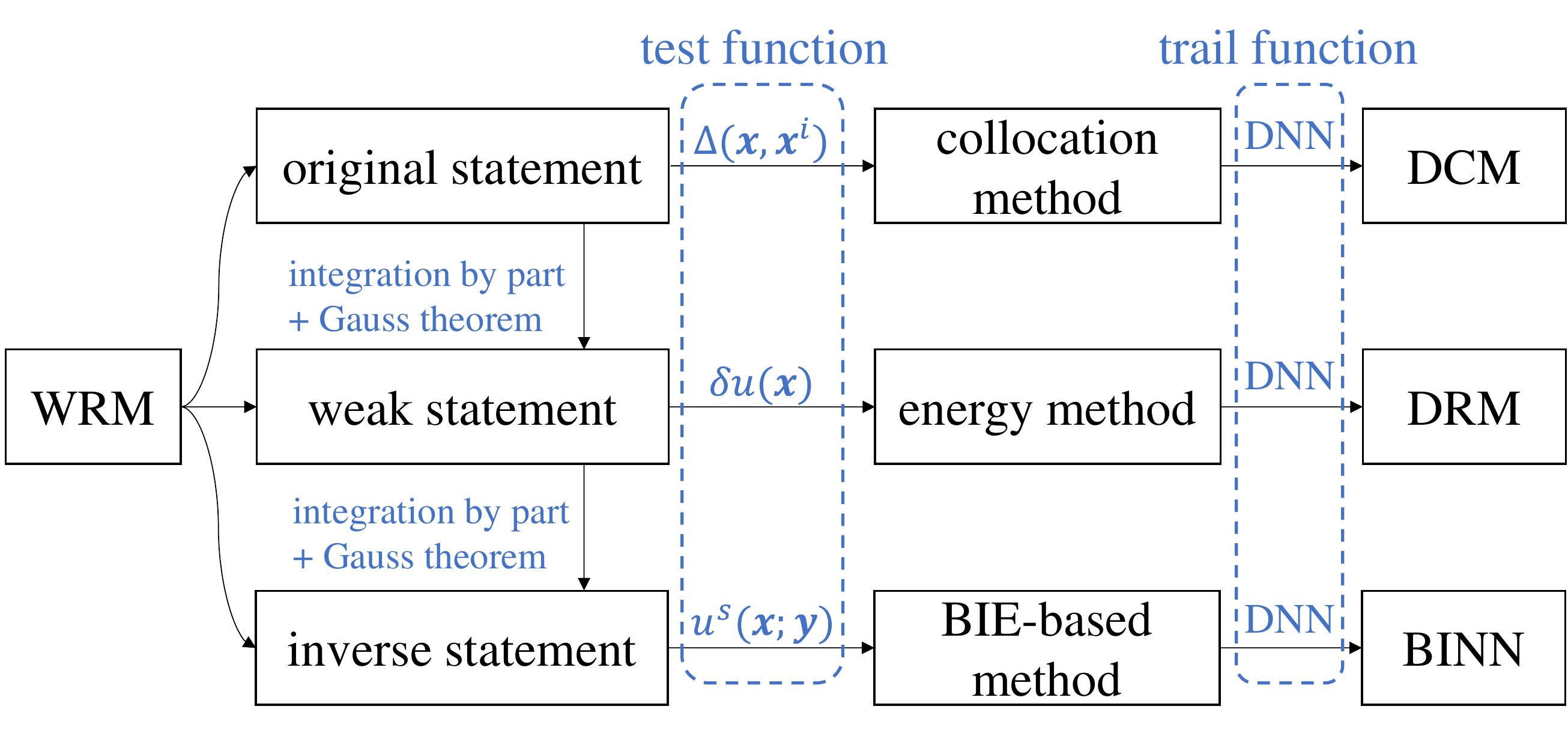}  

}\\

\subfigure[]{
\centering  
\includegraphics[width=10cm]{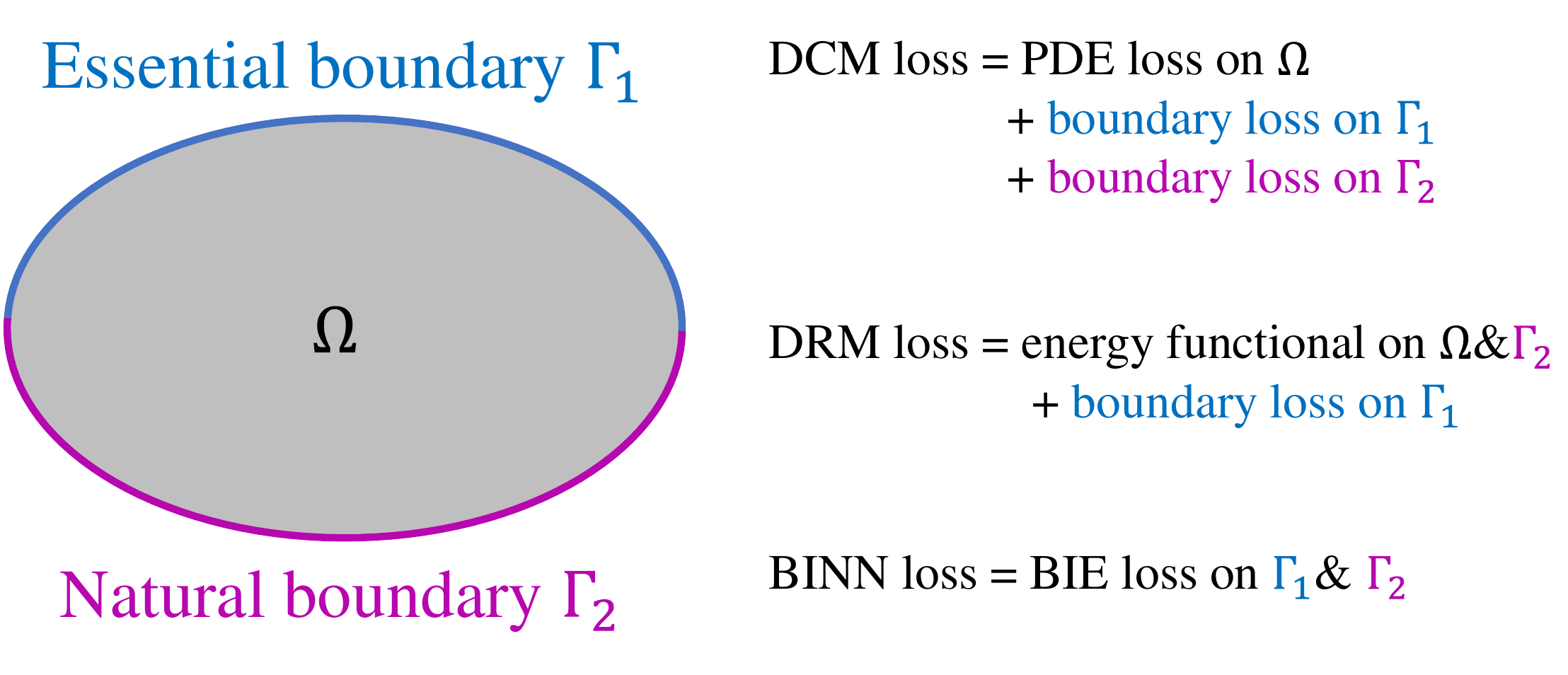}
}

\caption{(a) The relation and the difference of DCM, DRM, and BINN. All of them can be derived from WRM. The test function of them are Dirac delta function $\Delta(\boldsymbol{x}-\boldsymbol{x}^i)$, variation of the trail function $\delta(u(\boldsymbol{x}))$, the fundamental solution $u^s(\boldsymbol{x};\boldsymbol{y})$, respectively. Deep neural networks (DNNs) are involved in the trail function for all three methods. (b) The comparison of the loss function for the three methods. Poisson equations are taken as a demonstration.}  
\label{compara_w}  
\end{figure}
Next, we will illustrate the basic idea of BINN. Unlike DCM or DRM, BINN is based on the inverse statement and the resulting BIE formula eq(\ref{eq_BIE}). Fig.\ref{compara_w}(a) lists the relation and comparison of DCM, DRM, and BINN. A neural network $\boldsymbol{\phi}(\boldsymbol{x;\theta})$ will still be employed to represent the unknowns $\boldsymbol{u}(\boldsymbol{x})$. The difference is that as a BIE-based method, all the interior unknowns have been eliminated and only boundary unknowns are approximated. This is significantly distinct from other PINN-based methods. 


Let us take the Poisson equation as an example. The resulting BIE formula has been shown in eq(\ref{poi_bie}). The unknowns on the boundary with Dirichlet BC (denoted by $\Gamma_1$) and Neumann BC (denoted by $\Gamma_2$) are $\partial u/\partial \boldsymbol{n}$ and $u$, respectively. Then we have the following approximation:
\begin{equation}
\begin{aligned}
&\frac{\partial u(\boldsymbol{x})}{\partial\boldsymbol{n}}\approx\frac{\partial \phi(\boldsymbol{x;\theta})}{\partial\boldsymbol{n}},\quad &\boldsymbol{x}\in\Gamma_1,\\
&u(\boldsymbol{x})\approx \phi(\boldsymbol{x;\theta}),\quad &\boldsymbol{x}\in\Gamma_2.
\end{aligned}
\label{eq_app_BINN_poi}
\end{equation}

Note that only the unknowns are approximated, and the network itself does not have to satisfy the given boundary conditions, i.e., generally we have $\phi(\boldsymbol{x;\theta})\neq \bar{u}(\boldsymbol{x})$ on $\Gamma_1$ and $\partial\phi(\boldsymbol{x;\theta})/\partial \boldsymbol{n}\neq \bar{q}(\boldsymbol{x})$ on $\Gamma_2$. The given boundary conditions can be directly substituted into the BIE formula eq(\ref{poi_bie}).
From this perspective, BINN is an ``economy" strategy.

Substituting eq(\ref{eq_app_BINN_poi}) and all the boundary conditions into eq(\ref{poi_bie}) we can calculate the residual for a given source point $\boldsymbol{y}$:
\begin{equation}
\begin{aligned}
    R(\boldsymbol{y;\theta})=&c(\boldsymbol{y})\hat{u}(\boldsymbol{y})
    +\int_{\Gamma_1}\frac{\partial \phi(\boldsymbol{x;\theta})}{\partial \boldsymbol{n}}u^{s}(\boldsymbol{x;y})d\Gamma
    -\int_{\Gamma_2}\frac{\partial u^{s}(\boldsymbol{x;y})}{\partial \boldsymbol{n}}\phi(\boldsymbol{x;\theta})d\Gamma\\
    & -\int_{\Gamma_1}\frac{\partial u^{s}(\boldsymbol{x;y})}{\partial \boldsymbol{n}}\bar{u}(\boldsymbol{x})d\Gamma
    +\int_{\Gamma_2}\bar{q}(\boldsymbol{x})u^{s}(\boldsymbol{x;y})d\Gamma
    +\int_{\Omega} f(\boldsymbol{x})u^{s}(\boldsymbol{x;y})d\Omega,\quad \boldsymbol{y}\in \Gamma,
\end{aligned}
\label{poi_BINN}
\end{equation}
where 
\begin{equation}
    \hat{u}(\boldsymbol{y})= \left\{  \begin{matrix}\bar{u}(\boldsymbol{y}), &\boldsymbol{y}\in \Gamma_1,\\ \phi(\boldsymbol{y;\theta}), &\boldsymbol{y}\in \Gamma_2.
\end{matrix}
\right.
\end{equation}

Note that in eq(\ref{poi_BINN}) the network $\phi(\boldsymbol{x;\theta})$ is only involved in the first three terms in the right-hand side. The normal derivative $\partial\phi(\boldsymbol{x;\theta})/\partial\boldsymbol{n}$ is calculated with the automatic differentiation techniques \cite{baydin_1}. Suppose $N_s$ source points are allocated on the boundary, then the loss function is taken as the average of the residuals among the source points $\{\boldsymbol{y}^i\}$:
\begin{equation}
\begin{aligned}
    L^{bie}(\boldsymbol{\theta}) = \frac{1}{N_s}\sum^{N_{s}}_{i=1}\left\| R(\boldsymbol{y}^i;\boldsymbol{\theta})\right\|^2.
\end{aligned}
\label{poi_loss}
\end{equation}
The boundary unknowns can be obtained by minimizing the loss function through the training process:
\begin{equation}
    \boldsymbol{\theta}^* = \mathop{\arg\min}_{\boldsymbol{\theta}\in\boldsymbol{\Theta}}L^{bie}(\boldsymbol{\theta}).
\label{poi_minimam}
\end{equation}

Comparing the BINN loss in eq(\ref{poi_loss}) with DCM loss in eq(\ref{mse}) and DRM loss in eq(\ref{Ritz_loss}), it can be observed that BINN loss does not contain any extra term to impose the boundary conditions. The loss function only contains the residuals of the BIE formula, where all the boundary conditions have been naturally considered. The comparison of the loss function for Poisson equations is shown in fig.\ref{compara_w}(b).

Once we have solved the boundary unknowns, the value of any interior point $\boldsymbol{y}\in\Omega$ can be directly obtained through the integral:
\begin{equation}
\begin{aligned}
    &u(\boldsymbol{y})=
    -\int_{\Gamma_1}\frac{\partial \phi(\boldsymbol{x;\theta})}{\partial \boldsymbol{n}}u^{s}(\boldsymbol{x;y})d\Gamma
    +\int_{\Gamma_2}\frac{\partial u^{s}(\boldsymbol{x;y})}{\partial \boldsymbol{n}}\phi(\boldsymbol{x;\theta})d\Gamma
    +\int_{\Gamma_1}\frac{\partial u^{s}(\boldsymbol{x;y})}{\partial \boldsymbol{n}}\bar{u}(\boldsymbol{x})d\Gamma\\
    &-\int_{\Gamma_2}\bar{q}u^{s}(\boldsymbol{x;y})d\Gamma
    -\int_{\Omega} f(\boldsymbol{x})u^{s}(\boldsymbol{x;y})d\Omega,\quad \boldsymbol{y}\in \Omega.
\end{aligned}
\label{poi_d_inner}
\end{equation}

Eq(\ref{poi_d_inner}) can be directly derived by substituting eq(\ref{poi_weight}) to eq(\ref{poi_inv3}) and employing the property of the Dirac delta function.

Consider the general BIE in eq(\ref{eq_BIE}). The term $b.t.(\boldsymbol{u,u}^s)$ is comprised of the boundary integrals involving $u$ and its derivatives. A part of them are given as boundary conditions, and the others are unknowns to be solved.
Suppose the boundary $\Gamma$ is comprised of several parts $\Gamma=\Gamma_1\cap\Gamma_2\dots\cap\Gamma_n$, $\Gamma_i\cup\Gamma_j=\emptyset$ if $i\neq j$, and the unknowns on $\Gamma_i$ are $\mathcal{B}^i(\boldsymbol{u}(\boldsymbol{x}))$, where $\mathcal{B}^i$ is a boundary operator such as the normal derivatives $\partial/\partial\boldsymbol{n}$, the identity mapping, or any other forms such as the surface traction in elastostatic problems. We will apply the following approximation:
\begin{equation}
\mathcal{B}^i(\boldsymbol{u}(\boldsymbol{x}))\approx \mathcal{B}^i(\boldsymbol{\phi}(\boldsymbol{x;\theta})),\quad \boldsymbol{x}\in \Gamma_i, i=1,2,\cdots,n.
\label{eq_app_BINN}
\end{equation}
The derivatives of the network are calculated through the automatic differentiation techniques. The boundary term $b.t.(\boldsymbol{u,u}^s)$ in eq(\ref{eq_BIE}) can be divided into two parts:
\begin{equation}
b.t.(\boldsymbol{u,u}^s) = \boldsymbol{I}(\boldsymbol{u,u}^s)+\bar{\boldsymbol{I}}(\boldsymbol{u,u}^s),
\label{eq_BINN_bt}
\end{equation}
where $\boldsymbol{I}(\boldsymbol{u,u}^s)$ contains all the unknowns $\mathcal{B}^i(\boldsymbol{u}(\boldsymbol{x}))$, and $\bar{\boldsymbol{I}}(\boldsymbol{u,u}^s)$ contains all the given boundary conditions. 
Substituting eq(\ref{eq_app_BINN}) into $\boldsymbol{I}(\boldsymbol{u,u}^s)$ and the boundary conditions into $\bar{\boldsymbol{I}}(\boldsymbol{u,u}^s)$ we can calculate the residual of eq(\ref{eq_BIE}):
\begin{equation}
\begin{aligned}
    \boldsymbol{R}(\boldsymbol{y;\theta})=&
     \boldsymbol{C}(\boldsymbol{y})\cdot\hat{\boldsymbol{u}}(\boldsymbol{y})+\int_{\Omega}\left[-\boldsymbol{u}^{s}(\boldsymbol{x;y})\cdot\boldsymbol{f(x)}\right]d\Omega+
     \bar{\boldsymbol{I}}(\boldsymbol{u,u}^s) +
     \boldsymbol{I}(\boldsymbol{\phi(\boldsymbol{x;\theta}),u}^s),\quad \boldsymbol{y}\in \Gamma,
\label{eq_res_BINN}
\end{aligned}
\end{equation}
where 
\begin{equation}
    \hat{\boldsymbol{u}}(\boldsymbol{y})= \left\{  \begin{matrix}\bar{\boldsymbol{u}}(\boldsymbol{y}), &\boldsymbol{y}\in \Gamma_u,\\ \boldsymbol{\phi}(\boldsymbol{y;\theta}), &\boldsymbol{y}\in \Gamma_{nu}.
\end{matrix}
\right.
\end{equation}
where $\Gamma_u$ denotes the boundary where $\boldsymbol{u}$ is specified, and $\Gamma_{nu}$ denotes the boundary where $\boldsymbol{u}$ is unknown. Then the loss function can be similarly formulated as eq(\ref{poi_loss}) and the unknowns will be solved with the minimization problem eq(\ref{poi_minimam}). Once we obtain all the boundary unknowns, the value of any point $\boldsymbol{y}$ on the interior region can be calculated following eq(\ref{eq_INNER}):
\begin{equation}
    \boldsymbol{u}(\boldsymbol{y})=\int_{\Omega}\left[\boldsymbol{u}^{s}(\boldsymbol{x;y})\cdot\boldsymbol{f(x)}\right]d\Omega- \bar{\boldsymbol{I}}(\boldsymbol{u,u}^s) -
     \boldsymbol{I}(\boldsymbol{\phi(\boldsymbol{x;\theta}),u}^s),\quad \boldsymbol{y}\in \Omega.
\label{eq_INNER_B}
\end{equation}

In summary, we conclude the algorithm of BINN in Algorithm 1.
\begin{algorithm}
    \caption{Numerical scheme of BINN}
    \label{alg:algorithm-label}
    \begin{algorithmic}[1]
        \State Allocate $N_s$ source points $\{\boldsymbol{y}^i\}$ and $N_f$ integration points $\{\boldsymbol{x}^i\}$ on the boundary. If the non-homogeneous terms are not zero, allocate $N_d$ integration points $\{\boldsymbol{q}_i\}$ inside the domain.
        \State Compute the Jacobian $J$ on all the $\{\boldsymbol{x}^i\}$ and $\{\boldsymbol{q}_i\}$. Compute the outward normal vectors on $\{\boldsymbol{x}^i\}$.
        \State For each $\boldsymbol{y}^i$, compute the value of the kernel function on all the integration points $\{\boldsymbol{x}^i\}$ and $\{\boldsymbol{q}_i\}$.
        \State For each $\boldsymbol{y}^i$, evaluate the integrals $\bar{\boldsymbol{I}}(\boldsymbol{u},\boldsymbol{u}^s)$ in eq(\ref{eq_res_BINN}) with given boundary conditions. If the non-homogeneous terms are not zero, evaluate the domain integrals. 
        \State Initialize the network parameters $\theta$, e.g., with the Xavier initialization \cite{Glorot_1}. Choose the number of the iteration steps $N_{max}$.
        \For{$k=1:N_{max}$}
        \State Compute the unknowns $\mathcal{B}^i(\boldsymbol{\phi}(\boldsymbol{x};\boldsymbol{\theta}))$ in eq(\ref{eq_app_BINN}) on all the boundary integration points ${\boldsymbol{x}^i}$.
        \State For each $\boldsymbol{y}^i$, evaluate the integrals $\boldsymbol{I}(\boldsymbol{\phi(\boldsymbol{x};\boldsymbol{\theta})},\boldsymbol{u}^s)$. Compute $\boldsymbol{\phi}(\boldsymbol{y}^i;\boldsymbol{\theta})$ if $\boldsymbol{y}^i\in \Gamma_{nu}$.
        \State Compute the residuals $\boldsymbol{R}(\boldsymbol{y}^i;\boldsymbol{\theta})$ for all the source point following eq(\ref{eq_res_BINN}), then compute the loss function following eq(\ref{poi_loss}).
        \State Update the parameters $\boldsymbol{\theta}$ with optimization algorithm.
        \EndFor
        \State After the training process, all the boundary unknowns have been solved. Then the interior results can be calculated through eq(\ref{eq_INNER_B}).
    \end{algorithmic}
\end{algorithm}
\subsubsection{Comparison with the boundary element method}
As a BIE-based method, it is necessary to compare BINN with the traditional boundary element method (BEM) to gain better recognition. The comparisons can be summarized as follows:
\begin{enumerate}
\item In BEM, the boundary will be discretized into several elements, and the unknowns will be approximated with piece-wise interpolation functions called shape functions in each element. While BINN is exactly a mesh-free method that a neural network will be employed as the approximate function in all the boundaries. Hence BINN will be more flexible for arbitrary geometry.
\item In the boundaries with essential BC, the boundary unknowns will contain the derivatives of the field function. In BEM, these derivatives will be directly approximated using the shape function, hence the computation of the derivatives is not required. Therefore, the continuity requirement of the shape function is very loose in BEM, and even constant elements can be adopted. While in BINN, the derivatives are computed with automatic differentiation techniques. To ensure the continuity requirement, $\tanh()$ function is chosen as the activation function in the network architecture.
\item Like other traditional methods such as the finite difference method and finite element method, BEM involves the solution to a system of linear equations. The coefficient matrices in BEM are usually dense and asymmetric, hence direct solvers such as Gauss elimination or $\mathbf{LU}$ decomposition are commonly adopted. Some fast algorithms such as the fast multipole method (FMM) and the method based on hierarchical matrices are also developed to solve BEM equations in large-scale problems, where an iteration solver such as the generalized minimum residual method (GMRES) should be applied to solve the linear equations. While in BINN, the solution is obtained through the training process, where gradient descent-based algorithms are employed to decide the parameters $\boldsymbol{\theta}$ in the network.
\item The preference for the treatment of the singular integrals is also different between BEM and BINN. We will detail this issue in section \ref{sec_integrals}
\end{enumerate}

\section{The numerical scheme of BINN}
\label{sec_3}
In this section, we will detail the numerical implementation of BINN. The core of BINN is the evaluation of the boundary integrals in eq(\ref{eq_res_BINN}). In the BIE formulations, the fundamental solution $\boldsymbol{u}^s(\boldsymbol{x};\boldsymbol{y})$ plays a major role to eliminate the unknowns in the interior region. However, $\boldsymbol{u}^s(\boldsymbol{x};\boldsymbol{y})$ is singular when $\boldsymbol{x}\rightarrow \boldsymbol{y}$, which brings the problem of singular integrals. Common integral strategies such as the Gaussian quadrature rule or Monte Carlo method cannot be directly employed since lots of integration points are required to ensure accuracy.
Although the computation of the singular integrals has been well-discussed in traditional boundary element methods, the integration strategy for BINN should be carefully chosen due to its special features.

\subsection{Allocation of the source and integration points}
In this section, we will introduce our strategy for allocating the source and integration points.
Before proceeding into the details, we first explain our principles on how to choose the integration strategy in BINN:
\begin{enumerate}
\item The accuracy of the integrals should be guaranteed, especially the singular integrals whose values might be dominant. This is important for the final accuracy of BINN.

\item Since the loss function in BINN contains all the values on the integration points, the back-propagation during the training process will also involve the derivatives at these points. Too many integration points will lead to huge computational costs. To reduce the number of integration points, we always want to evaluate both the singular and non-singular integrals with the same set of integration points.
\end{enumerate}

In order to evaluate all the singular integrals accurately, sufficient integration points should be allocated around each source point. The commonly used strategy in the PINN-based method is the Monte Carlo method. However, the integration points in the Monte Carlo method are generated randomly, which is hard to control the point distribution and may lead to sparse results in some area. Therefore, we adopt the piece-wise Gaussian quadrature rule in BINN: The boundary are divided into several segments, each segment centered at a source point. And sufficient integration points are allocated in each segment to ensure accuracy. 
For a given source point, the singular integrals are only involved in the segment that contains it, and the integrals on other segments are treated with regular integrals.

\subsection{Evaluation of the integrals}
\label{sec_integrals}
In this section, we will discuss the concept on how to evaluate the integrals in BINN.
In traditional methods such as FEM, the Gaussian quadrature rule is commonly used to compute the integrals due to its high algebraic accuracy. The DRM also involves the evaluation of integrals, where the Monte Carlo method is preferred since it is convenient for mesh-free methods and is efficient to evaluate the integrals in high-dimensional spaces.
As demonstrated before, BINN involves singular integrals, where the direct implementation of the above quadrature rules will usually lead to a large error. 

According to the singularity of the integrand, there are three types of integrals for BINN in the present work: The regular integrals which contain no singularity, the weakly singular integrals, and the strongly singular integrals (the Cauchy-principle value of integrals). The evaluation of these integrals will be detailed in the following sections. For the sake of simplicity, 2D problems are considered in the present work as a demonstration.
\begin{remark}
\label{remark_cond}
The singularity comes from the fundamental solutions $\boldsymbol{u}^s(\boldsymbol{x};\boldsymbol{y})$ when $\boldsymbol{x}\rightarrow \boldsymbol{y}$. Since the
source points are allocated on the boundary, when evaluate the boundary integral, the integrand will be singular near the source point. A natural idea is to move the source points outside the interested domain $\Omega\cup \Gamma$, then we always have $\boldsymbol{x}\neq \boldsymbol{y}$ in the boundary integral, and the singularity is removed. However, the condition of the resulting BIE will be more pathological. For the sake of simplicity, we give a brief analysis in \ref{App_cond}, by taking the Poisson equations as a demonstration.
\end{remark}

\subsubsection{Evaluation of regular integrals}
\label{sec_reg_int}
In BINN, the regular integrals correspond to the most common case. For a given source point, only the integrals on the segment that contains it are singular, otherwise the integrals are regular. For 2D problems, the regular integrals on segment $\Gamma_s$ can be evaluated directly through the Gauss-Legendre quadrature rule:
\begin{equation}
    \int_{\Gamma_s}f(\boldsymbol{x})d\Gamma = \sum_{i=1}^{n_g} f(\boldsymbol{x}(\xi^i))J(\xi^i)w^i,
\end{equation}
where $n_g$ is the order of the quadrature rule, i.e., $n_g$ Gaussian quadrature points are allocated for each segment. $\xi^i$ and $w^i$ denote the Gauss points and the weights of the Gauss-Legendre quadrature rule, respectively. $J = \partial\Gamma/\partial\xi$ is Jacobian of the transformation from the arc length to the local variable $\xi$. In this article, we choose $n_g=10$ in all the examples.

\subsubsection{Evaluation of weakly singular integrals}
\label{sec_sing_int}
 In the present work, the weakly-singular integrals can always be transferred into the following form, regarding that the source point is located at the center of the segment:
\begin{equation}
    \int_{-a}^{a} \ln{|t|}f(t) dt,
\label{eq_weak}
\end{equation}
where $f(t)$ is a regular term contains all the external terms such as the networks $\phi(\boldsymbol{x}(t);\boldsymbol{\theta})$, the Jacobian, etc. The singular term $\ln{|t|}$ comes from the fundamental solution. The weakly singular integral belongs to the improper integral, which is mathematically integrable in the ordinary sense. However, the standard Gaussian quadrature rule cannot be directly employed due to the $O(\ln{(t)})$ singularity of the integrands.

In the traditional boundary element method, weakly singular integrals can be analytically computed for low-order elements such as constant element. However, this is not suitable for BINN due to the complexity of $\phi(\boldsymbol{x}(t);\boldsymbol{\theta})$. Another common strategy is to use the logarithmic Gaussian quadrature formulas:
\begin{equation}
    \int_{0}^{1}\ln{\frac{1}{\eta}}f(\eta)d\eta = \sum_{i=1}^{n_l} f(\eta^i)w^{i'},
\end{equation}
where $n_l$ denotes the order of the quadrature rule, $\eta^i$ and $w^{i'}$ are the Gauss points and the weights of the logarithmic Gaussian quadrature formulas, respectively. However, the integration points in logarithmic Gaussian quadrature formulas $\boldsymbol{x}(\eta^i)$ are not coincide with those in the standard Gauss-Legendre quadrature rule $\boldsymbol{x}(\xi^i)$, which means we have to allocate extra integration points around each source point. As demonstrated before, this will lead to higher computational cost in BINN. In order to evaluate the integrals eq(\ref{eq_weak}) with standard Gaussian quadrature rule, regularization techniques are employed to remove the singularity. In the present work, we adopt the subtraction and addition method, i.e., we first subtract a term from the singular integral to make it regular and easy to be computed numerically, and the subtracted term will be computed analytically and added back on. Such strategy has been well discussed in literature \cite{RN90}. For the integral in eq(\ref{eq_weak}), we rewritten it as:
\begin{equation}
\begin{aligned}
    \int_{-a}^{a} \ln{|t|}f(t) \,dt &= \int_{-a}^{a} \ln{|t|}\left[f(t)-f(0)\right] \,dt+\int_{-a}^{a} \ln{|t|}f(0)\,dt\\
    &= \int_{-a}^{a} \ln{|t|}\left[f(t)-f(0)\right] \,dt+2f(0)(a\ln{a}-a) .
\end{aligned}
\label{eq_weakreg}
\end{equation}
Note that $f(0)$ is exactly the value on the source point. The integrand $ \ln{|t|}\left[f(t)-f(0)\right]$ will be regular if the function $f(t)$ satisfies certain continuity conditions, such as the Lipschitz condition. A brief proof is given in \ref{App_1}.
In BINN, the neural networks with $\tanh()$ as activation function are always differentiable, which is even stronger than the Lipschitz condition. The source points are located at the center of each segment, hence the Jacobian is also smooth enough in the segment. Therefore, the Lipschitz continuity of $f(t)$ can be always guaranteed, and regularization in eq(\ref{eq_weakreg}) is always available. In fact, to derive the regularity of the integral, the Lipschitz continuity is an over-strong constraint. For function $f(t)$ with a worse condition, eq(\ref{eq_weakreg}) may still be available. 

\subsubsection{Evaluation of strongly singular integrals}
In the present work, the strongly singular integrals have the following form:
\begin{equation}
    \int_{-a}^{a} \frac{1}{t}f(t) \,dt,
\label{eq_strong}
\end{equation}
where $f(t)$ is a regular function that contains all the external terms. Unlike the weakly singular integral, the value of eq(\ref{eq_strong}) only exists in the Cauchy principle sense. The singular term $1/t$ comes from the derivative of the fundamental solution. In traditional BEM, such integrals can also be calculated analytically for constant elements, which is not available for BINN. For the general case, a common strategy to evaluate eq(\ref{eq_strong}) is using simple solutions such as rigid body displacement. Roughly speaking, the singular integral in eq(\ref{eq_strong}) along with the coefficient $\boldsymbol{C}(\boldsymbol{y})$ in eq(\ref{eq_BIE}) can be calculated from the summation of other regular integrals over the boundary. However, this is also not available in BINN. Again the regularization techniques are employed. It should be noticed that the subtraction and addition methods are still available for the integrals in eq(\ref{eq_strong}). However, regarding that the source point is always located at the center of the segment in this article, a more convenient choice is the formula \cite{1958On}:
\begin{equation}
    \int_{-1}^{1} \frac{1}{\xi}f(\xi) \,d\xi = \int_{-1}^{1} \frac{1}{\xi}\left[f(\xi)-f(-\xi)\right] \,d\xi.
\label{eq_regstrong}
\end{equation}
Eq(\ref{eq_regstrong}) can be derived by separating the integrand $\frac{1}{\xi}f(\xi)$ into the odd and even part, then the integral of the odd part over $[-1,1]$ become zero. Similarly, it can be proven that the integrand in the left-hand side of eq(\ref{eq_regstrong}) is regularized if the function $f(t)$ satisfies the Lipschitz condition. Then the integral can be evaluated with the standard Gaussian quadrature rule. Eq(\ref{eq_regstrong}) is convenient if the interval is symmetric so that both $\boldsymbol{x}(\xi)$ and $\boldsymbol{x}(-\xi)$ are Gaussian points, which is exactly the case in this article. A drawback for Eq(\ref{eq_regstrong}) is that we can only use the even order Gaussian quadrature rule, where $\xi=0$ is not the abscissa. For the general case, i.e., the source points are not located in the center of the segments, we can still apply the subtraction and addition methods as mentioned in section \ref{sec_sing_int}. 

\subsection{Training strategy}
After the evaluation of all the boundary integrals, the residuals in of BIE eq(\ref{eq_res_BINN}) and loss function in eq(\ref{poi_loss}) can be calculated. Then the boundary unknowns can be obtained by minimizing the loss function. The optimization algorithms have been quite mature in deep learning, most of which are variations of the stochastic gradient descent (SGD) method, which is a first-order algorithm. Second-order algorithms such as the L-BFGS method have also been applied in PINN-based methods \cite{RN9}. All these algorithms have been embedded into modern machine learning frameworks such as Pytorch and TensorFlow. In the present work, the networks are built with the framework Pytorch, and trained with the built-in $Adam$ optimizer \cite{RN70}.

By the way, the batch training strategy has been a mature technique in deep-learning \cite{RN63,RN69}: The training set will be divided into several mini-batches, then the loss function will be evaluated on each mini-batch, and the iterations in the training process are also implemented batch-wisely. The strategy has been also applied in PINN-based methods such as DCM, where the collocation points are divided into mini-batches to implement batch training. The batch training strategy is helpful to reduce over-fitting and save storage cost, especially for large training sets (or large collocation point sets in PINNs). Such a strategy is also available for BINN. Note that the residual of BIE can be calculated individually for each source point, we can divide the source points into mini-batches to implement batch training, which might be helpful in large problems.
In this article, the numbers of source points in all the examples are no more than 400, hence we do not employ the batch training strategy.

\subsection{Evaluation of the field variables on the interior region}
\label{sec_inner}
After the training process, we have solved all the boundary unknowns, and the parameters $\boldsymbol{\theta}$ in the network will be frozen. Then the results on the interior region can be calculated through the eq(\ref{eq_INNER_B}). 
Although the integrals in eq(\ref{eq_INNER_B}) are regular, there is a problem of the evaluation for nearly singular integrals when the interior point $\boldsymbol{y}$ is closed to the boundary. 
In this case, the integrand is not exactly singular but changes dramatically near the source point $\boldsymbol{y}$. Therefore, the standard Gaussian quadrature rule is also not suitable. Such integrals also occur in thin-walled structures like coating.
There has been a lot of research on the evaluation of nearly singular integrals \cite{RN92,RN93,RN91,RN94,RN95,RN96}, such as self-adaptive integral techniques, regularization, or variable transformation. Most of the research is related to the integrals in BEM, and some of them may be still available in BINN. However, in the present work, we find that the nearly singular integrals only occur when the inner points are extremely close to the boundary. For the sake of simplicity, we do not apply extra treatments and just evaluate them with the standard Gaussian quadrature rule. We will give a brief analysis of them in the first example in section \ref{sec_results}, and in the all examples, the inner points are assumed to have a small margin $\varepsilon$ from the boundary. 
by the way, in some special cases such as thin-walled structures, the nearly singular integrals will also occur in the BIE formula eq(\ref{eq_BIE}) and should be treated carefully, which is beyond the scope of this article.

\section{Results}
\label{sec_results}
In this section, we will give some results of the BINN. In the present work, we mainly focus on the potential problems governed by Poisson equations, and the elastostatic problems governed by Navier equations.
\subsection{Potential problems}
In this section, we will investigate the performance of BINN on potential problems governed by the Poisson equation. Note that the non-homogeneous term $f(\boldsymbol{x})$ are only involved in the integral $\int_{\Omega} f(\boldsymbol{x})u^{s}(\boldsymbol{x;y})d\Omega$ in eq(\ref{poi_BINN}). This integral does not involve the trail function and is a constant for a given source point $\boldsymbol{y}$. For the sake of simplicity, we will consider the homogeneous problem with $f(\boldsymbol{x})=0$, i.e., the Laplace equation in the following examples.
\subsubsection{Potential problems on a flower-shaped region}
\begin{figure}
\centering  
\subfigure[]{   
\centering    
\includegraphics[width=3.5cm]{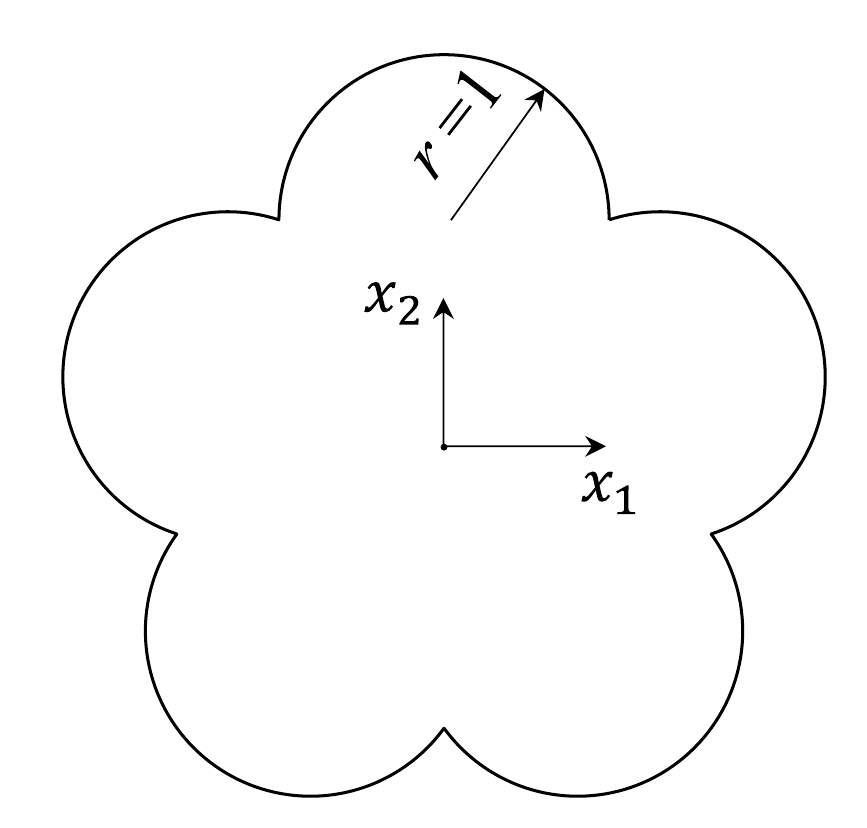}  

}
\quad
\subfigure[]{

\centering  
\includegraphics[width=5.8cm]{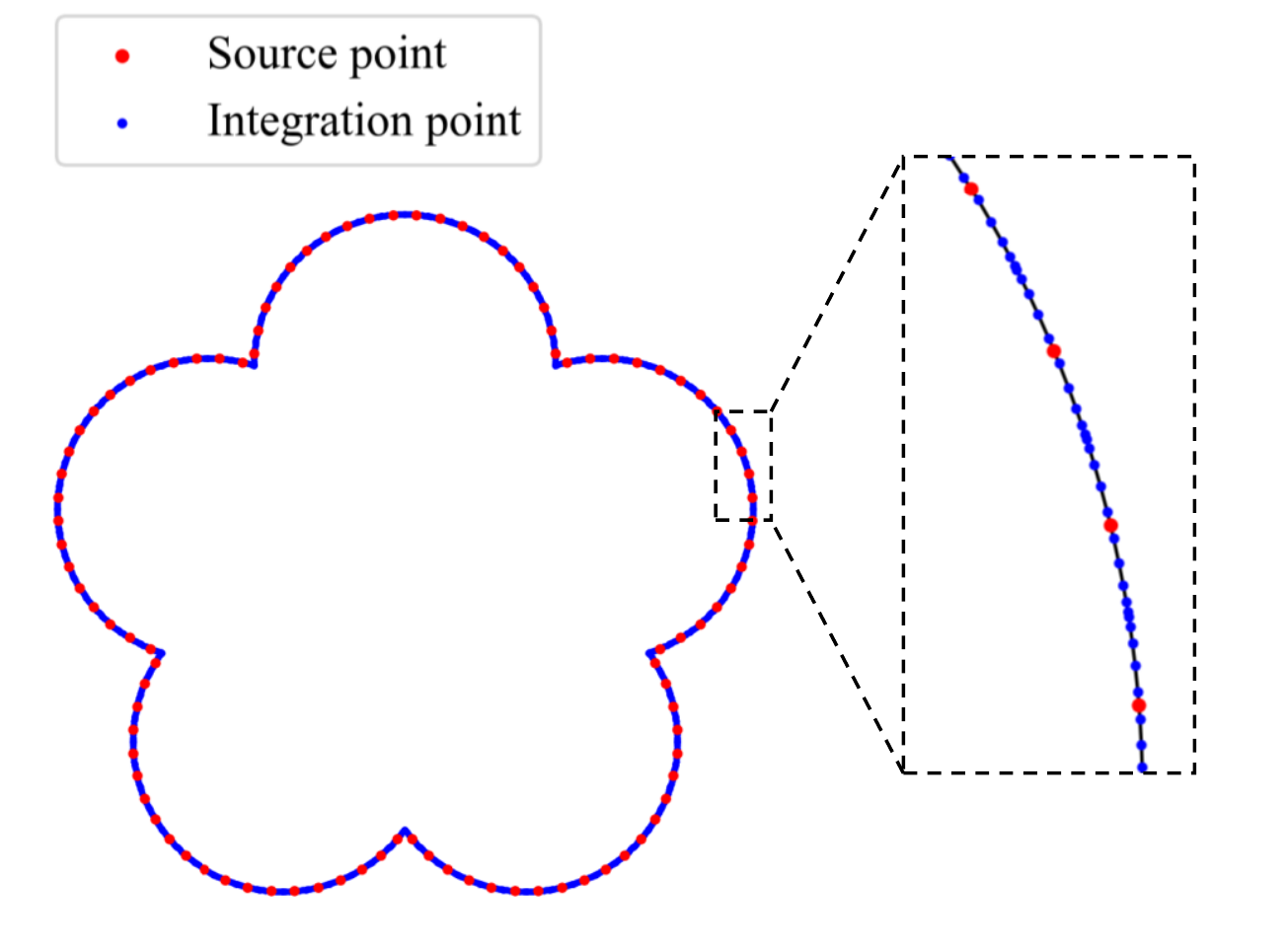}
}
\quad
\subfigure[]{

\centering  
\includegraphics[width=3.5cm]{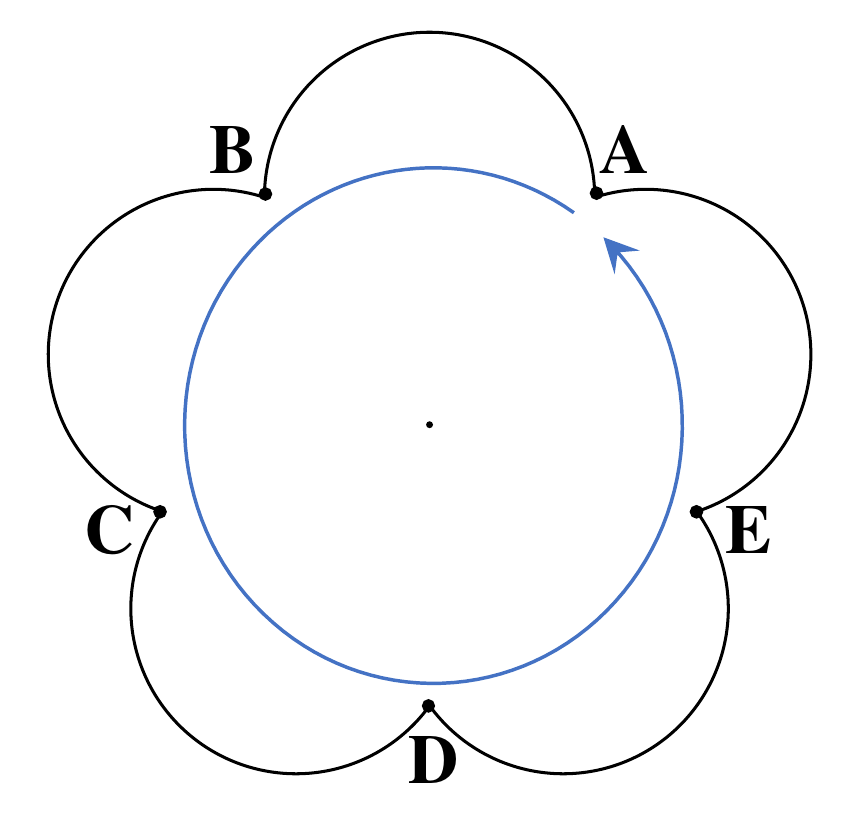}
}
\caption{(a) The geometry of the flower-shaped region. (b) The distribution of the source points and integration points. (c) The trajectory to show the boundary results.}  
\label{geo_flower}  
\end{figure}

To highlight the ability of BINN on complex-shaped geometries, consider the following 2-D Laplace equation on a flower-shaped region with Dirichlet boundary conditions:
\begin{equation}
\left\{ 
    \begin{array}{rlc}
    \nabla^2 u(\boldsymbol{x}) &= 0 &in \quad \Omega,\\
    u(\boldsymbol{x})&=\bar{u}(\boldsymbol{x})& in\quad \Gamma,\\
    \end{array}
\right.
\label{Potential_flower}
\end{equation}
where $\bar{u}(\boldsymbol{x})$ is given by the analytical solution:
\begin{equation}
    u(\boldsymbol{x}) = \sin(x_1)\sinh(x_2)+\cos(x_1)\cosh(x_2).
\label{flower_essentialBC}
\end{equation}

\begin{figure}
\centering  
\subfigure[]{   
\centering    
\includegraphics[height=3.7cm]{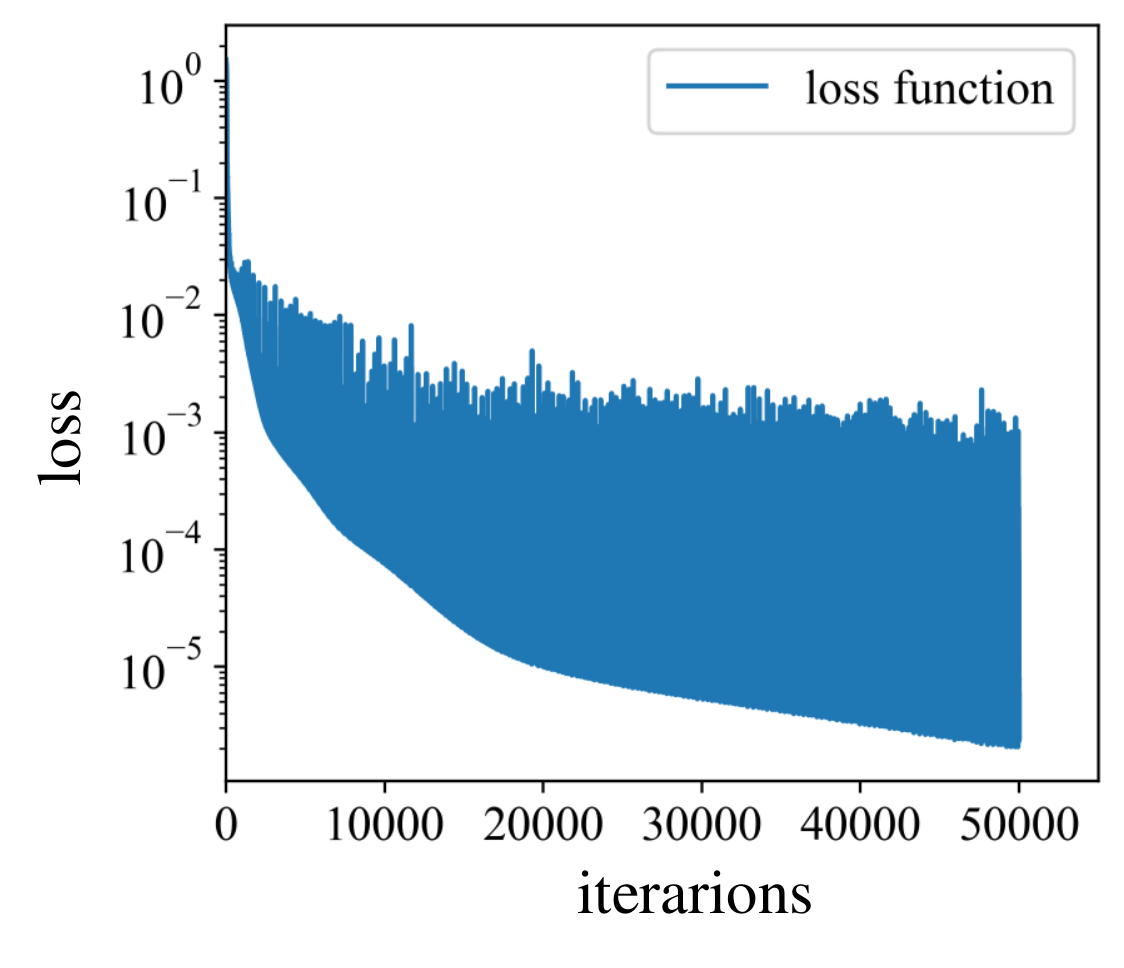}  

}
\quad
\subfigure[]{   
\centering    
\includegraphics[height=3.7cm]{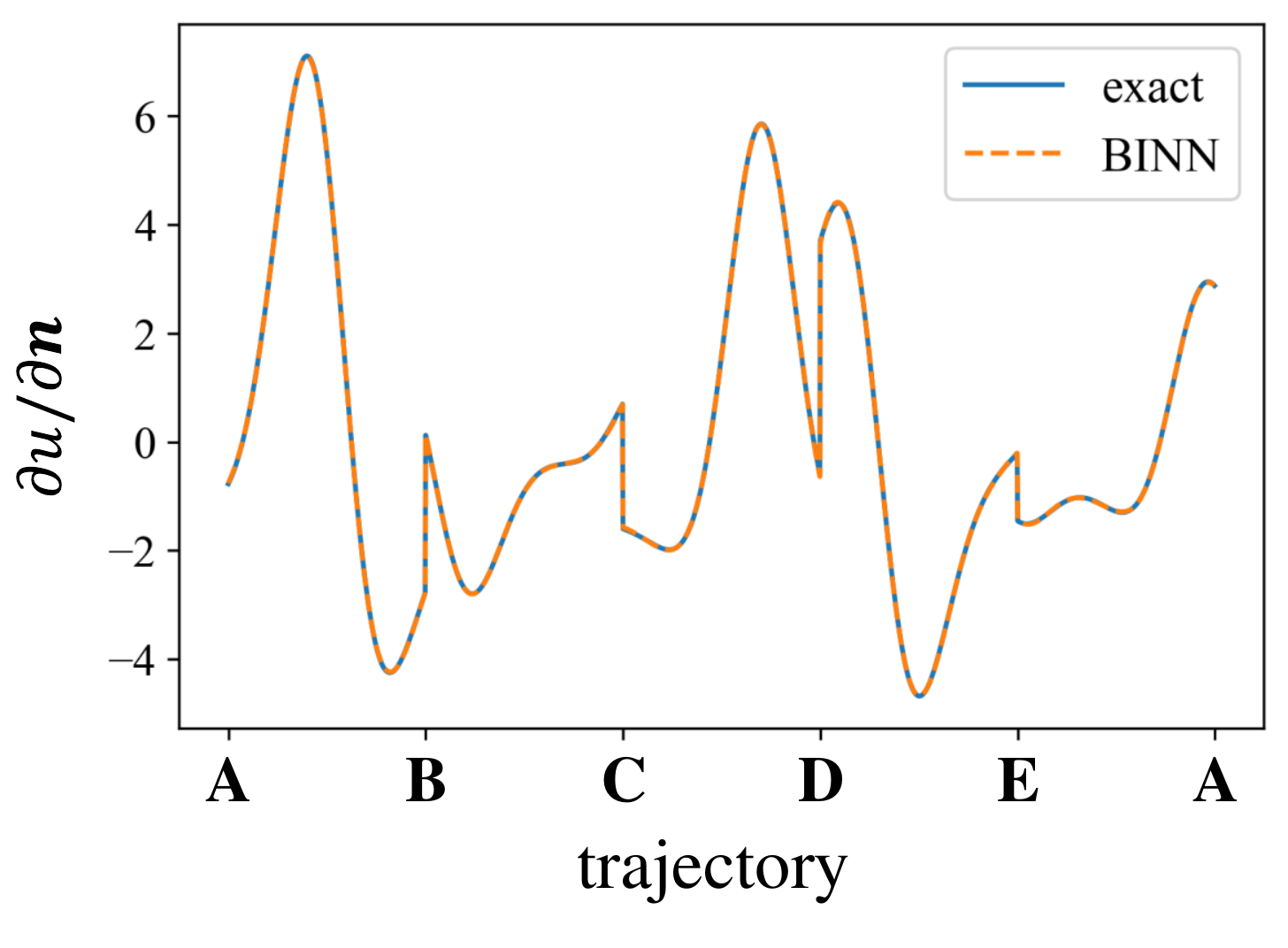}  

}
\quad
\subfigure[]{
\centering  
\includegraphics[height=3.7cm]{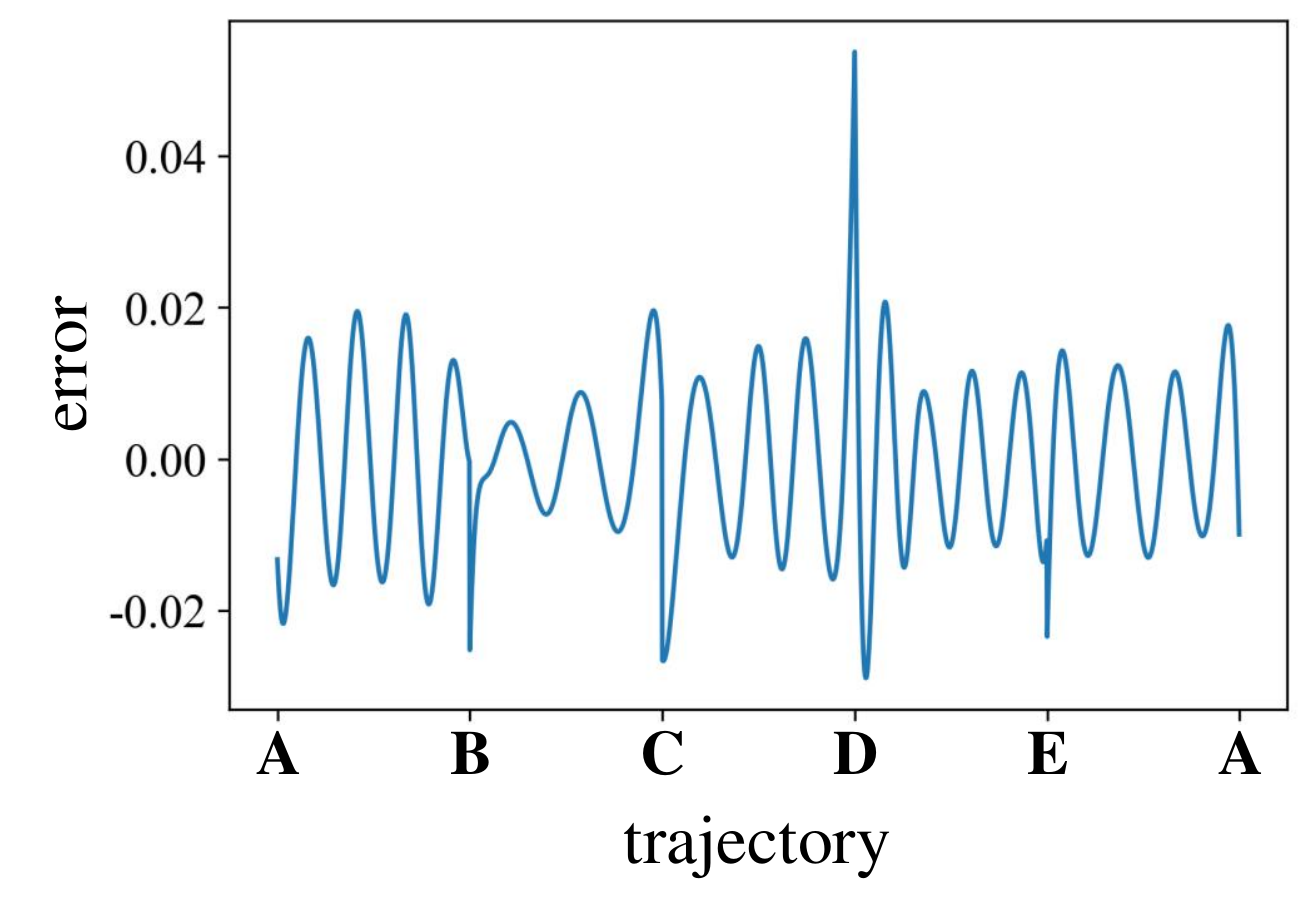}
}

\caption{(a) Evolution of the loss function during the training process of the potential problem on the flower-shaped region. (b) The BINN solution and the exact value of $\partial u/\partial \boldsymbol{n}$ along the boundary. (c) The absolute error between the prediction and the exact value of $\partial u/\partial \boldsymbol{n}$ along the boundary.}  
\label{flower_boundary}  
\end{figure}

The geometry is shown as fig.\ref{geo_flower}(a). The flower-shaped region is composed of 5 semi-circles with the radius $r=1$. 100 source points are allocated uniformly on the boundary, as shown in fig.\ref{geo_flower}(b). The boundary integrals (including the regular integrals and the regularized singular integrals) are evaluated piece-wisely using the Gaussian quadrature rule with 10 quadrature points. The neural networks are trained with 50000 iterations.
Fig.\ref{flower_boundary}(a) shows the evaluation of the loss function during the training process, which is quite typical in PINN-based methods. In this example, the unknowns are $\partial u/\partial \boldsymbol{n}$ on the boundary. The boundary results are presented along the trajectory demonstrated in fig.\ref{geo_flower}(c) with 2000 evenly distributed points. Fig.\ref{flower_boundary}(b) shows the results of the BINN solution and exact solution of $\partial u/\partial \boldsymbol{n}$ along the trajectory. 
Fig.\ref{flower_boundary}(c) shows the absolute error $\partial u/\partial \boldsymbol{n}-\partial \phi(\boldsymbol{x};\boldsymbol{\theta})/\partial \boldsymbol{n}$ along the trajectory. It can be observed that the solution of BINN agrees well with the accurate results.
\begin{figure}
\centering  
\includegraphics[width=14cm]{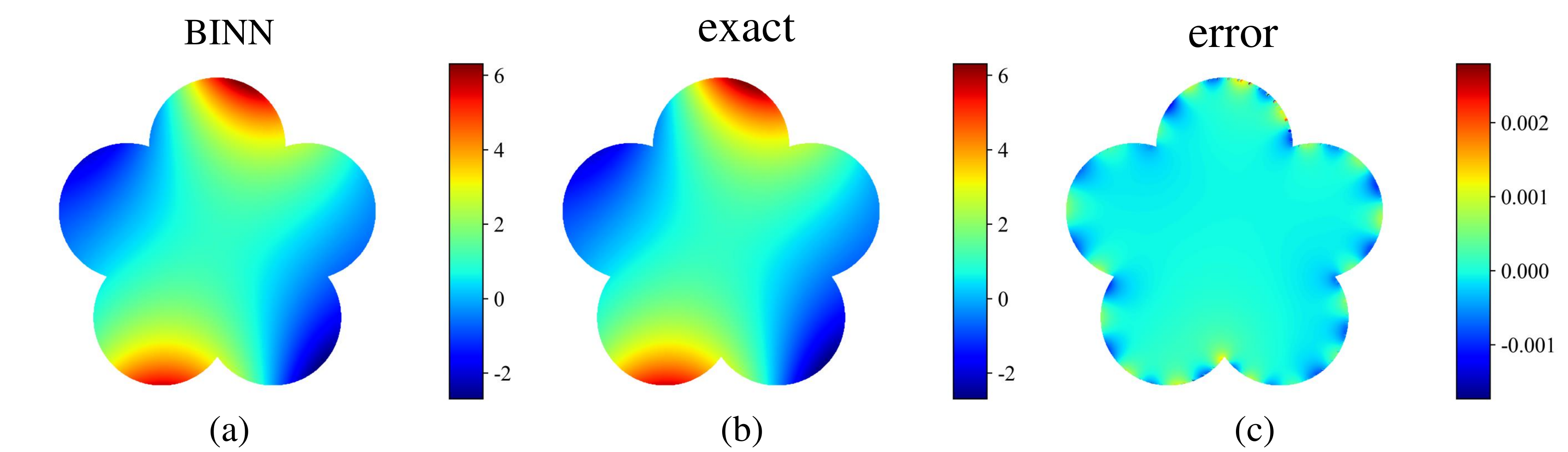}
\caption{ Results of the potential problem on a flower-shaped region. (a) The solution by BINN. (b) The exact solution. (c) The distribution of the absolute error. }  
\label{flower_inner}  
\end{figure}
\begin{figure}
\centering  
\subfigure[2$\times$ quadrature points]{   
\centering    
\includegraphics[height=3.5cm]{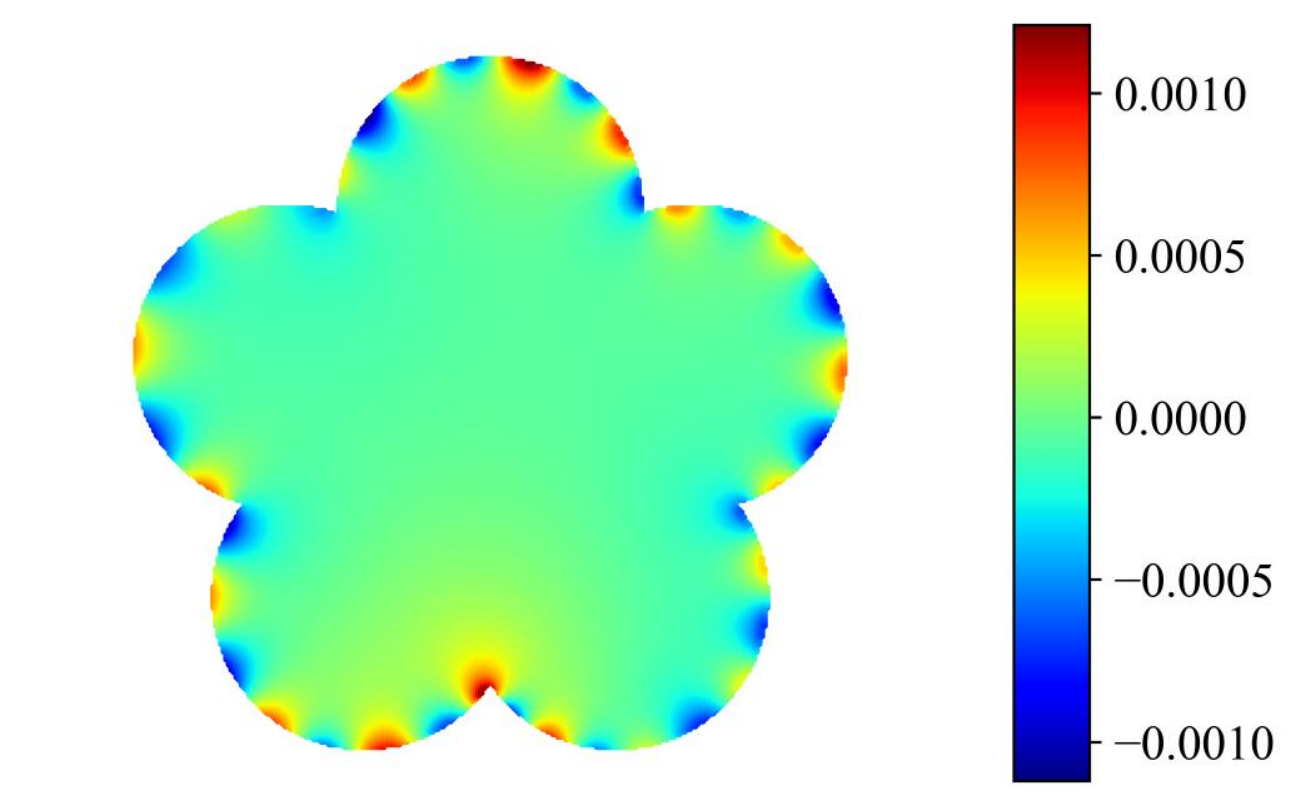}  

}
\quad
\subfigure[ 4$\times$ quadrature points]{

\centering  
\includegraphics[height=3.5cm]{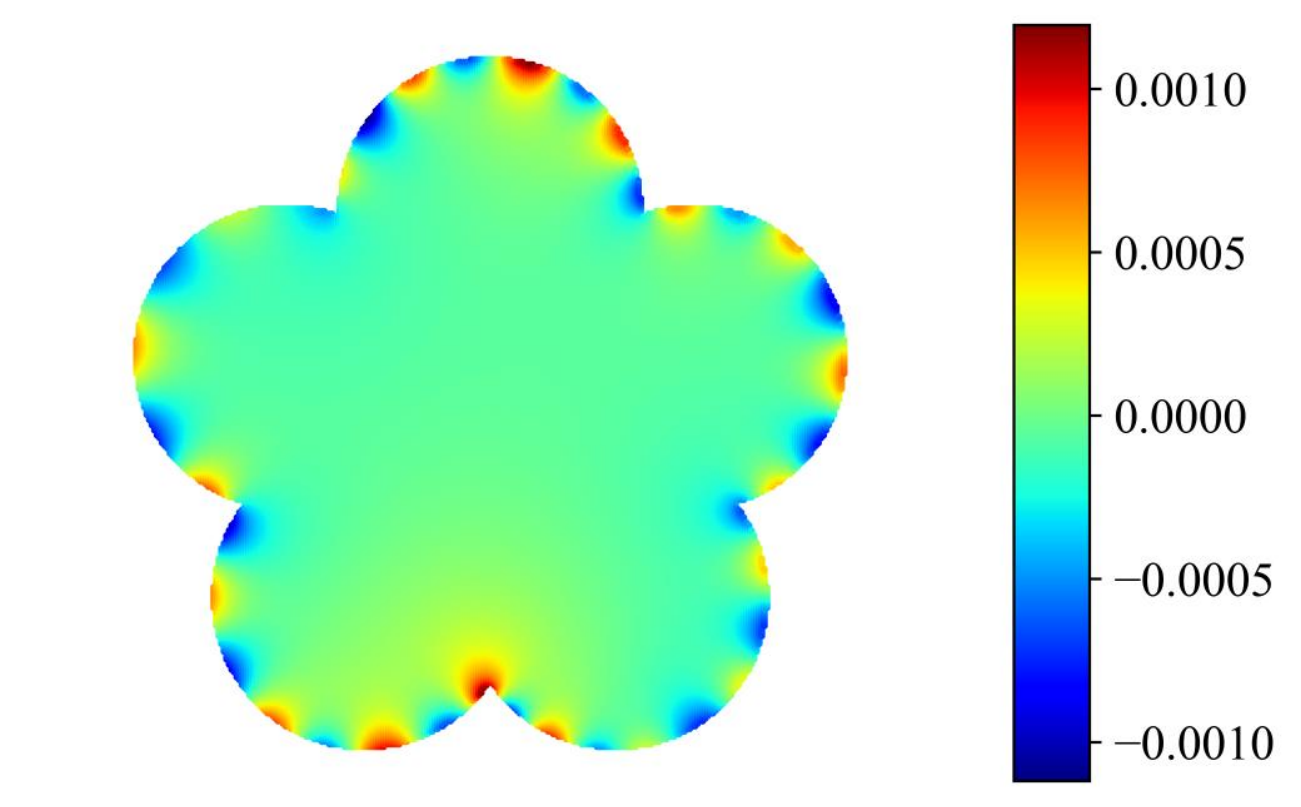}
}

\caption{(a) Error distribution with 2 times quadrature points. (b) Error distribution with 4 times quadrature points. Note that the parameters in the network are frozen in this stage.}  
\label{more_qpoints}  
\end{figure}

\begin{figure}
\centering  
\subfigure[]{   
\centering    
\includegraphics[height=4cm]{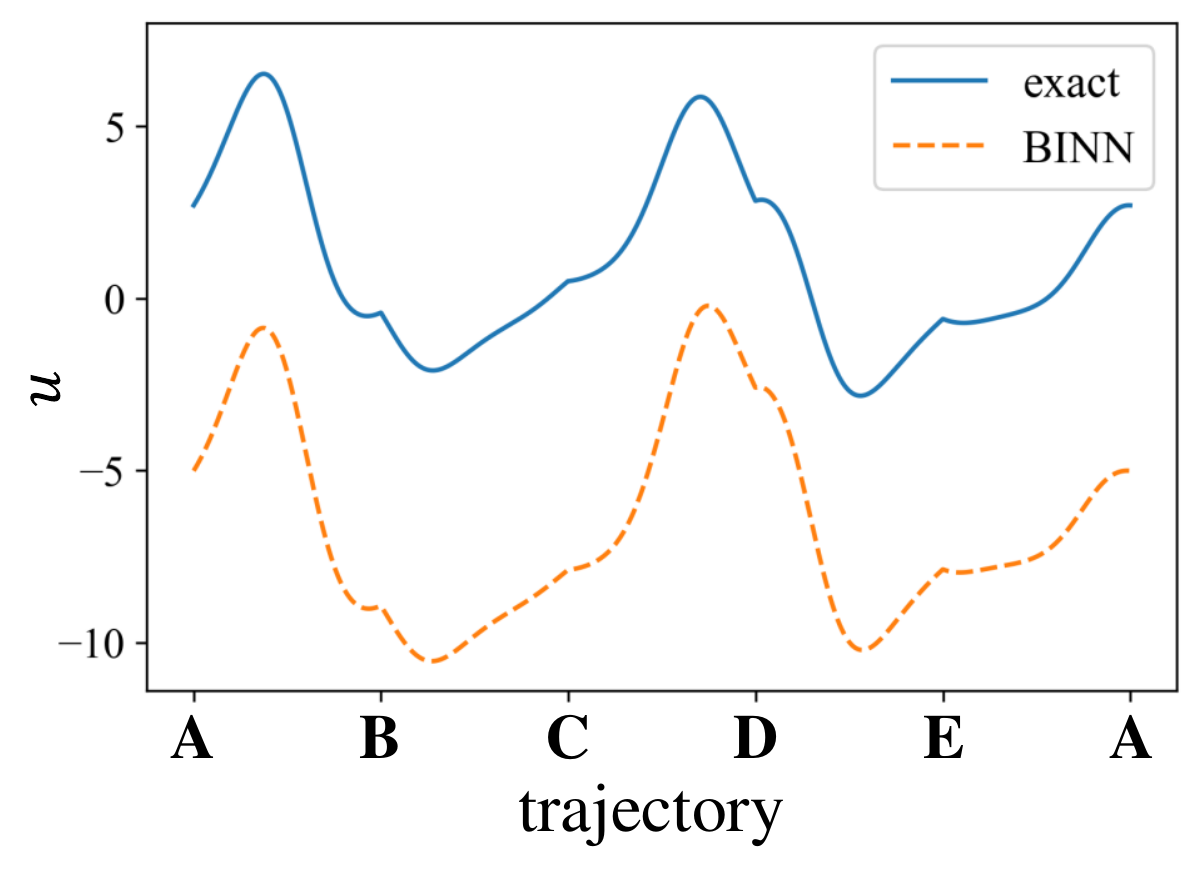}  

}
\quad
\subfigure[]{

\centering  
\includegraphics[height=4cm]{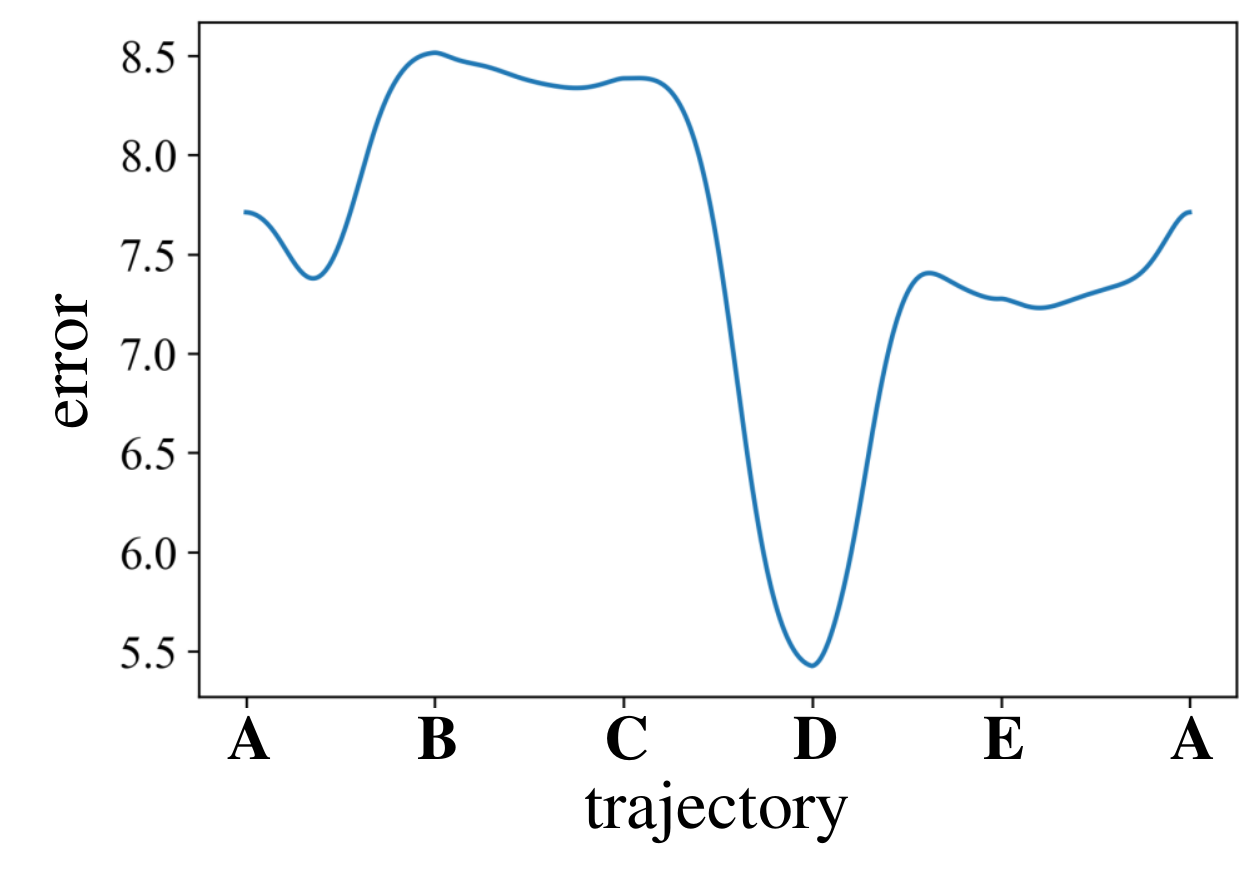}
}

\caption{ Results for the given boundary conditions of the potential problem on the flower-shaped region. (a) The results of $u(\boldsymbol{x})$ and $\phi(\boldsymbol{x;\theta})$ along the trajectory. (b) The discrepancy between $u(\boldsymbol{x})$ and $\phi(\boldsymbol{x;\theta})$ along the trajectory. It can be seen that the network itself does not have to satisfy the boundary conditions }  
\label{flower_ess}  
\end{figure}
After the acquisition of both the boundary value $u$ and $\partial u/\partial \boldsymbol{n}$, the values on the interior points are computed with eq(\ref{poi_d_inner}) with standard Gaussian quadrature rule. 
The results of the interior region are shown in fig.\ref{flower_inner}. As mentioned in section \ref{sec_inner}, the standard Gaussian quadrature rule is not available when the interior points are closed to the boundary, hence the interior points are kept with a distance $\varepsilon=0.03$ to the boundary in all the examples. It can be seen that the proposed method can produce a very precise result for the interior points, even better than the boundary value. The reason is that the errors on the boundary in fig.\ref{flower_boundary}(b) oscillate around zero. When computing the integrals in eq(\ref{poi_d_inner}), the boundary errors will cancel each other somehow, hence the results of the inner points have better accuracy. It can be also observed that the errors near the boundary are higher. This is induced by two main contributions: The first part is the error from the trail function itself $\phi(\boldsymbol{x;\theta})$. The second part is the integration error. As demonstrated before, when the inner points are close to the boundary, the fundamental solution will vary rapidly, hence the integration points may be not enough to accurately evaluate the integrals. The second part can be reduced by simply adding the number of quadrature points. It should be stressed that the parameters in the network have been frozen in this stage, and we can obtain $\partial{\phi(\boldsymbol{x;\theta})}/\partial{\boldsymbol{n}}$ at any boundary points. Therefore, when calculating the interior results, we can select a different set of integration points other than the ones used in the training stage. We evaluate the results of the interior region with 2 times and 4 times quadrature points than the training stage by simply adding the segments, and the errors are shown in fig.\ref{more_qpoints}.  Comparing fig.\ref{flower_inner}(c) with fig.\ref{more_qpoints}(a) it can be seen that the error can be reduced with more quadrature points. However, if we keep increasing the number of the quadrature points, the error will finally converge and be mainly dominated by the first part.

As demonstrated before, in BINN, the neural networks (or their derivative) are only requested to approximate the unknowns on the boundary to satisfy the boundary integral equations eq(\ref{poi_bie}). The network itself does not necessarily satisfy the given boundary conditions on the boundary. In this example, the unknowns are flux $\partial u/\partial \boldsymbol{n}\approx \partial{\phi(\boldsymbol{x;\theta})}/\partial{\boldsymbol{n}}$ on the boundary, and the value of $\phi(\boldsymbol{x;\theta})$ and the exact potential $u(\boldsymbol{x})$ produced by the trained model on the boundary are shown in fig.\ref{flower_ess}. It can be observed that although the flux is finely approximated, the network itself does not necessarily satisfy the essential boundary conditions, i.e., $\phi(\boldsymbol{x;\theta})\neq u(\boldsymbol{x})$.

\subsubsection{2D potential flow around a circular cylinder}

As demonstrated before, BINN can be conveniently employed for exterior problems, i.e., problems on the infinite or semi-infinite domain. Such problems are also common in practice such as a small cavity inside a sufficiently large body, airflow over an aerofoil, etc. In this example, we study a basic problem of 2D potential flow around a circular cylinder with uniform onset velocity $\boldsymbol{v}_0$ in the $x_1$ direction. The radius of the cylinder is $a=1.5$ and the magnitude of the onset velocity is $\|\boldsymbol{v}_0\|=3$. In potential flow problems, the velocity field $\boldsymbol{v}(\boldsymbol{x})$ can be determined through a potential function $u(\boldsymbol{x})$ defined as
\begin{equation}
    \boldsymbol{v}(\boldsymbol{x}) = \nabla u(\boldsymbol{x}).
\label{eq_potential}
\end{equation}

\begin{figure}
\centering
\subfigure[]{
\centering  
\includegraphics[height=4.5cm]{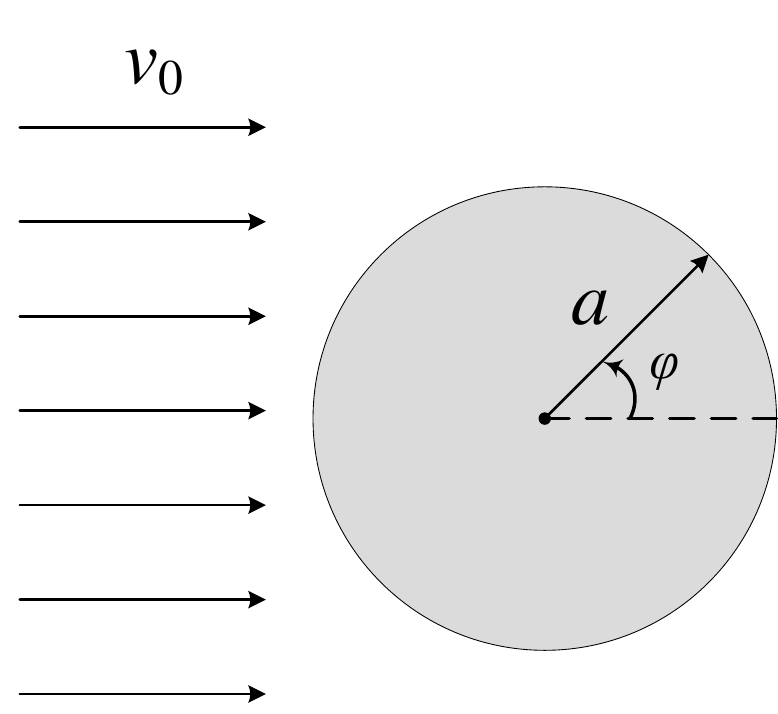}
}
\quad
\subfigure[]{

\centering  
\includegraphics[height=4.0cm]{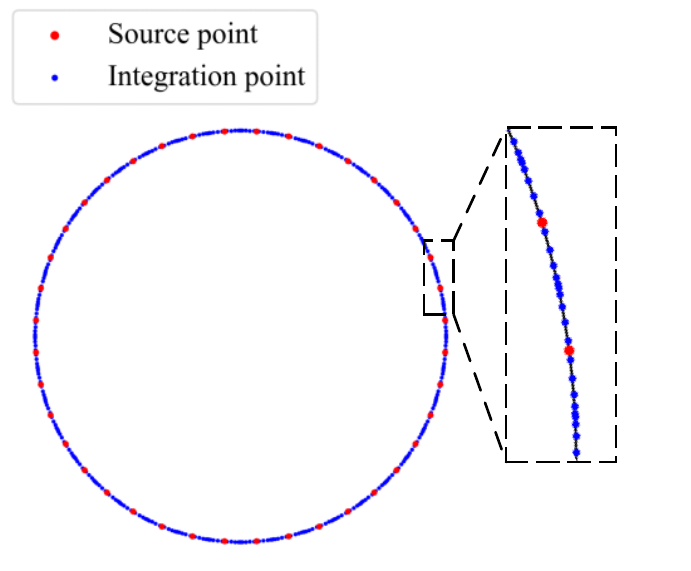}
}
\caption{(a) 2D potential flow over a cylinder. (b) The distribution of the source points and integration points}  
\label{flow_geometry}  
\end{figure}
\begin{figure}
\centering  
\subfigure[]{   
\centering    
\includegraphics[height=4cm]{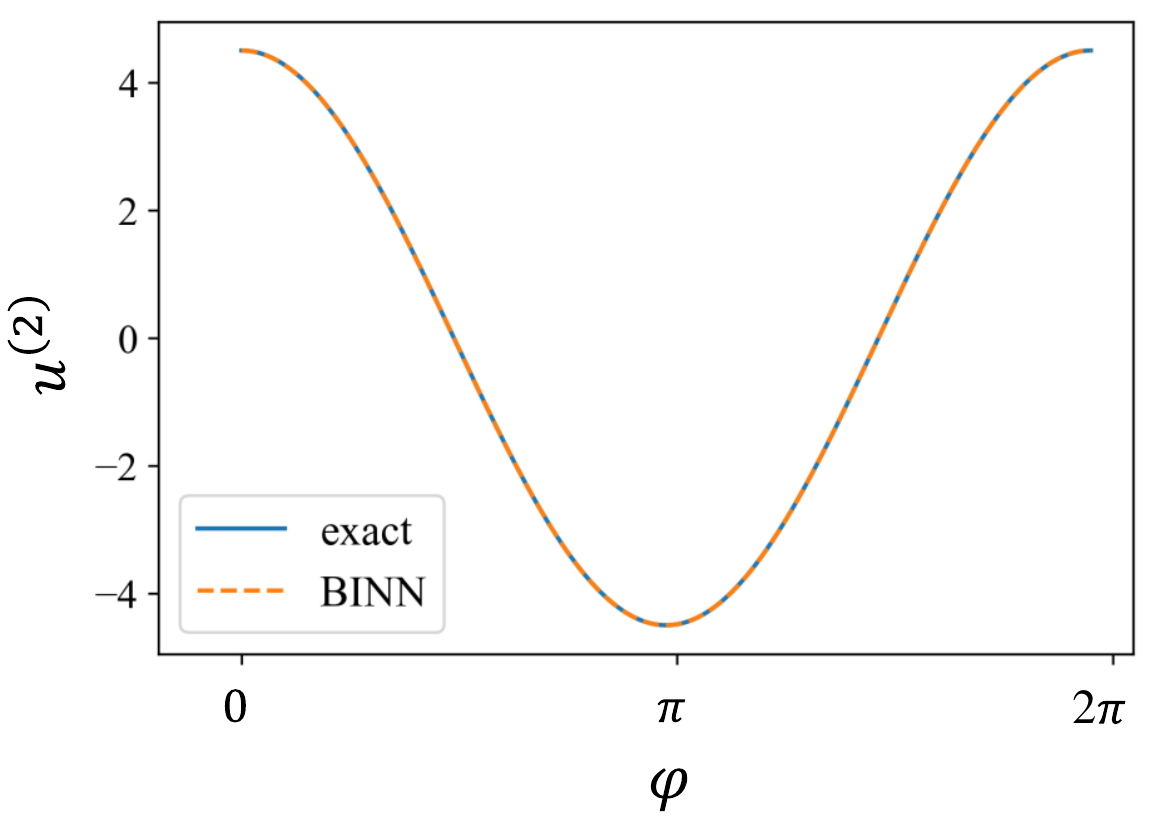}  

}
\quad
\subfigure[]{

\centering  
\includegraphics[height=4cm]{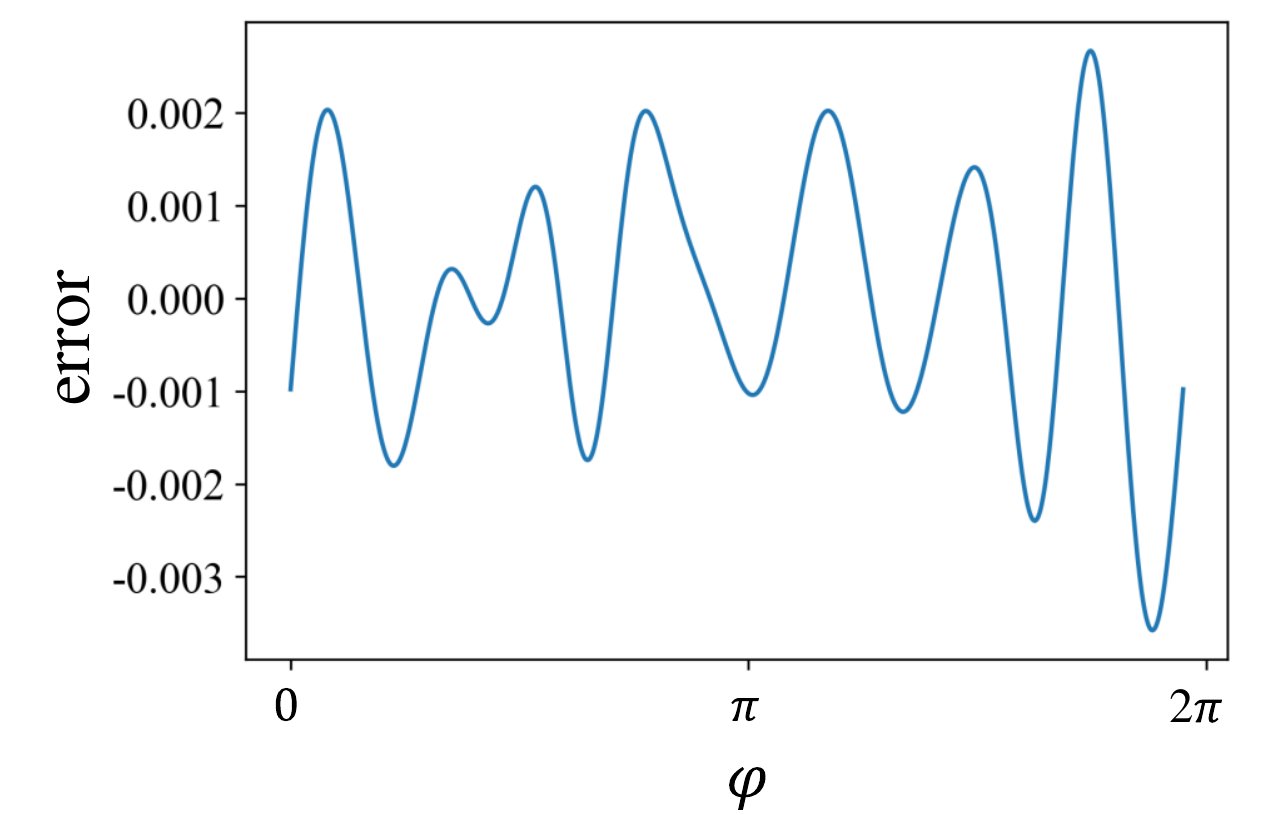}
}
\caption{ Results of the boundary solution to the 2D potential flow around a cylinder. (a) The solution of $u^{(2)}$ along the boundary. (b) The absolute error of $u^{(2)}$ between BINN and exact solution along the boundary.}  
\label{flow_boundary}  
\end{figure}
\begin{figure}
\centering  
\includegraphics[width=14cm]{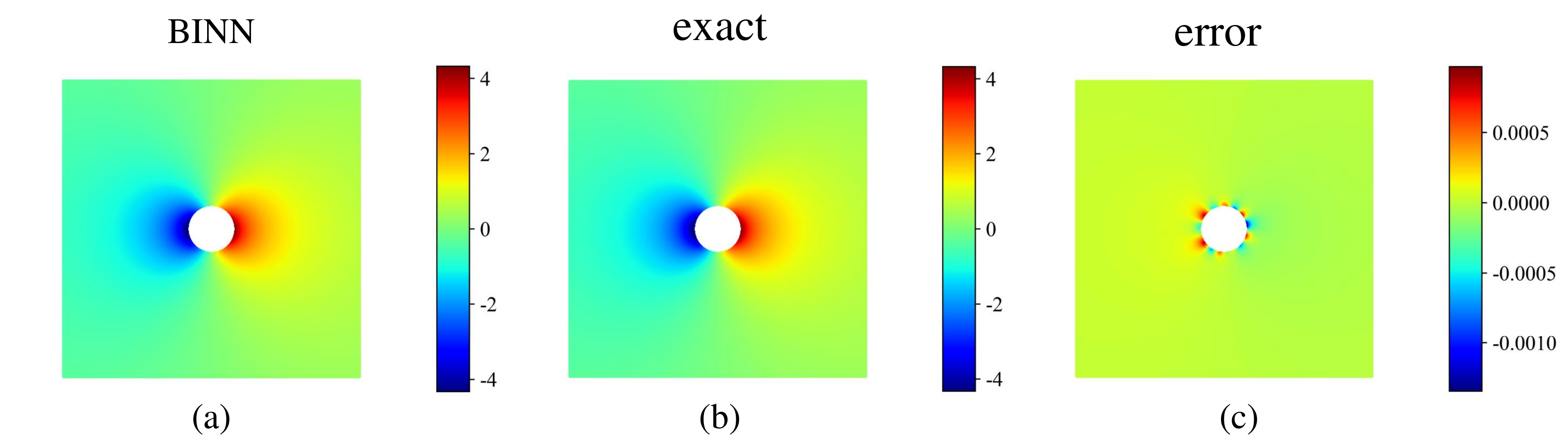}
\caption{ The results of the perturbation potential $u^{(2)}$ in the interior region.
(a) The solution by BINN. (b) The exact solution. (c) The distribution of the absolute error.}  
\label{flow_inner}  
\end{figure}
The potential function $u(\boldsymbol{x})$ is governed by the Laplace equation with Neumann boundary conditions:
\begin{equation}
\left\{ 
    \begin{aligned}
    \nabla^2 u(\boldsymbol{x}) &= 0 &in \quad \Omega,\\
    \dfrac{\partial u(\boldsymbol{x})}{\partial \boldsymbol{n}}&=0& in\quad \Gamma,\\
    \nabla u(\boldsymbol{x})&=\boldsymbol{v}_0 &when \quad \boldsymbol{x}\rightarrow \infty,\\
    \end{aligned}
\right.
\label{eq_flow}
\end{equation}
where $\Gamma$ denotes the boundary of the cylinder. In potential flow problems, a common treatment is to separate the potential function into two parts,
\begin{equation}
    u(\boldsymbol{x})=u^{(1)}(\boldsymbol{x})+u^{(2)}(\boldsymbol{x}),
    \label{eq_separate}
\end{equation}
where $u^{(1)}(\boldsymbol{x})=\|\boldsymbol{v}_0\|x_1$ defines the steady onset flow and $u^{(2)}(\boldsymbol{x})$ is a perturbation potential decays at infinity. Then the problem can be solved in terms of the perturbation $u^{(2)}(\boldsymbol{x})$. Substituting eq(\ref{eq_separate}) and  $u^{(1)}(\boldsymbol{x})=\|\boldsymbol{v}_0\|x_1$ into eq(\ref{eq_flow}) we have
\begin{equation}
\left\{ 
    \begin{aligned}
    \nabla^2 u^{(2)}(\boldsymbol{x}) &= 0 &in \quad \Omega,\\
    \dfrac{\partial u^{(2)}(\boldsymbol{x})}{\partial \boldsymbol{n}}&=-\|\boldsymbol{v}_0\|n_1& in\quad \Gamma,\\
    \nabla u^{(2)}(\boldsymbol{x})&=0 &when \quad \boldsymbol{x}\rightarrow \infty.\\
    \end{aligned}
\right.
\label{eq_flow_solution}
\end{equation}
The analytical solution of the potential function is:
\begin{equation}
    u(\boldsymbol{x}) = \|\boldsymbol{v}_0\|x_1+\frac{a^2\|\boldsymbol{v}_0\|x_1}{\left(x_1\right)^2+\left(x_2\right)^2}.
\label{eq_sol_flow}
\end{equation}

 The numerical model is shown in fig.\ref{flow_geometry}(a). 40 source points with 400 integration points are allocated on the boundary, as shown in fig.\ref{flow_geometry}(b). The network is trained with 10000 iterations. The comparison of BINN and accurate solution of $u^{(2)}$ along with the error distribution are shown in fig.\ref{flow_boundary} with 1000 evenly distributed points. Note that in this example, the ``interior" region is the infinite domain outside the cylinder, but the integration points are only required on the boundary of the cylinder. Once we obtain all the $\partial u/\partial \boldsymbol{n}$ and $u$ on the boundary, we can compute $u$ 
on any point outside the cylinder with eq(\ref{poi_d_inner}). For the sake of simplicity, we only present the results of $u^{(2)}$ in a $20\time20$ domain, as shown in fig.\ref{flow_inner}. It can be seen that the error mainly occurs near the boundary, and the result is pretty accurate in the far field.  

\begin{remark}
As shown in this example, it is convenient to implement BINN on problems with infinite regions, which is a typical superiority of BIE-based methods. It should be noticed that the solution for these problems is not unique if the boundary condition at infinity is not assigned, and the BIE formula is valid only when certain regularity conditions at infinity are fulfilled. For the sake of complicity, we only list the conclusion here, and the details can be found in  \cite{RN55}. Let $\Gamma$ denote the inner boundary, and $R$ denote the distance from the center of $\Gamma$. In BIE-based methods, to extend eq(\ref{poi_bie}) for infinite region, it is assumed that $u(\boldsymbol{x})$ behaves at most as $O(\ln{R})$ when $R\rightarrow \infty$ in 2D problems, and $O(1/R)$ in 3D problems. If the above condition is not satisfied, a common treatment is to decompose the undetermined solution into two parts: A particular part that meets the condition at infinity, and an undetermined part that meets the regularity condition. This is exactly what we did in eq(\ref{eq_separate}). The perturbation $u^{(2)}(\boldsymbol{x})$ governed by eq(\ref{eq_flow}) vanishes at infinity, which is indeed lower than $O(\ln{R})$.
\end{remark}

\subsection{Elastostatic problems}
Next, we will present the performance of BINN on elastostatic problems, which are governed by the Navier equations:
\begin{equation}
\left\{ 
    \begin{aligned}
    \nabla^2 \boldsymbol{u}(\boldsymbol{x})&+\frac{1}{1-2 \nu} \nabla (\nabla\cdot \boldsymbol{u}(\boldsymbol{x}))+\frac{1}{G} \boldsymbol{f}(\boldsymbol{x})=0 &in \quad \Omega,\\
    \boldsymbol{u}(\boldsymbol{x})&=\bar{\boldsymbol{u}}(\boldsymbol{x}) &on\quad \Gamma_{u},\\
    \boldsymbol{t}(\boldsymbol{x})&=\bar{\boldsymbol{t}}(\boldsymbol{x}) &on\quad \Gamma_{t},\\
    \end{aligned}
\right.
\label{Navier_eq}
\end{equation}
where $\boldsymbol{u}(\boldsymbol{x})$ is the displacement field, $\Gamma_{u}$ and $\Gamma_{t}$ denotes the essential and natural boundary, respectively. For a well-posed problem we have $\Gamma_{u}\bigcap\Gamma_{t}=\emptyset$ and $\Gamma_{u}\bigcup\Gamma_{t}=\Gamma$.  $\boldsymbol{t=\sigma\cdot n}$ denotes the traction on the boundary, where $\boldsymbol{n}$ is the outward normal vector, $\boldsymbol{\sigma}$ denotes the stress tensor calculated with the generalized Hooke law:
\begin{equation}
    \boldsymbol{\sigma} = 2G\boldsymbol{\epsilon}+\frac{2G\nu}{1-2\nu} tr\left(\boldsymbol{\epsilon}\right)\mathbf{I},
    \label{Hooke}
\end{equation}
where $G$ and $\nu$ denote the shear modulus and the Poisson ratio, respectively. $tr(\cdot)$ denotes the trace of the tensor, $\boldsymbol{\epsilon}$ is the Cauchy strain tensor defined with the geometric equation:
\begin{equation}
    \boldsymbol{\epsilon} = \frac{1}{2}\left(\nabla\boldsymbol{u}+(\nabla\boldsymbol{u})^T\right).
    \label{geometric}
\end{equation}

There are more than one ways to derive the boundary integral equations of the elastostatic problems. A derivation starting from the weighted residual method can be found in  \cite{RN55}. Similar to potential problems, the body force $\boldsymbol{f(x)}$ will only introduce a constant to the resulting BINN formulation. In the present work, we considered the problems with zero body force for simplicity.  The resulting equations in the component form are:
\begin{equation}
C_{\alpha \beta}(\boldsymbol{y}) u_{\beta}(\boldsymbol{y})+\int_{\Gamma} t_{\alpha \beta}^{\rm s}(\boldsymbol{x}; \boldsymbol{y}) u_{\beta}(\boldsymbol{x}) {\rm d} \Gamma(\boldsymbol{x})=\int_{\Gamma} u_{\alpha \beta}^{\rm s}(\boldsymbol{x}; \boldsymbol{y}) t_{\beta}(\boldsymbol{x}) {\rm d} \Gamma(\boldsymbol{x}),\quad \boldsymbol{x,y}\in \Gamma,\quad \alpha,\beta = 1,2,
\label{BIE}
\end{equation}
where $C_{\alpha \beta}(\boldsymbol{x})$ is a parameter depending on the continuity of the boundary at $\boldsymbol{x}$. $C_{\alpha \beta}(\boldsymbol{x})-\delta_{\alpha \beta}/2$ if the boundary is smooth, i.e., the tangent line is continuous at $\boldsymbol{x}$, where $\delta_{\alpha \beta}$ denotes the Kronecker delta. The fundamental solutions $u_{\alpha \beta}^{\rm s}(\boldsymbol{x}, \boldsymbol{y})$ and
$t_{\alpha \beta}^{\rm s}(\boldsymbol{x}, \boldsymbol{y})$ are  exactly the displacement and traction derived from the Kelvin solution of the 2D case, respectively:
\begin{equation}
\begin{aligned}
u_{\alpha \beta}^{\rm s}(\boldsymbol{x}, \boldsymbol{y})=&\frac{1}{8 \pi G(1-\nu)}\left[(3-4 \nu) \ln \left(\frac{1}{r}\right) \delta_{\alpha \beta}+r_{,\alpha} r_{,\beta}\right],\\
t_{\alpha \beta}^{\rm s}(\boldsymbol{x}, \boldsymbol{y})=&-\frac{1}{4 \pi (1-\nu)}\left\{\frac{\partial r}{\partial n}\left[(1-2 \nu) \delta_{\alpha \beta}+2r_{,\alpha} r_{,\beta}\right]+(1-2 \nu)\left( r_{,\alpha} n_{\beta}-r_{,\beta} n_{\alpha}\right)\right\},
\end{aligned}
\label{fund_elastic}
\end{equation}
where $r=\|\boldsymbol{x}-\boldsymbol{y}\|$ is the distance between $\boldsymbol{x}$ and $\boldsymbol{y}$. $n_{\alpha}$ is the outward normal vector. 
Once we have obtained all the boundary displacement and traction, the displacement field can be obtained by
\begin{equation}
 u_{\alpha}(\boldsymbol{y}) =\int_{\Gamma} u_{\alpha \beta}^{\rm s}(\boldsymbol{x};\boldsymbol{y}) t_{\beta}(\boldsymbol{x}) {\rm d} \Gamma(\boldsymbol{x})- \int_{\Gamma} t_{\alpha \beta}^{\rm s}(\boldsymbol{x}; \boldsymbol{y}) u_{\beta}(\boldsymbol{x}) {\rm d} \Gamma(\boldsymbol{x}),\quad \boldsymbol{x}\in \Gamma,\quad \boldsymbol{y}\in \Omega,\quad \alpha,\beta = 1,2.
\label{Navier_inner}
\end{equation}

Eq(\ref{Navier_inner}) is known as Somigliana's identity for displacements \cite{Somigliana_1}.
Similar to the potential problems, we will use a network $\boldsymbol{\phi}(\boldsymbol{x};\boldsymbol{\theta})$ to approximate the boundary unknowns:
\begin{equation}
\left\{ 
    \begin{aligned}
     \boldsymbol{t}(\boldsymbol{x})&\approx\hat{\boldsymbol{t}}(\boldsymbol{x};\boldsymbol{\theta}) &in \quad \Gamma_{u},\\
    \boldsymbol{u}(\boldsymbol{x})&\approx\boldsymbol{\phi}(\boldsymbol{x};\boldsymbol{\theta}) &in \quad \Gamma_{t},\\
    \end{aligned}
\right.
\label{Navier_approx}
\end{equation}
where the approximate traction $\hat{\boldsymbol{t}}(\boldsymbol{x};\boldsymbol{\theta})$ is computed by substituting  $\boldsymbol{u}(\boldsymbol{x})=\boldsymbol{\phi}(\boldsymbol{x};\boldsymbol{\theta})$ into eq(\ref{geometric}), eq(\ref{Hooke}) and $\boldsymbol{t=\sigma\cdot n}$.
For a given source point $\boldsymbol{y}^i$, the residual of the BIE formula can be computed similarly by substituting eq(\ref{Navier_approx}) and all the boundary conditions into eq(\ref{BIE}), then we can compute the loss function of the same form as eq(\ref{poi_loss}) and train the network to solve all the boundary unknowns.

\subsubsection{Beam under shear loading}

We first consider the solution of the elastic beam under shear loading, which is a common benchmark for elastostatic problems. The analytical solution of the displacement field is given by:
\begin{equation}
\begin{aligned}
u_{1}=&-\frac{Px_2}{6EI}\left[\left(6L-3x_1\right)x_1+\left(2+\nu\right)\left(x_2^2-\frac{D^2}{4}\right)\right],\\
u_{2}=&\frac{P}{6EI}\left[3\nu x_2^2\left(L-x_1\right)+\left(4+5\nu\right)\frac{D^2x_1}{4}+\left(3L-x_1\right)x_1^2\right],
\end{aligned}
\label{eq_beam}
\end{equation}
where $P$ denotes the magnitude of shear force integrated from the shear stress. $E,v$ denotes Young's modulus and the Poisson ratio, respectively. $L$ and $D$ denote the length and height of the beam, respectively. $I$ is the inertia moment of the beam section. The stress field can be calculated from eq(\ref{eq_beam}) using eq(\ref{geometric}) and eq(\ref{Hooke}):
\begin{equation}
\begin{aligned}
\sigma_{11}=&-\frac{P(L-x_1)x_2}{I},\\
\sigma_{22}=&0,
\\
\sigma_{12}=&\frac{P}{2I}(\frac{D^2}{4}-x_2^2).
\end{aligned}
\label{eq_beam_s}
\end{equation}

\begin{figure}
\centering  
\subfigure[]{
\centering  
\includegraphics[height=4.2cm]{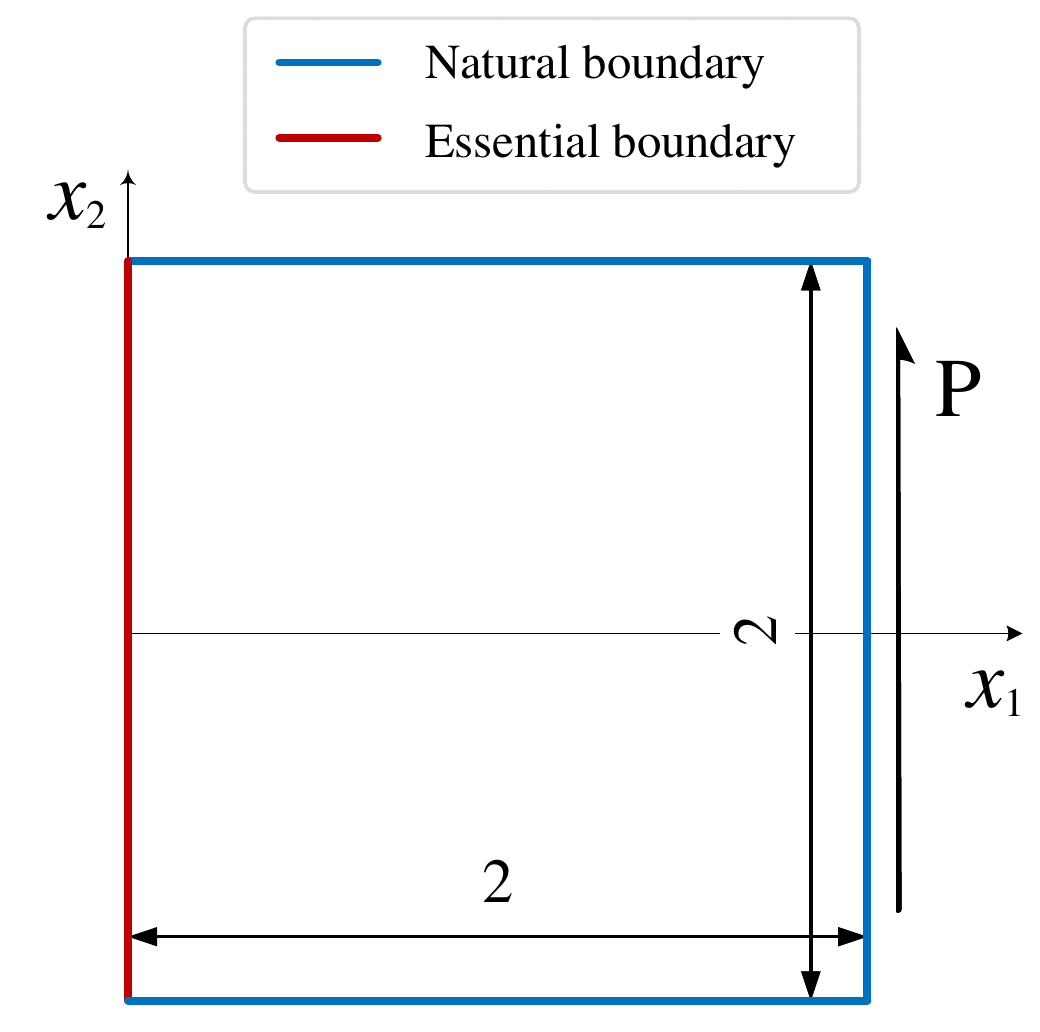}
}
\quad
\subfigure[]{

\centering  
\includegraphics[height=4.0cm]{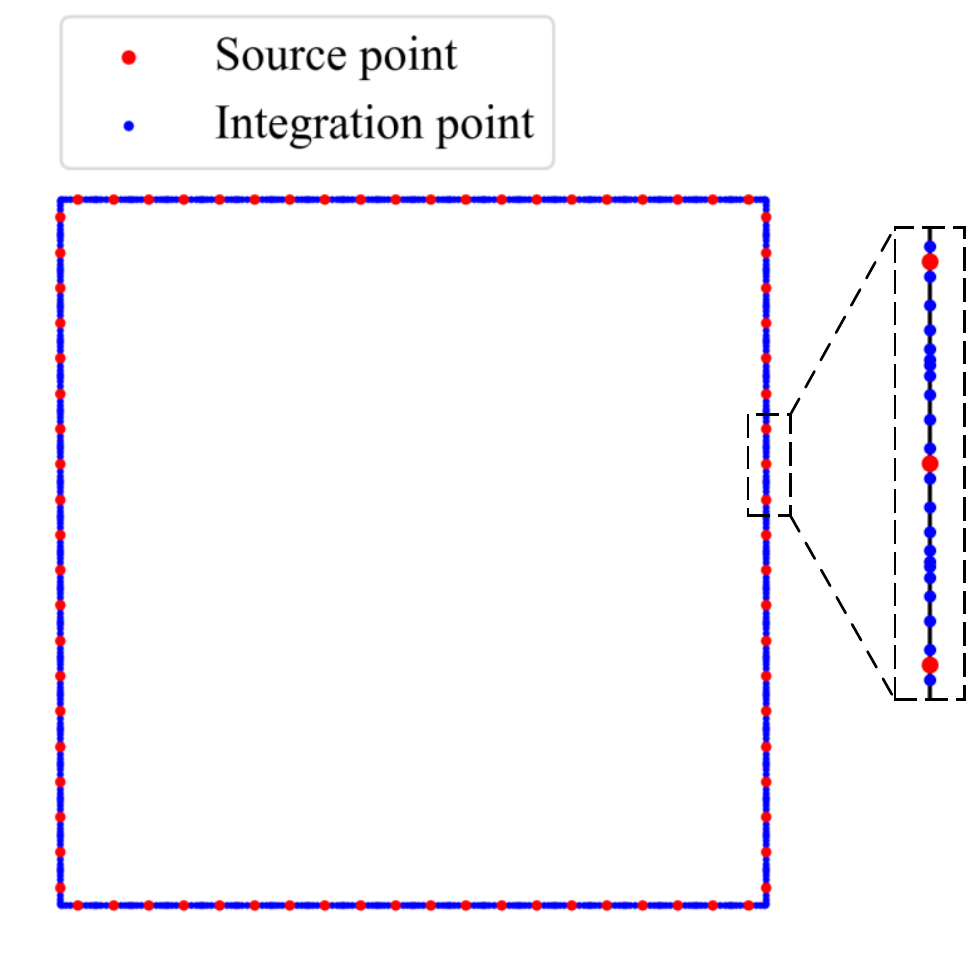}
}
\quad
\subfigure[]{

\centering  
\includegraphics[height=3.5cm]{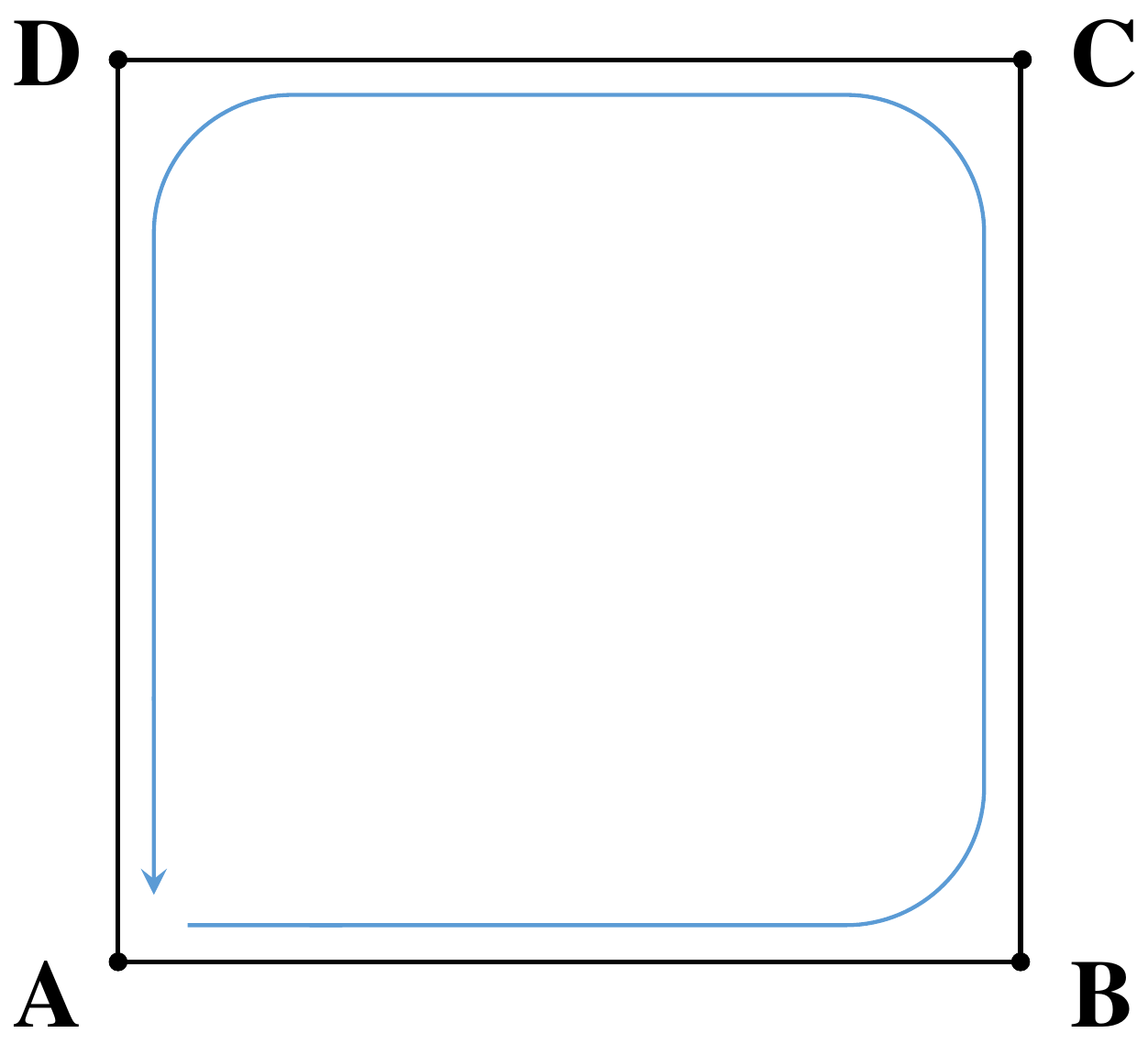}
}
\caption{(a) The geometry and the type of boundary conditions for the beam block. (b) The distribution of the source points and integration points. (c) The trajectory to show the boundary results.}
\label{beam_geometry}  
\end{figure}

\begin{figure}
\centering  
\includegraphics[width=12cm]{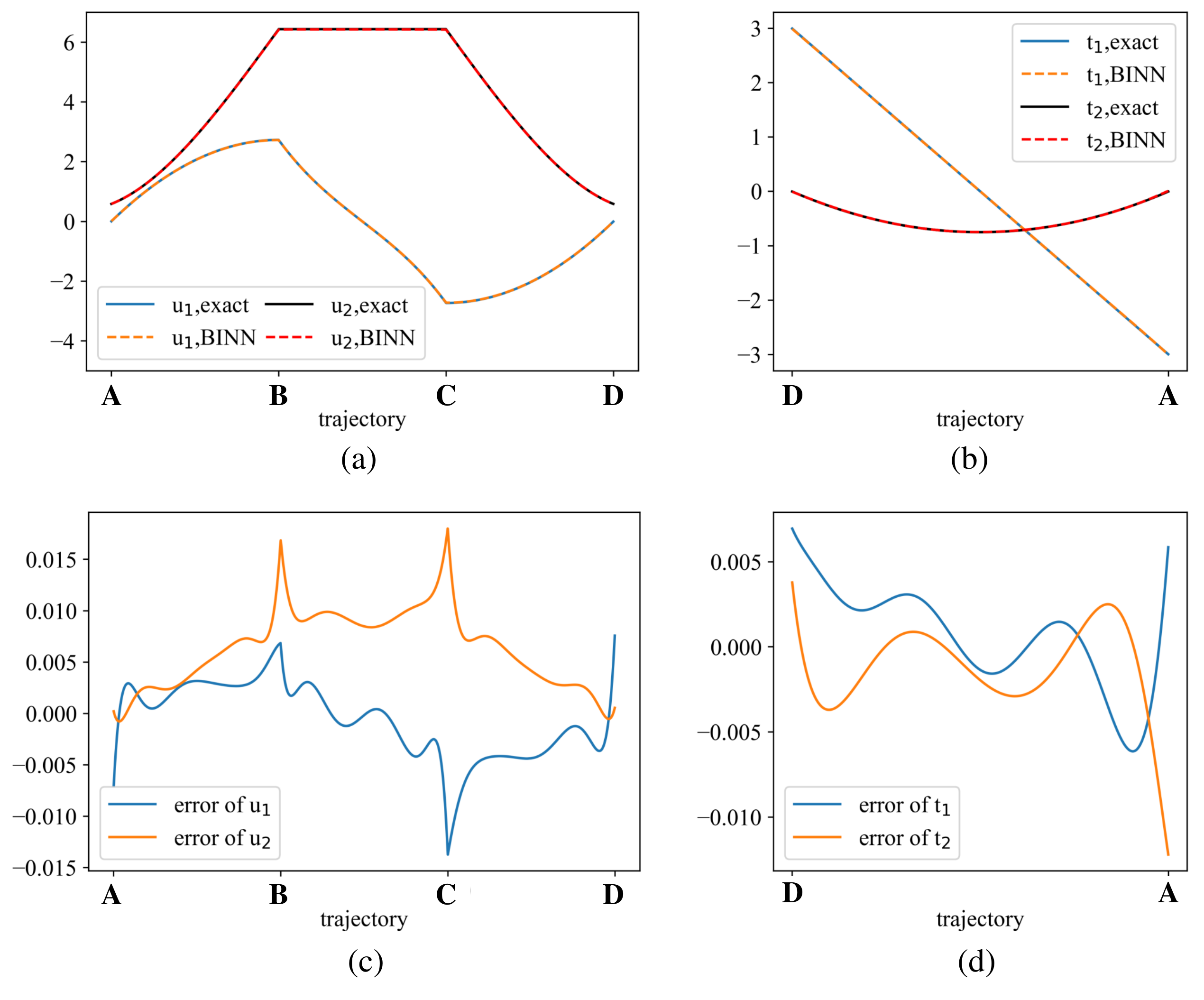}
\caption{ The results of the boundary solution to the beam problem. On the trajectory $\textbf{A}$-$\textbf{B}$-$\textbf{C}$-$\textbf{D}$, the boundary unknowns are $\boldsymbol{u}$, and the results are shown in (a), while on the trajectory $\mathbf{D}$-$\textbf{A}$ the unknowns are $\boldsymbol{t}$, and the results are shown in (b). The corresponding distributions of the absolute error between BINN and the exact solution along the trajectories are shown in (c) and (d), respectively. }  
\label{beam_boundary}  
\end{figure}

\begin{figure}
\centering  
\includegraphics[width=12cm]{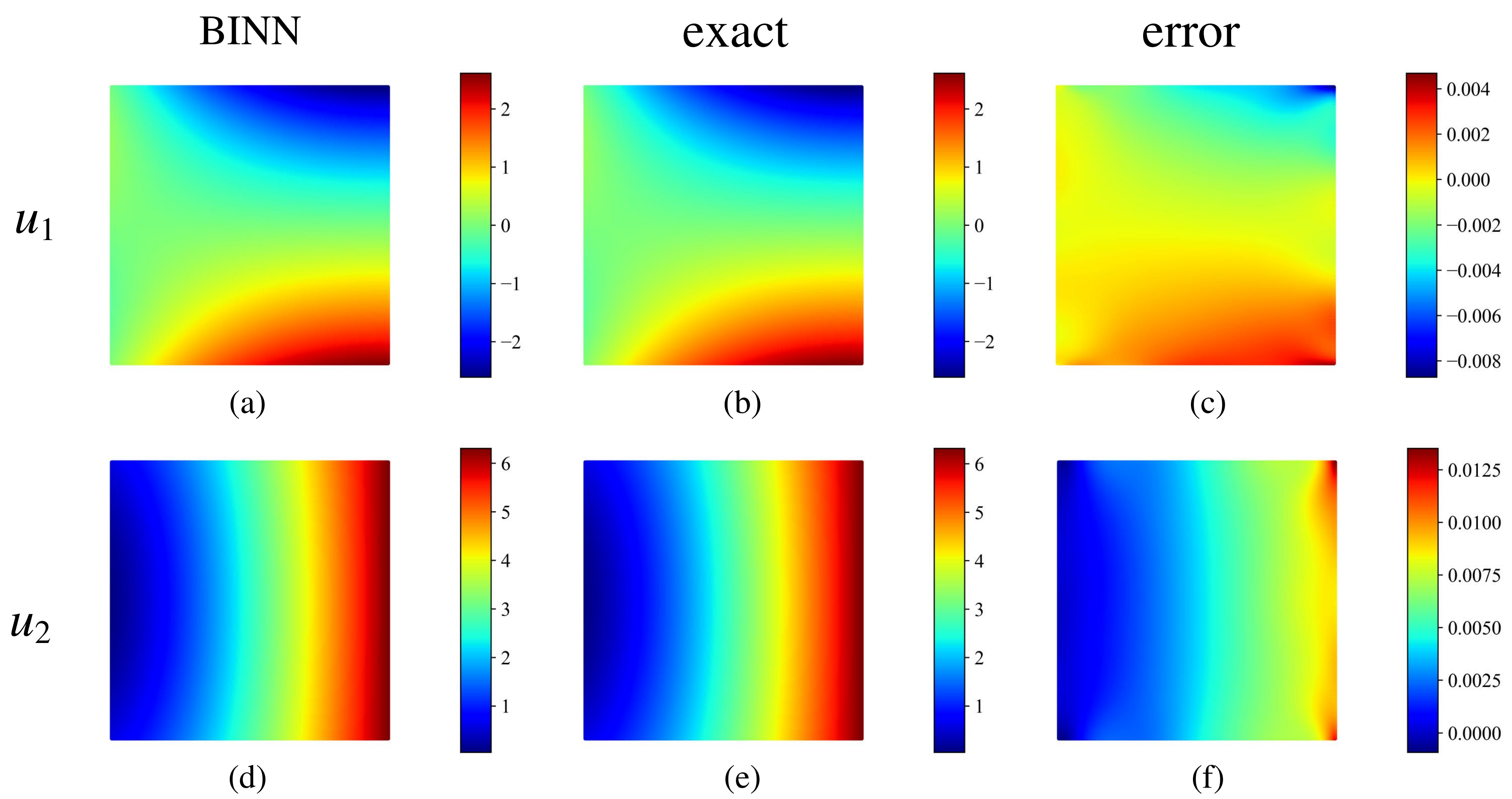}
\caption{The interior results of the beam problem. (a) and (d) are the displacement results $u_1$ and $u_2$ by BINN, respectively. (b) and (e) are the exact solution of the displacement $u_1$ and $u_2$, respectively. (c) and (f) are the  distribution of the absolute error between BINN and exact solution.}  
\label{beam_interia}  
\end{figure}
In this example, we consider a $2\times2$ domain, i.e., $L=D=2$ in eq(\ref{eq_beam}) and eq(\ref{eq_beam_s}). The shear force is taken as $P=1$. The elastic constants are $E=1$ and $v=0.3$. Mixed boundary conditions are considered, where essential BC is applied for the left boundary and natural BC is applied for the rest boundaries, as shown in fig.\ref{beam_geometry}(a). Note that the boundary conditions are applied directly following the analytical solution eq(\ref{eq_beam}) and eq(\ref{eq_beam_s}), hence the domain does not require to be slender to follow the hypothesis of the beam structure, and can be viewed as a block separated from the beam. 80 source points are evenly allocated on the boundary. The distribution of the source and integration points are shown in fig.\ref{beam_geometry}(b). The network was trained with 50000 iterations. The boundary results are observed on 4000 evenly distributed points along the trajectory in fig.\ref{beam_geometry}(c).
The solutions of the boundary unknowns are shown in fig.\ref{beam_boundary}(a)-(b). On the trajectory $\textbf{A}$-$\textbf{B}$-$\textbf{C}$-$\textbf{D}$, the boundary unknowns are $u$, while on the trajectory $\mathbf{D}$-$\textbf{A}$ the unknowns are $t$. The corresponding errors are shown in fig.\ref{beam_boundary}(c)-(d).
The results of the displacement field on the interior region are shown in fig.\ref{beam_interia}. It can be seen that BINNs could produce accurate results in the present problem. 

\subsubsection{Hertz contact}
\begin{figure}[htbp]
\centering
\subfigure[]{

\centering  
\includegraphics[height=3.5cm]{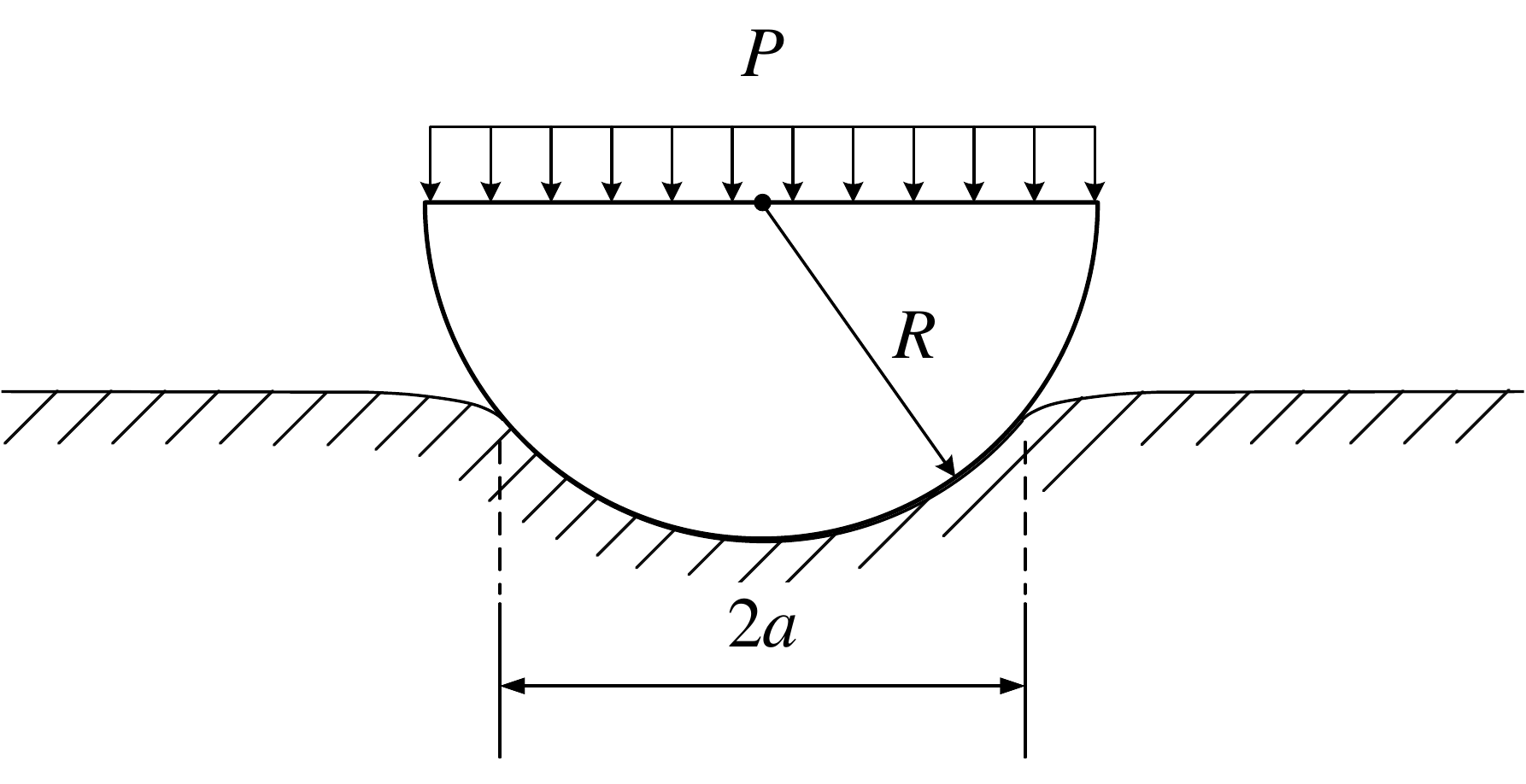}
}
\quad
\subfigure[]{

\centering  
\includegraphics[height=3.5cm]{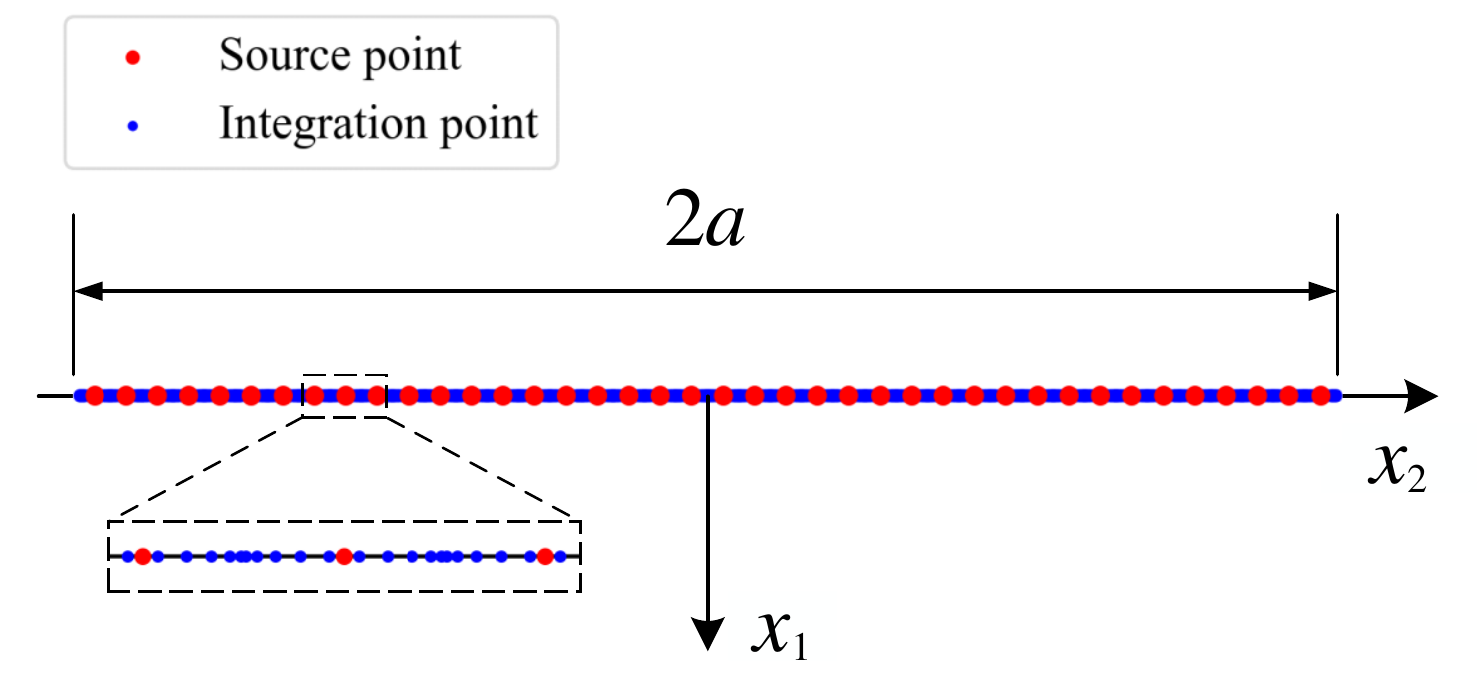}
}
\quad
\subfigure[]{

\centering  
\includegraphics[height=3.5cm]{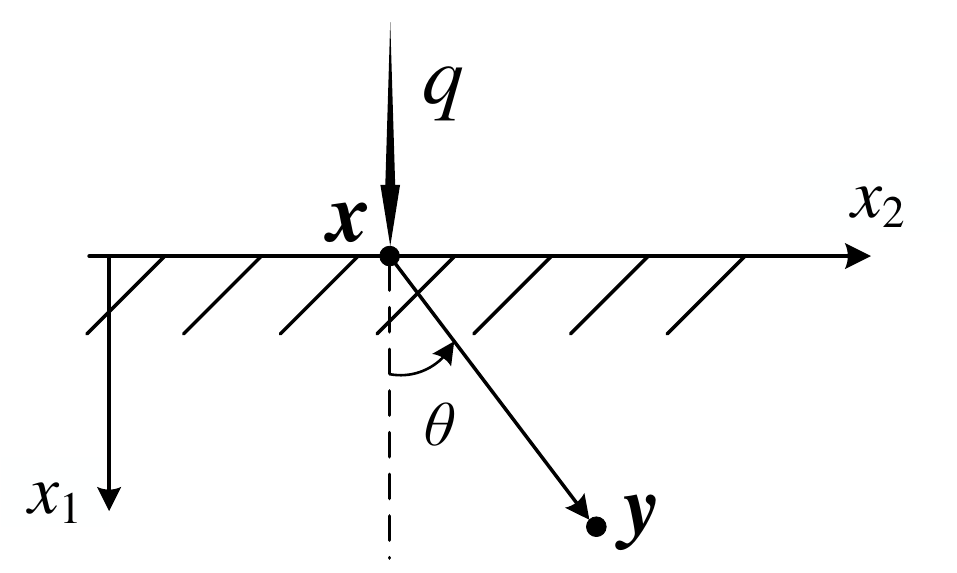}
}
\caption{(a) Illustration of the Hertz contact problem. (b) The distribution of the source points and integration points. (c) Illustration of the Flamant solution.}  
\label{hertz_geometry}  
\end{figure}

\begin{figure}
\centering
\subfigure[]{

\centering  
\includegraphics[height=4cm]{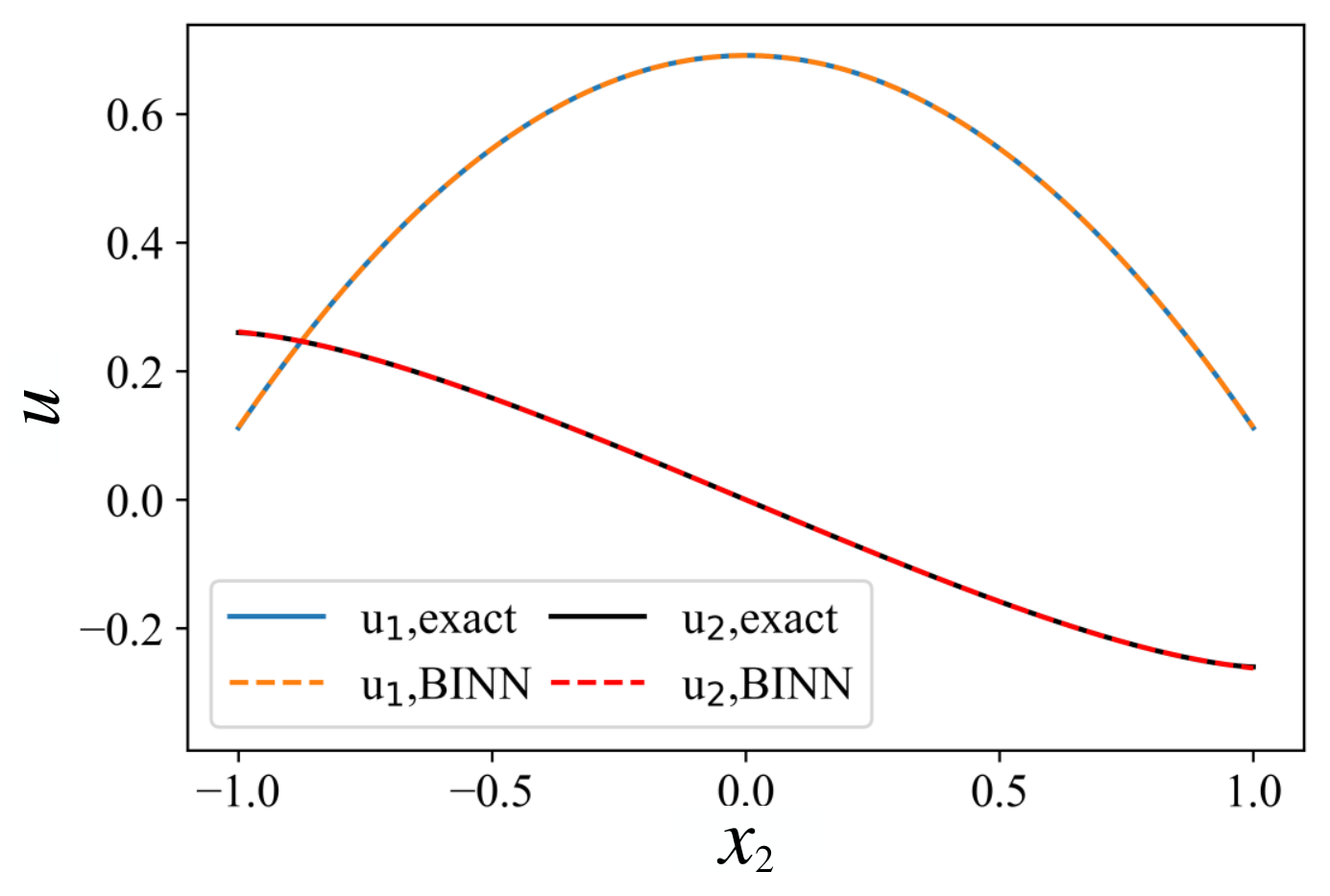}
}
\quad
\subfigure[]{

\centering  
\includegraphics[height=4cm]{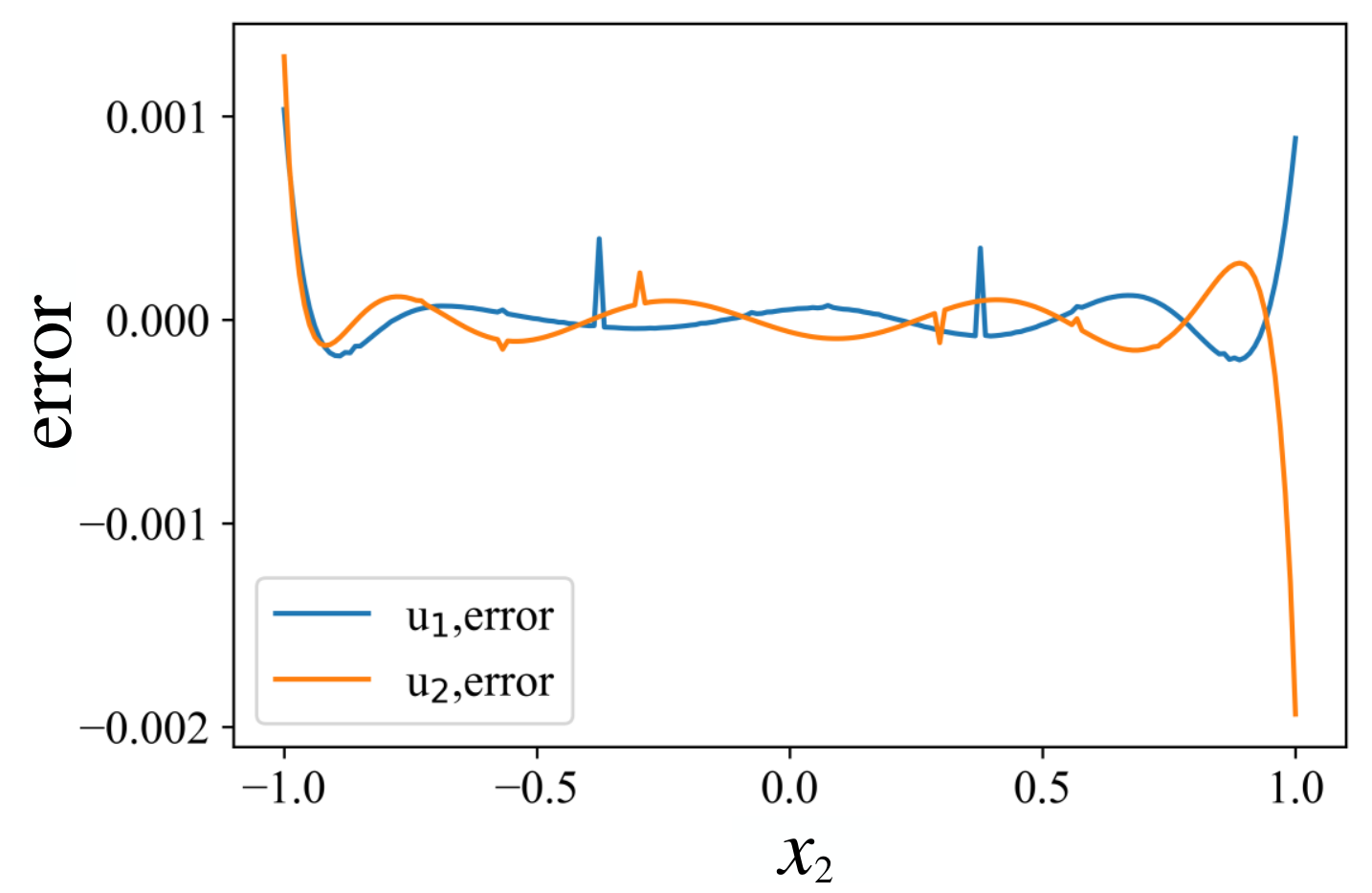}
}
\caption{ Results of the boundary solution to the Hertz contact problem. (a) The BINN solution and the exact value of $\boldsymbol{u}$ on the contact area. (b) The distribution of the absolute error between BINN and exact solution along on the contact area.}  
\label{hertz_boundary}  
\end{figure}
\begin{figure}
\centering  
\includegraphics[width=13cm]{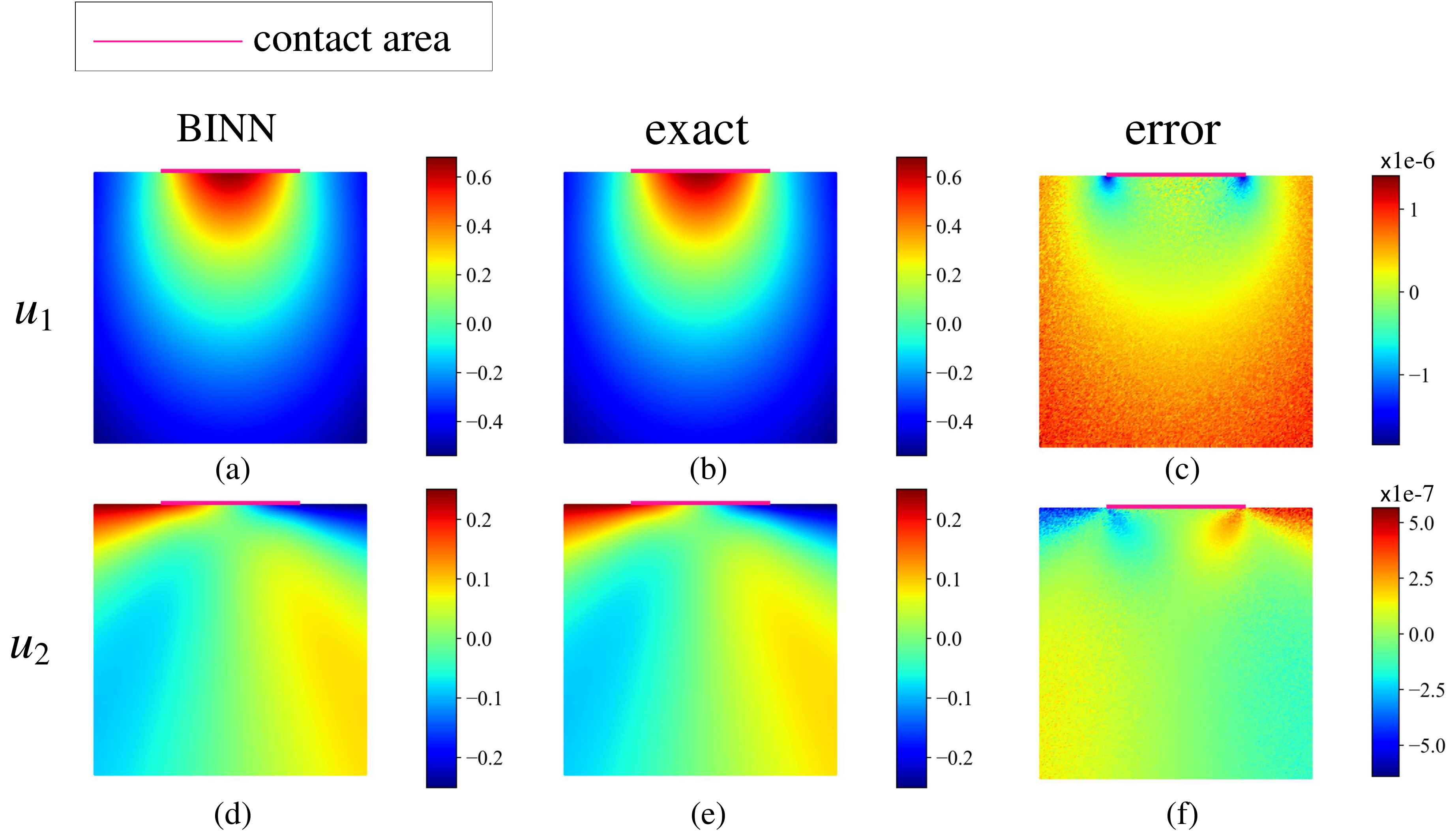}
\caption{The interior results of the Hertz contact problem. (a) and (d) are the displacement results $u_1$ and $u_2$ by BINN, respectively. (b) and (e) are the exact solution of the displacement $u_1$ and $u_2$ computed by the semi-analytical formula eq(\ref{eq_Flamant}), respectively. (c) and (f) are the corresponding error distribution.}  
\label{hertz_interia}  
\end{figure}
As mentioned before, BINN can be easily implemented to problems with semi-infinite region. In BIE-based methods, the unknowns are only assigned to the loading area of the interface, which is much simpler and more convenient for numerical implementation. Therefore, the BIE-based methods, such as the boundary element method, have been widely used in the investigations of contact mechanics. Hertz contact problem is one of the most important models in contact mechanics. In this example, we consider a typical case of the 2D Hertz contact problem: The indentation of a long rigid cylinder into an elastic half-plane. All surfaces are assumed to be frictionless. The analytical solution of the contact force is given by:
\begin{equation}
p(\boldsymbol{x}) = \frac{2P}{\pi a^2}\sqrt{(a^2-x_2^2)},
\label{eq_contact_f}
\end{equation}
where $P$ is the total indentation force integrated from the pressure, $a = \sqrt{(4PR(1-\nu^2)/(\pi E))}$ is the half-width of the contact area. In this example, we take $P=1, a=1$ with the elastic constants $E=1, v=0.3$.
The reference solution of the displacement field can be calculated semi-analytically with the superposition principle of the Green function:
\begin{equation}
u_{\alpha}(\boldsymbol{y}) = \int_{\Gamma}p(\boldsymbol{x})u_{\alpha}^{F}(\boldsymbol{x,y})d\Gamma(\boldsymbol{x}) ,\quad \alpha=1,2,
\label{eq_Flamant}
\end{equation}
where $u_{\alpha}^{F}(\boldsymbol{x,y})$ is a special case of the Flamant solutions as demonstrated in fig.\ref{hertz_geometry}(c), which means the displacement at $\boldsymbol{y}$ induced by a unit concentrated force $\boldsymbol{q}$ vertically acting on the interface an elastic half-plane point at $\boldsymbol{x}$:
\begin{equation}
\begin{aligned}
u_{1}^{F} &= -\frac{1}{2\pi G}\left[2(1-\nu)\ln{r}-\cos^2\theta\right],\\
u_{2}^{F} &= -\frac{1}{2\pi G}\left[(1-2\nu)\theta-\cos^2\theta\sin^2\theta\right],
\end{aligned}
\label{eq_Flamant_1}
\end{equation}
where $r$ is the distance between $\boldsymbol{x}$ and $\boldsymbol{y}$, $G$ is the shear module and $\nu$ is the Poisson ratio.
In the present work, the reference solution is computed with eq(\ref{eq_Flamant}) using the adaptive quadrature function $integral()$ in MATLAB with 12 decimal places of accuracy.

The fundamental solution for elastic half-plane is the superposition of the kelvin solution eq(\ref{fund_elastic}) and an auxiliary solution \cite{Telles_2}. For the sake of simplicity, we put the formulas of the auxiliary solution in \ref{App_2}.
In this example, the natural boundary condition is considered, i.e., we apply a Hertz contact force that follows the form eq(\ref{eq_contact_f}) on the boundary, and the unknowns are the boundary displacement on the contact area.
It should be stressed that in the semi-infinite problems, we only require to solve the properties on the contact area, hence the scale of the problem is greatly reduced, as shown in fig.\ref{hertz_geometry}(b). 20 source points with 200 integration points are allocated on the contact surface. 

The networks are trained by 10000 iterations. The results of the boundary displacement observed with 1000 evenly distributed points on the contact area are shown in fig.\ref{hertz_boundary}. The BINN solution agrees well with the exact solution. Similar to the case of the infinite domain, the displacement of any point in the infinite region can be computed through eq(\ref{Navier_inner}). For simplicity, we present the results on a $4\times4$ domain. The results are shown in fig.\ref{hertz_interia}.

\subsubsection{Inclusion}
\begin{figure}
\centering  
\includegraphics[width=12cm]{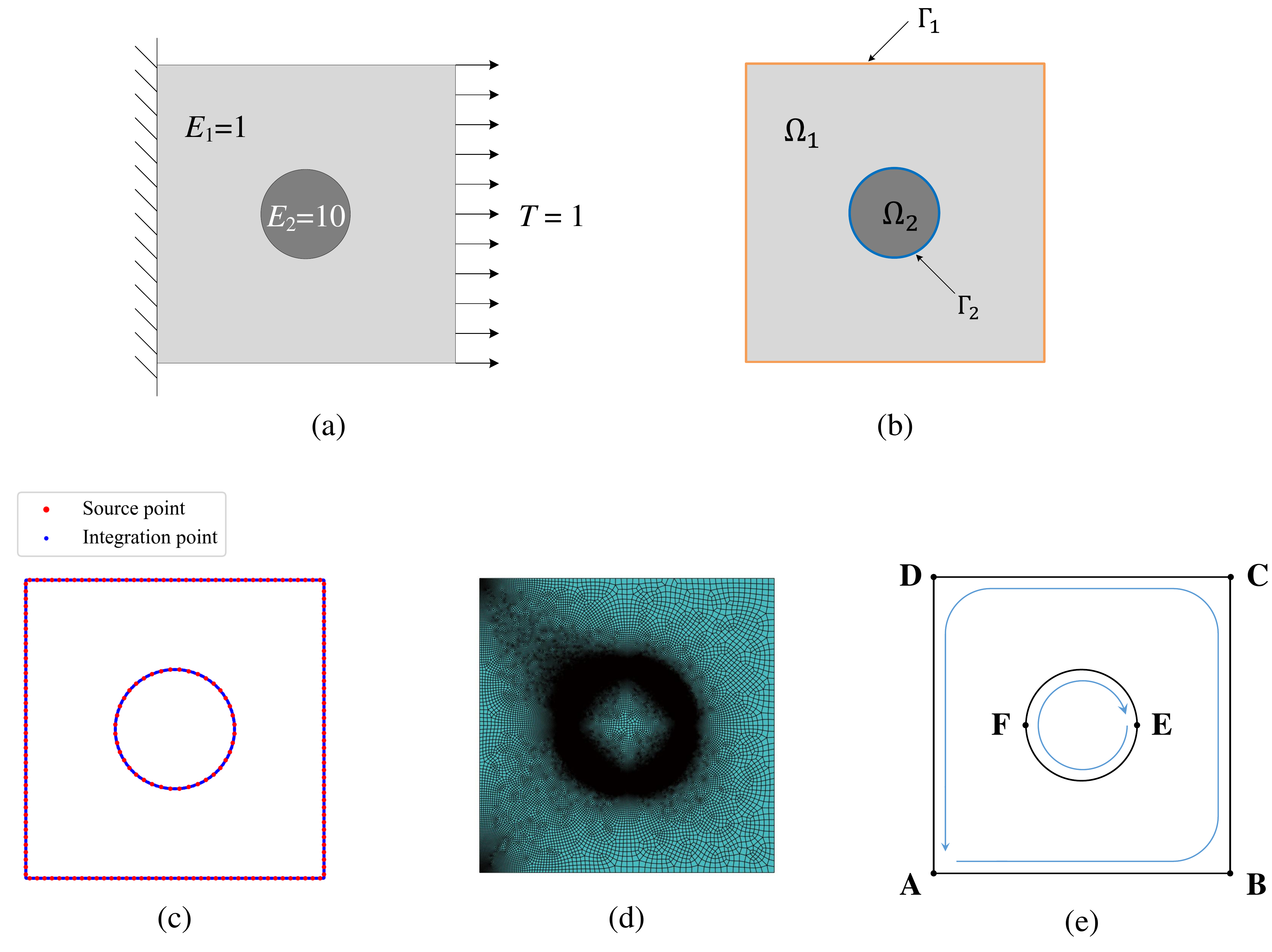}

\caption{(a) The geometry and the boundary condition for the heterogeneous problem. (b) Notations of the boundary and domain. (c) The distribution of the source points and integration points. (d) The mesh in FEM to generate the reference solution. (e) The trajectory to show the boundary results.}  
\label{geo_inclusion}  
\end{figure}

\begin{figure}
\centering
\includegraphics[width=16cm]{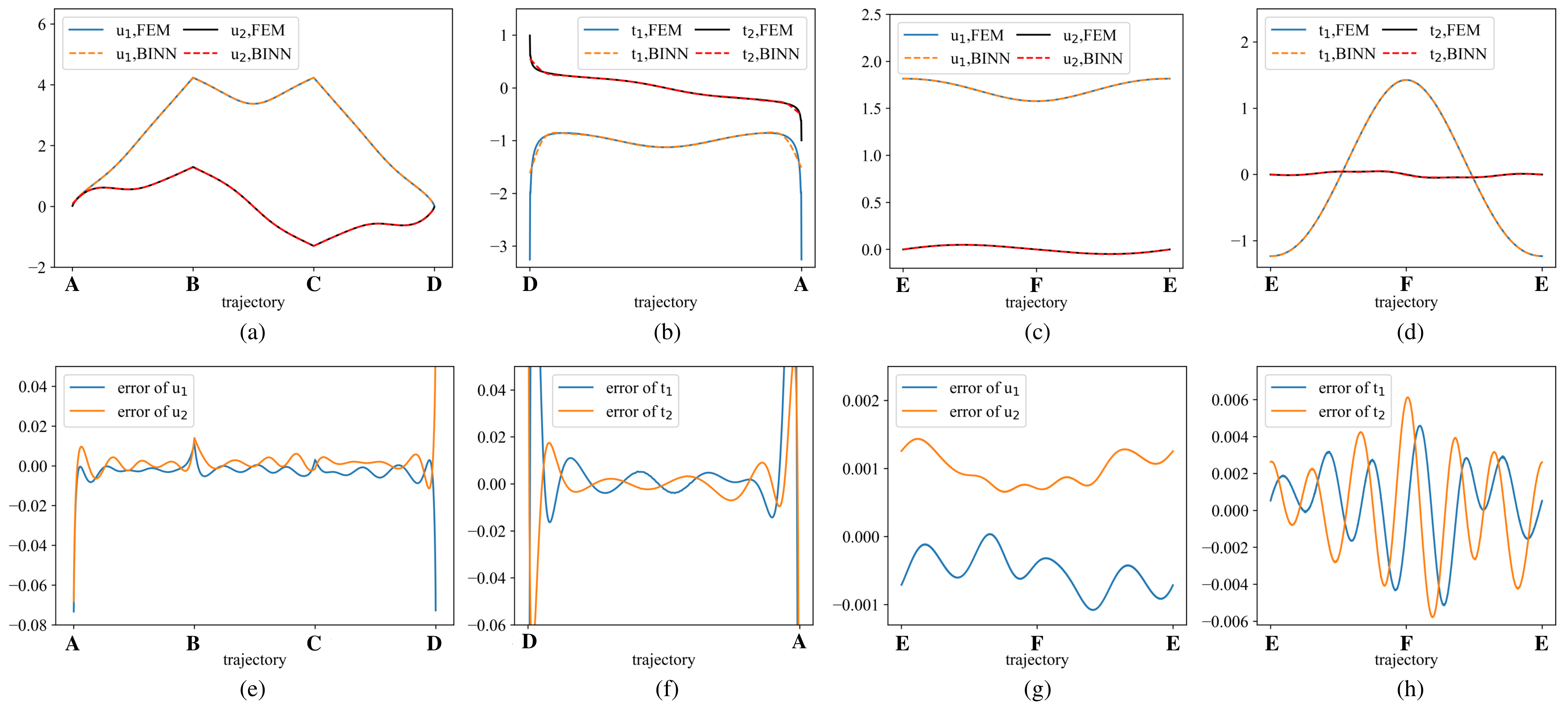}
\caption{ Results of the boundary solution to the heterogeneous problem. (a)-(d): Comparison of the solution for boundary unknowns between FEM and BINN along the trajectories.  (e)-(h): The corresponding distribution of the absolute error between BINN and the exact solution along the trajectories.}  
\label{inclusion_boundary}  
\end{figure}

\begin{figure}
\centering
\includegraphics[width=13cm]{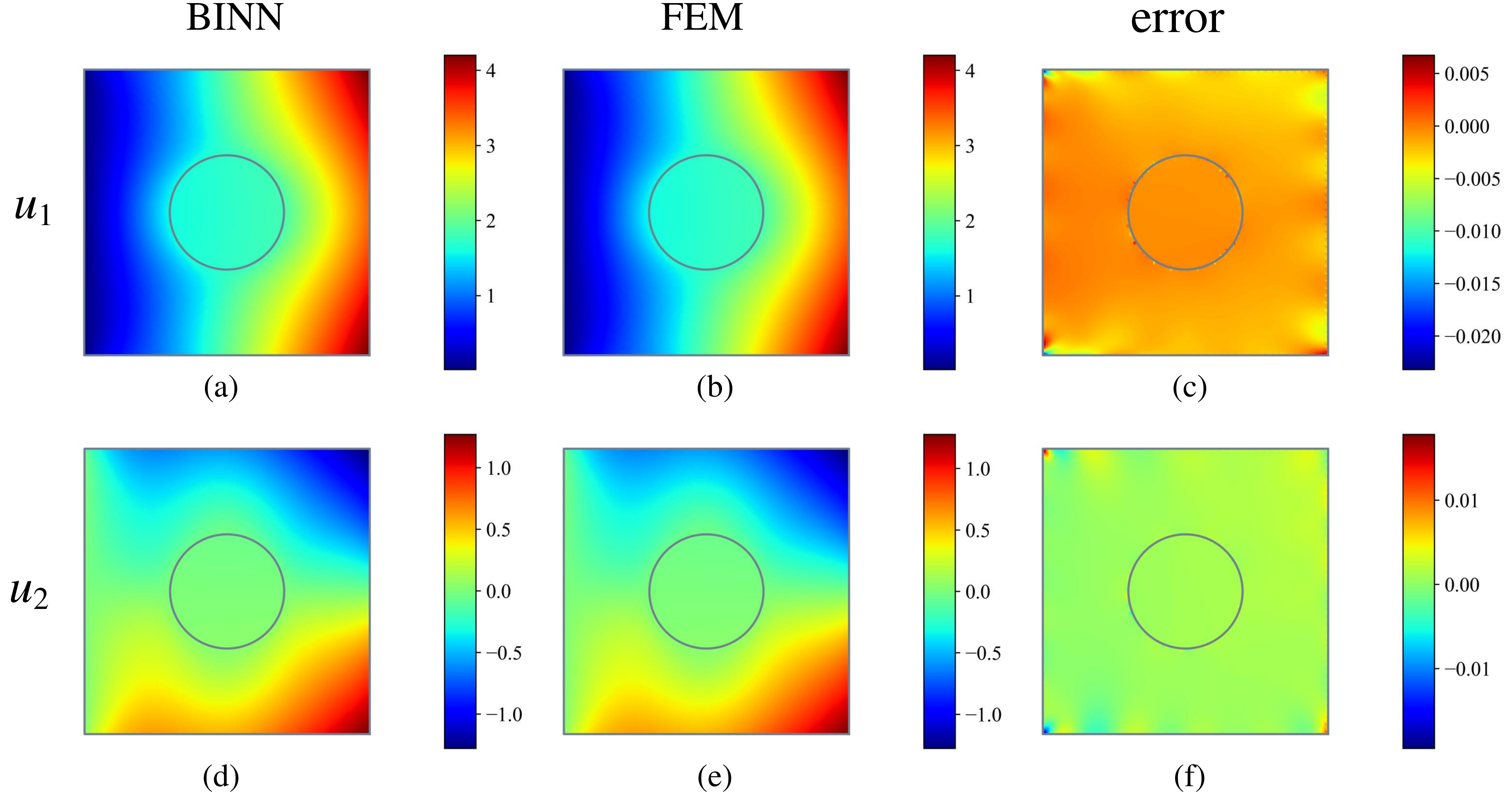}
\caption{The interior results of the heterogeneous problem. (a) and (d) are the displacement results $u_1$ and $u_2$ by BINN, respectively. (b) and (e) are the reference solutions of the displacement $u_1$ and $u_2$ by FEM, respectively. (c) and (f) are the corresponding distributions of the absolute error between BINN and FEM solution.}  
\label{inclusion_inner}  
\end{figure}

In this example, we will investigate the heterogeneous problems, which are important in the analysis of flawed structures, composite materials, meso-mechanics, etc. In traditional schemes of PINN-based models such as DCM and DRM, the networks are adopted to approximate the properties of the whole region, which is assumed to be sufficiently smooth. Thus the original versions of these methods are not suitable for heterogeneous problems since the derivatives are no longer continuous across the interface, which is hard for a single neural network to produce the exact results in the nearby regions. Therefore, some investigations such as cPINN \cite{RN53} or CENN \cite{RN115}, are using the idea of subdomains to introduce the discontinuity, i.e., the whole region is decomposed into several subdomains with respect to their material properties, and each subdomain will be assigned with an individual network that coupled with other adjacent ones on the interfaces. Such procedures can effectively overcome the discontinuity issue, but the training cost for multiple networks is quite expensive. In contrast, as we will show in the following example, BINN could solve heterogeneous problems with a single network, which is more convenient and efficient.

The geometry of the model is shown in fig.\ref{geo_inclusion}(a), a $5\times5$ square plate with a circular inclusion of radius $R=1$ is considered.  Young's modulus of the matrix and the inclusion is $E_1=1$ and $E_2=10$, respectively. The Poisson ratio for both materials is 0.3. The left side of the plate is clamped, and a uniform tension $T=1$ is applied on the right side of the plate. Plane strain condition is considered.

For heterogeneous problems, both the displacement $\boldsymbol{u}$ and traction $\boldsymbol{t}$ on the interface between the different materials are unknowns to be approximated. It should be noticed that the displacement $\boldsymbol{u}$ and traction $\boldsymbol{t}=\boldsymbol{\sigma}\cdot \boldsymbol{n}^s$ are still continuous across the interface, where $\boldsymbol{n}^s$ denotes the norm of the interface, although the spatial gradient of the displacement $\boldsymbol{u}$ is not. 
As shown in fig.\ref{geo_inclusion}(b), let $\Gamma_1$ and $\Gamma_2$ denote the boundary of the square and interface,  respectively. Then the boundary of the matrix $\Omega_1$ is $\Gamma_1\cup\Gamma_2$, and the boundary of the inclusion $\Omega_2$ is $\Gamma_2$. The boundary integral equations for the matrix and inclusion can be separately written as
\begin{subequations}
\begin{align}
&C_{\alpha \beta}(\boldsymbol{y}) u_{\beta}(\boldsymbol{y})+\int_{\Gamma_1\cup\Gamma_2} t_{\alpha \beta}^{\rm s\left(1\right)}(\boldsymbol{x}; \boldsymbol{y}) u_{\beta}(\boldsymbol{x}) {\rm d} \Gamma(\boldsymbol{x})
=\int_{\Gamma_1\cup\Gamma_2} u_{\alpha \beta}^{\rm s\left(1\right)}(\boldsymbol{x}, \boldsymbol{y}) t_{\beta}(\boldsymbol{x}) {\rm d} \Gamma(\boldsymbol{x}),\quad &\boldsymbol{y}\in \Gamma_1\cup\Gamma_2,\\
&C_{\alpha \beta}(\boldsymbol{y}) u_{\beta}(\boldsymbol{y})
+\int_{\Gamma_2} t_{\alpha \beta}^{\rm s\left(2\right)}(\boldsymbol{x}, \boldsymbol{y}) u_{\beta}(\boldsymbol{x}) {\rm d} \Gamma(\boldsymbol{x})=\int_{\Gamma_2} u_{\alpha \beta}^{\rm s\left(2\right)}(\boldsymbol{x};\boldsymbol{y}) t_{\beta}(\boldsymbol{x}) {\rm d} \Gamma(\boldsymbol{x}),\quad &\boldsymbol{y}\in \Gamma_2,
\end{align}
\label{BIE_inclusion}
\end{subequations}
where the superscript $(1), (2)$ denote the fundamental solutions with Young's modulus of $E_1, E_2$, respectively. The continuity conditions on the interface $\Gamma_2$ are: 
\begin{subequations}
\begin{align}
u_{\alpha}(\boldsymbol{x})|_{\Gamma_2^+} = u_{\alpha}(\boldsymbol{x})|_{\Gamma_2^-}\label{cont_u},\\
t_{\alpha}(\boldsymbol{x})|_{\Gamma_2^+} = -t_{\alpha}(\boldsymbol{x})|_{\Gamma_2^-}\label{cont_t}.
\end{align}
\label{BIE_interface}
\end{subequations}

In BINN, the unknowns on the boundary are approximated with a single network, hence eq(\ref{cont_u}) is automatically satisfied. To satisfy eq(\ref{cont_t}), the surface traction $t_{\beta}$ on the interface in both equations of eq(\ref{BIE_inclusion}) should be computed with the same elastic constants. In the present work, we choose $E_2$ to compute the traction. The loss function is the summation of the residuals for BIEs in eq(\ref{BIE_inclusion}):
\begin{equation}
    L^{bie}(\boldsymbol{\theta}) = \frac{1}{N_s^{\left(1\right)}+N_s^{\left(2\right)}}\sum^{N_{s}^{\left(1\right)}+N_s^{\left(2\right)}}_{i=1}\left\| \boldsymbol{R}(\boldsymbol{y}^{i\left(M\right)};\boldsymbol{\theta})\right\|^2+\beta\frac{1}{N_s^{\left(2\right)}}\sum^{N_{s}^{\left(2\right)}}_{i=1}\left\| \boldsymbol{R}(\boldsymbol{y}^{i\left(I\right)};\boldsymbol{\theta})\right\|^2,
\end{equation}
where $\boldsymbol{y}^{i\left(M\right)}$ denotes the source point on $\Gamma_1\cup\Gamma_2$, and $\boldsymbol{y}^{i\left(I\right)}$ denotes the source point on $\Gamma_2$.  Note that  $\boldsymbol{y}^{i\left(M\right)}$ and $\boldsymbol{y}^{i\left(I\right)}$ are coincide on $\Gamma_2$.
${N_s^{\left(1\right)}}$ and ${N_s^{\left(2\right)}}$ denote the numbers of source points on $\Gamma_1$ and $\Gamma_2$, respectively. $\beta$ is a parameter to adjust the magnitude of the terms due to the different elastic constants. We suggest $\beta$ to be exactly the ratio of Young's modulus between the two materials, which is 10 in this example.

200 source points with 2000 integration points are allocated in this example, as shown in fig.\ref{geo_inclusion}(c). The network is trained by 100000 iterations. As a comparison, a reference solution is calculated using FEM by ABAQUS (version 6.14) with a very fine mesh (175313 quadratic quadrilateral elements of type CPE8), as shown in fig.\ref{geo_inclusion}(d). The mesh is refined especially around the corner of the fixed boundary and the interface to gain enough accuracy. The results on the boundary are presented along the trajectory shown in fig.\ref{geo_inclusion}(e). The solution of the boundary unknowns is shown in fig.\ref{inclusion_boundary}. For trajectory $\textbf{A}$-$\textbf{B}$-$\textbf{C}$-$\textbf{D}$, the unknowns are the displacement $\boldsymbol{u}$ as shown in fig.\ref{inclusion_boundary}(a), while for trajectory $\textbf{D}$-$\textbf{A}$, the unknowns are the traction $\boldsymbol{t}$ as shown in fig.\ref{inclusion_boundary}(b). 
On the interface $\textbf{E}$-$\textbf{F}$-$\textbf{E}$, both $\boldsymbol{u}$ and $\boldsymbol{t}$ are unknowns to be solved, and the results are shown in fig.\ref{inclusion_boundary}(c) and (d), respectively. The corresponding error distributions are shown in fig.\ref{inclusion_boundary}(e)-(h). It can be observed from fig.\ref{inclusion_boundary}(e) and (f) that the results near the corner $\mathbf{A}$ and $\mathbf{D}$ are relatively inferior, which is due to the strong stress concentration around the corner, and the traction changes dramatically in the nearby region. As demonstrated in fig.\ref{geo_inclusion}(c), we did not employ any refinement around the corner in BINN, thus the sparse distribution of the integration points can not capture such rapid changes of the traction, which will also influence the accuracy of the displacement results in the nearby region. Nevertheless, the results are still accurate in other parts of the boundary.
The results on the interior region are shown in fig.\ref{inclusion_inner}. It can be seen that the results from BINN agree well with the FEM solution.

\section{Concluding Remarks}
\label{sec_conclusion}
In this article, we proposed BINN, an architecture for solving PDEs with neural networks based on boundary integral equations. The neural networks are employed as the function approximation machine and the BIE formulas are embedded as the constraint in the loss function. We demonstrated the differences and the relations of the proposed method with the existing Deep Collocation method (DCM) and Deep Ritz method (DRM) in view of different statements of the weighted residual method (WRM): Unlike DCM that can be derived from the original statement of WRM, or DRM derived from the weak statement of WRM, BINN is derived from the inverse statement of WRM. We also illustrated the differences between PINN and the traditional BEM, which is a widely used technique also based on BIE. We discussed the principle of how to choose the strategy for evaluating the singular integrals in BINN, and regularization techniques are suggested and employed in this work. As a demonstration, we investigated the performance of BINN with potential problems governed by Laplace equations and the elastostatic problems governed by Navier equations. Unlike Deep Collocation method and Deep Ritz method, the loss function in BINN only contains the residuals of BIE, where all the boundary conditions have been naturally considered without any special construction on the approximate function. Hence BINN can be easily employed to arbitrarily shaped regions. As a BIE-based method, BINN can be conveniently implemented to the problems on infinite/semi-infinite regions. Moreover, BINN could solve heterogeneous problems with a single network, without suffering from discontinuity on the interface. Numerical examples have shown the remarkable performance of BINN on these problems.


A limitation of BINN is that it relies on the existence of the BIE formulation and the fundamental solution. In many non-linear problems, the unknowns in the interior region cannot be completely eliminated. Moreover, the fundamental solution may be hard to obtain or even not exist. Note that the traditional BEM has already been employed to these non-linear problems. The fundamental solution in the linear case is usually adopted, and there will be an extra domain integral that involves the unknowns in the interior region. There are several investigations focusing on the treatment of these domain integrals \cite{Nowak_1,Gao_1}, and some of them may still be available in BINN. Alternatively, we would like to mention a recent research named DeepGreen \cite{deepgreen} that aims at finding the fundamental solution for non-linear problems with deep neural networks, which may also be employed to extend the power of BINN.

Another possible direction to improve the present version of BINN is combining it with the fast multipole method (FMM). On the one hand, the loss function in BINN involves the computation of boundary integral, which can be efficiently computed with FMM. On the other hand, FMM has to combine with an iterative solver, which is compatible with the training process of BINN. The combination of BINN and FMM may be meaningful for large-scale problems.

Finally, it should be also clarified that the proposed method is still in its early stage. The depth and breadth of the present research are still far to be comparable with that of traditional methods such as BEM. There are still many problems to investigate on BINN such as the efficiency, convergence, or robustness for specific problems. However, the proposed method is an important supplement to the existing PINN-based methods and has the potential to be developed along with the fast growth of data-driven scientific computing.


\section*{Acknowledgments}
This study is supported by the projects from the National Natural Science Foundation of China, under Grant No.11672155 and No.12090033.

\appendix
\section{The condition analysis for the case that the source point is out of the interested domain}
\label{App_cond}
As demonstrated in remark \ref{remark_cond}, the singularity of the boundary integral can be removed by allocating the source point out of the interested domain instead of on the boundary. However, the problem will be much more pathological.
Taking the Poisson equation as an example, the non-singular BIE can be derived by substituting eq(\ref{poi_inv3}) into eq(\ref{poi_weight}) and employing the property of the Dirac delta function:
\begin{equation}
\begin{aligned}
&\int_{\Gamma_1}\frac{\partial u(\boldsymbol{x})}{\partial \boldsymbol{n}}u^{s}(\boldsymbol{x;y})d\Gamma
    -\int_{\Gamma_2}\frac{\partial u^{s}(\boldsymbol{x;y})}{\partial \boldsymbol{n}}u(\boldsymbol{x})d\Gamma=\int_{\Gamma_1}\frac{\partial u^{s}(\boldsymbol{x;y})}{\partial \boldsymbol{n}}\bar{u}(\boldsymbol{x})d\Gamma
    \\
    &-\int_{\Gamma_2}\bar{q}u^{s}(\boldsymbol{x;y})d\Gamma
    -\int_{\Omega} f(\boldsymbol{x})u^{s}(\boldsymbol{x;y})d\Omega,\quad \boldsymbol{y}\in \mathbb{R}^{n_d}\backslash(\Omega\cup \Gamma)
\end{aligned}
\label{eq_nonsingluar}
\end{equation}

 The condition analysis can be roughly made by recalling the Fredholm theorem \cite{Korn_1}. For problems with pure Neumann BC, the eq(\ref{poi_bie}) corresponds to the Fredholm equation of the second kind, which is usually well-posed, while eq(\ref{eq_nonsingluar}) corresponds to the Fredholm equation of the first kind, whose condition is much worse. For problems with pure Dirichlet BC, both eq(\ref{poi_bie}) and eq(\ref{eq_nonsingluar}) correspond to the Fredholm equation of the first kind, but 
it has been proved that eq(\ref{poi_bie}) with weakly-singular kernels has better condition \cite{Hsiao_1}. And of course, there is a spectral problem for the mixed boundary conditions. Although not presented in this article, we have tried the form of eq(\ref{eq_nonsingluar}) to formulate BINN, and according to our observation, the accuracy and robustness are much less due to its pathological property.
\section{Proof for the regularity of the integrand in eq(\ref{eq_weakreg})}
\label{App_1}
Suppose the function $f(t)$ in eq(\ref{eq_weakreg}) satisfies the Lipschitz condition, i.e.,
\begin{equation}
|f(x_1)-f(x_2)|<K|x_1-x_2|,\quad \forall x_1,x_2 \in [-a,a]
\label{eq_Lipschitz}
\end{equation}
where $K\in \mathbb{R}$ is a constant. Then we can easily derive that:
\begin{equation}
\lim_{t\to 0} \lvert\ln{\lvert t\rvert}\left[f(t)-f(0)\right]\rvert<\lim_{t\to 0}K|t\ln{|t|}|=0,
\end{equation}
Hence the integrand has no singularity at $t=0$.
\section{The fundamental solution of elastostatic problems for half-plane}
\label{App_2}

For elastostatic problems on half-plane, the fundamental solution $\boldsymbol{u}^s$ and $\boldsymbol{t}^s$ can be written as the superposition of two parts \cite{Telles_2}:
\begin{equation}
\begin{aligned}
    \boldsymbol{u}^s(\boldsymbol{x},\boldsymbol{y}) = \boldsymbol{u}^{s(K)}(\boldsymbol{x},\boldsymbol{y})+ \boldsymbol{u}^{s(C)}(\boldsymbol{x},\boldsymbol{y})\\
    \boldsymbol{t}^s(\boldsymbol{x},\boldsymbol{y}) = \boldsymbol{t}^{s(K)}(\boldsymbol{x},\boldsymbol{y})+ \boldsymbol{t}^{s(C)}(\boldsymbol{x},\boldsymbol{y})
\end{aligned}
\end{equation}
where $\boldsymbol{u}^{s(K)}$ and $\boldsymbol{t}^{s(K)}$ denotes the Kelvin solution for the 2D case as shown in eq(\ref{fund_elastic}). $\boldsymbol{u}^{s(C)}$ is the auxiliary solution that can be expressed as
\begin{equation}
\begin{aligned}
    &u_{11}^{s(C)} = K_d\left\{-[8(1-\nu)^2-(3-4\nu)]\ln{R} + \frac{[(3-4v)R^2_1-2c\bar{x}]}{R^2}+\frac{4c\bar{x}R^2_1}{R^4} \right\}\\
    &u_{12}^{s(C)} = K_d\left\{\frac{(3-4\nu)r_1r_2}{R^2}+\frac{4c\bar{x}R_1r_2}{R^4}-4(1-\nu)(1-2\nu)\theta \right\}\\
    &u_{21}^{s(C)} = K_d\left\{\frac{(3-4\nu)r_1r_2}{R^2}-\frac{4c\bar{x}R_1r_2}{R^4}+4(1-\nu)(1-2\nu)\theta \right\}\\
    &u_{22}^{s(C)} = K_d\left\{-[8(1-\nu)^2-(3-4\nu)]\ln{R} + \frac{[(3-4v)r^2_2+2c\bar{x}]}{R^2}-\frac{4c\bar{x}r^2_2}{R^4} \right\}\\
\end{aligned}
\end{equation}
\setcounter{figure}{0}    
\begin{figure}
\centering  
\includegraphics[width=6cm]{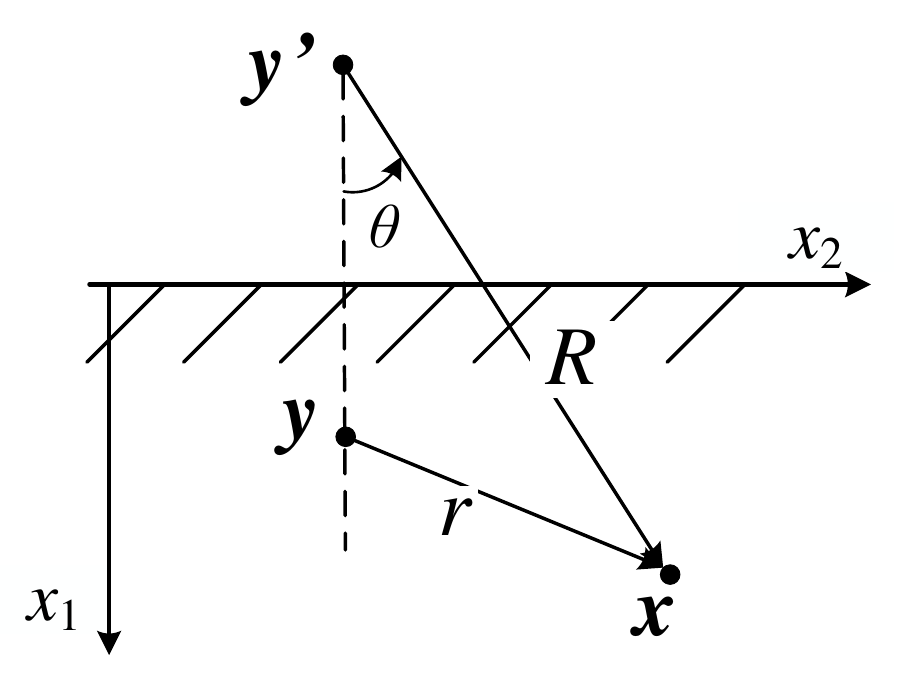}
\caption{Illustration of the auxiliary solution.}  
\label{Aux}  
\end{figure}
The meaning of the notations are illustrated in fig.\ref{Aux}. $\boldsymbol{y}'$ is the mirror point of $\boldsymbol{y}$ with respect to the interface, and
\begin{equation}
\begin{aligned}
   &r_{\alpha} = x_{\alpha}(\boldsymbol{x})-x_{\alpha}(\boldsymbol{y}),\quad
   &R_{\alpha} = x_{\alpha}(\boldsymbol{x})-x_{\alpha}(\boldsymbol{y}'),\\
   &r=(r_{\alpha}r_{\alpha})^{1/2},\quad &R=(R_{\alpha}R_{\alpha})^{1/2},\\
   &c=x_1(\boldsymbol{y}),\quad
   &\bar{x}=x_1(\boldsymbol{x}),\\
   &\theta = \arctan{(\frac{R_2}{R_1})},\quad
   &K_d = \frac{1}{8\pi(1-\nu)G}
\end{aligned}
\end{equation}

The corresponding traction solution $\boldsymbol{t}^{s(C)}$ can be calculated with:
\begin{equation}
    t^{s(C)}_{\alpha\beta} = \sigma^{s(C)}_{\alpha\beta\gamma}n_{\gamma},\quad \alpha,\beta,\gamma=1,2
\end{equation}
where $n_{\gamma}$ denotes the component of the outward normal vector, $\sigma^{s(C)}_{\alpha\beta\gamma}$ is the stress solution derived from the displacement solution  $u^{s(C)}_{\alpha\beta}$ by employing the geometric equation and constitutive law.

 \bibliographystyle{elsarticle-num} 
 \bibliography{manuscript}





\end{document}